\documentclass[journal]{IEEEtran}

\makeatletter
\long\def\@makecaption#1#2{\ifx\@captype\@IEEEtablestring%
\footnotesize\begin{center}{\normalfont\footnotesize #1}\\
{\normalfont\footnotesize\scshape #2}\end{center}%
\@IEEEtablecaptionsepspace
\else
\@IEEEfigurecaptionsepspace
\setbox\@tempboxa\hbox{\normalfont\footnotesize {#1.}~~ #2}%
\ifdim \wd\@tempboxa >\hsize%
\setbox\@tempboxa\hbox{\normalfont\footnotesize {#1.}~~ }%
\parbox[t]{\hsize}{\normalfont\footnotesize \noindent\unhbox\@tempboxa#2}%
\else
\hbox to\hsize{\normalfont\footnotesize\hfil\box\@tempboxa\hfil}\fi\fi}
\makeatother

\usepackage{amsmath,amsfonts}
\usepackage{algorithmic}
\usepackage{algorithm}
\usepackage{array}
\usepackage[caption=false,font=footnotesize,labelformat=simple]{subfig}
\usepackage[switch]{lineno}
\usepackage{textcomp}
\usepackage{stfloats}
\usepackage{url}
\usepackage{verbatim}
\usepackage{graphicx}
\usepackage{cite}
\usepackage{svg}
\usepackage{xcolor}
\usepackage{siunitx}
\usepackage{multirow}
\usepackage{booktabs}
\usepackage{rotating}
\usepackage{pdflscape}
\usepackage{cprotect} %
\usepackage{enumitem} 
\usepackage{soul}

\newcommand{\etal}{\textit{et al.}~}

\newcommand{\eg}{\textit{e.g.},~}

\newcommand\added[1]{\textcolor{black}{#1}}
\newcommand\removed[1]{}

\begin{document}
\bstctlcite{IEEEexample:BSTcontrol}

\title{Beyond Mortality: Advancements in Post-Mortem Iris Recognition through Data Collection and Computer-Aided Forensic Examination}

\author{Rasel Ahmed Bhuyian,~\IEEEmembership{Student Member,~IEEE,}
        Parisa Farmanifard,
        Renu Sharma,
        Andrey Kuehlkamp,
        Aidan Boyd,
        Patrick J. Flynn,~\IEEEmembership{Fellow,~IEEE},
        Kevin W. Bowyer,~\IEEEmembership{Fellow,~IEEE},
        Arun Ross,~\IEEEmembership{Senior Member,~IEEE},
        Dennis Chute, and
        Adam Czajka,~\IEEEmembership{Senior Member,~IEEE}
\thanks{R. A. Bhuyian, P. J. Flynn, K. W. Bowyer, A. Kuehlkamp and A. Czajka are with the University of Notre Dame, Notre Dame, IN 46556, USA. Corresponding email address: aczajka@nd.edu}
\thanks{P. Farmanifard, R. Sharma and A. Ross are with the Department of Computer Science and Engineering, Michigan State University, East Lansing, MI 48824, USA.}
\thanks{A. Boyd is currently with Meta, New York, NY 10003, USA. Boyd was with the University of Notre Dame, Notre Dame, IN 46556, USA, during conducting this study.}
\thanks{Dennis Chute is with the Dutchess County Medical Examiner's Office, Poughkeepsie, NY 12601, USA.}
\thanks{Manuscript accepted for publication in {\it IEEE Transactions on Biometrics, Behavior, and Identity Science}; early access link: https://ieeexplore.ieee.org/document/11063436}}

\markboth{IEEE Transactions on Biometrics, Behavior, and Identity Science,~Vol.~, No.~, August~2022}%
{Bhuyian \MakeLowercase{\textit{et al.}}: Beyond Mortality: Advancements in Post-Mortem Iris Recognition ...}

\maketitle

\def\datasetname{NIJ-2018-DU-BX-0215}

\begin{abstract}
Post-mortem iris recognition brings both hope to the forensic community (a short-term but accurate and fast means of verifying identity) as well as concerns to  society (its potential illicit use in post-mortem impersonation). These hopes and concerns have grown along with the volume of research in post-mortem iris recognition. Barriers to further progress in post-mortem iris recognition include the difficult nature of data collection, and the resulting small number of approaches designed specifically for comparing iris images of deceased subjects. This paper makes several unique contributions to mitigate these barriers. First, we have collected and we offer a new dataset of NIR (compliant with ISO/IEC 19794-6 where possible) and visible-light iris images collected after demise from 259 subjects, with the largest PMI (post-mortem interval) being 1,674 hours. For one subject, the data has been collected before and after death, the first such case ever published. Second, the collected dataset was combined with publicly-available post-mortem samples to assess the current state of the art in automatic forensic iris recognition with five iris recognition methods and data originating from 338 deceased subjects. These experiments include analyses of how selected demographic factors influence recognition performance. Thirdly, this study implements a model for detecting post-mortem iris images, which can be considered as presentation attacks. Finally, we offer an open-source forensic tool integrating three post-mortem iris recognition methods with explainability elements added to make the comparison process more human-interpretable.

\end{abstract}

\begin{IEEEkeywords}
post-mortem iris recognition, presentation attack detection, iris image datasets.
\end{IEEEkeywords}

\section{Introduction}

\IEEEPARstart{T}{he} iris is a major biometric modality, and iris recognition has witnessed more than three decades of development \cite{Daugman_PAMI_1993}.
Applications of iris recognition have been expanded recently to law enforcement and criminal justice (\eg the FBI Next Generation Identification Iris Service \cite{FBI_NGI_webpage}). Post-mortem iris recognition was not considered viable until 2016 \cite{Trokielewicz_BTAS_2016,Bolme_BTAS_2016}, when successful matching was demonstrated for samples acquired within a few days of demise. Later studies \cite{Trokielewicz_TIFS_2019,Boyd_Access_2020} demonstrated that in favorable conditions (\eg low temperatures in which deceased individuals are kept, and lack of physical damage to the eyes), iris recognition can be performed for up to three weeks after demise. With iris image datasets constantly growing, and iris recognition proving its usefulness in law enforcement, post-mortem iris recognition has become an important biometrics-based modality in the forensic identification toolkit.

Understanding the capabilities of post-mortem iris recognition and its applicability are, however, still limited due to two main factors: (a) the exceptionally complicated process of data collection from deceased subjects and, hence, a very small number of longitudinal studies, especially related to pre-mortem vs. post-mortem comparisons, and (b) a lack of automated methods designed specifically for forensic examiners, that would both yield a candidate list of identities matching a post-mortem sample and facilitate forensic human examination of iris images. This paper describes the efforts of a large interdisciplinary team towards filling these gaps by offering the {\bf following contributions}:

\begin{itemize}[leftmargin=.15in]
    \item {\bf The NIJ-2018-DU-BX-0215 dataset}, acquired from 259 deceased subjects and composed of 5,770 near-infrared (NIR) and 4,643 visible-light (RGB) iris images captured in multiple sessions. The dataset has several unique elements: (a) the largest PMI (post mortem interval) (1,674 hours), (b) one case for which both pre-mortem (acquired near the time of death) and post-mortem samples have been obtained, (c) one case in which eyeball was dislocated outside the bony socket due to injury. The image data is accompanied by metadata including gender, age, PMI, and eye (left/right), and also by post-mortem iris-specific segmentation masks, generated separately for iris texture, post-mortem cornea wrinkles, and occlusions, all predicted by a deep learning-based segmentation model trained specifcally on human-annotated post-mortem iris samples.

    \item A series of {\bf presentation attack detection} (PAD) experiments using a state-of-the-art iris PAD model \cite{sharma2020d}, to determine if post-mortem irises can be detected as spoofs. We demonstrate that fine-tuning this PAD approach using a few samples of post-mortem iris images is sufficient to create an effective detector of post-mortem irises.

    \item A {\bf human examination software tool}, \verb+PMExpert+, incorporating three post-mortem iris recognition and explainability methods, along with ISO/IEC 29794-6 iris image quality metrics, to support human forensic examiners.
    
    \item A set of comprehensive {\bf post-mortem iris recognition experimental results} using five iris matchers implementing different approaches to iris recognition (classical Gabor-wavelets-based, human-saliency-based, deep learning-based and commercial). To deliver authoritative results, the experiments have been performed on the largest-to-date dataset of post-mortem samples sourced from already-published forensic iris datasets and from the NIJ-2018-DU-BX-0215 set, representing 338 deceased subjects.

    \item Analyses of how {\bf demographic factors} (age at time of death and gender) impact post-mortem iris recognition. 
    
\end{itemize}

The \datasetname~dataset and the forensic iris recognition tool \verb+PMExpert+ are offered with this paper and can be already obtained at no cost via the National Archive of Criminal Justice Data (NACJD) service \cite{Czajka_NACJD_2023}.

\section{Related Work}

This section summarizes important milestones achieved to date in post-mortem iris recognition area, related to data acquisition, our understanding of the feasibility of forensic iris recognition, and algorithmic advances.

\subsection{Post-mortem Iris Image Datasets}
\label{sec:related:datasets}

Collecting biometric data from deceased subjects is different from, and more difficult than, collecting data from living subjects. Collections need to be done by forensic technicians or medical doctors, and take place in mortuary or forensic anthropology facilities (``body farm''). Samples collected at crime scenes may not be approved for redistribution to research communities. Although collecting biometric samples from deceased subjects is not categorized as human subjects research in the United States, local regulations may call for such collections to undergo an approval from an ethics committee, based in part on a common-sense understanding of respect for the dead and members of the decedent's family. These realities explain why the existing datasets are fragmentary, poorly-balanced (in terms of demographic dimensions, PMI, and/or environmental factors), and relatively small. 

Trokielewicz \etal \cite{trokielewicz2016post,trokielewicz2016human} released the first public dataset of post-mortem iris images that we know of, {\it Warsaw-BioBase-Post-Mortem-Iris v1.0}. The dataset contains 480 NIR and 850 RGB images from 17 subjects, along with metadata such as age, gender, and cause of death. Images were captured using an IriShield M2120U iris sensor and Olympus TG-3 consumer-grade camera, with two to eight acquisition sessions per eye depending on subject availability. The first session was conducted 5-7 hours after death, while subsequent sessions were taken at intervals up to 17 days post-mortem. Its extension, {\it Warsaw-BioBase-Post-Mortem-Iris v2.0} \cite{trokielewicz2018iris}, contains 1,787 RGB and 1,200 NIR images from 37 subjects, captured up to 814 hours after death in the hospital mortuary. The third collection offered by Trokielewicz \etal \cite{trokielewicz2020post}, {\it Warsaw-BioBase-Post-Mortem-Iris v3.0}, includes 1,094 NIR and 785 RGB images from 42 cadavers, taken up to 369 hours after death. Most environmental conditions in these three collections are unknown, such as where cadavers were kept before entering the cold storage, air pressure and humidity. However, the mortuary room temperature was kept stable at approximately \ang{6} Celsius (\ang{42.8} Fahrenheit).

To our knowledge, the above three datasets are the only publicly-available collections offered to date, comprising 2,294 NIR and 2,572 RGB images from 79 deceased subjects, with a maximum PMI of 814 hours. None of these datasets, contrary to the new dataset described in this paper, contains images captured pre- and post-mortem from the same eyes.

\subsection{Automated Detection of Post-mortem Iris Images}

Iris recognition systems may lack the ability to detect if a presented iris pertains to a living or a deceased person. This is a vulnerability that could potentially allow unauthorized access if attackers use preserved iris patterns of deceased individuals to bypass security measures.

While there is a considerable body of research on post-mortem iris recognition, the specific area of post-mortem iris detection is relatively less explored.
Studies in this area typically focus on identifying unique post-mortem changes in iris texture, the absence of physiological responses (like pupillary light reflex), and structural alterations that occur after death. One previous approach known to us~\cite{trokielewicz2018presentation} involves leveraging deep learning techniques to classify iris images as those captured from living individuals or from deceased cases. 

Among the methods designed for living subjects, D-NetPAD~\cite{sharma2020d} demonstrated its success in the LivDet-Iris 2020 competition~\cite{das2020iris}.
It's based on the densely connected neural network architecture of DenseNet-121~\cite{Huang2017}, and allows for learning nuanced differences between live and post-mortem irises, in particular. The original D-NetPAD solution was thus selected to be adapted to this study. 

\subsection{Feasibility Studies of Post-mortem Iris Recognition}
\label{sec:related:feasibility}

To our knowledge, Sansola was the first to study the feasibility of post-mortem human iris recognition \cite{sansola2015postmortem}, carried out for images captured by IriShield M2120U iris sensor from 43 deceased subjects, and utilizing IriCore iris recognition software \cite{IriCore2013}. Depending on the PMI, the method resulted in false non-matches ranging from 19-30\% and no false matches were recorded. Sansola also found that the color of the eyes was correlated with the comparison scores, with blue/gray eyes resulting in lower correct match rates (59\%) compared to brown (82\%) or green/hazel eyes (88\%). 

Due to the challenges with data collection from deceased human subjects, some researchers opted to conduct experiments on animals. Saripalle \etal \cite{saripalle2015post} conducted a study for 17 eyes of domestic pigs, and found that the biometric capabilities of irises degrade gradually after being removed from the animal's body and are lost between 6 to 8 hours after death. Daugman's iris recognition approach \cite{daugman2007new} was used to assess the biometric potential of the extracted iris tissues. 

The first systematic longitudinal study of post-mortem iris recognition accuracy as a function of PMI, for data captured from 17 cadavers, was offered by Trokielewicz \etal \cite{trokielewicz2016post}, who demonstrated that iris can continue to be used as a biometric identifier for up to 27 hours after death. Three commercial matchers, IriTech \cite{IriCore2013}, VeriEye and MIRLIN \cite{mirlin2013}, and one open-source OSIRIS method \cite{othman2016osiris} were used in this study. 
Further, Trokielewicz \etal \cite{trokielewicz2016human} extended their experiments to a longer time span, including samples collected up to 17 days after death from the same 17 deceased people. In the short-term (60 hours after death) analysis they found that the IriCore method performed well, with an EER as low as 13\%. However, considering all samples collected up to 17 days, their long-term analysis revealed significant deterioration in iris recognition reliability: although correct matches could still occur, they were infrequent.

Bolme \etal \cite{bolme2016impact} first investigated the post-mortem biometric performance of face, fingerprint, and iris recognition during human decomposition in outdoor conditions. They conducted their experiment using 12 cadavers placed in an outdoor field laboratory (a ``body farm'') to assess the effect of time and environmental conditions on biometric recognition performance. The authors found that fingerprints and faces were moderately resilient to decomposition, but the irises degraded quickly, becoming useless for recognition only a few days after exposure to outdoor conditions.

A later study by Sauerwein \etal \cite{sauerwein2017effect} showed that irises can still be readable for up to 34 days after death if the cadavers were exposed to low temperatures in winter. In their study, instead of iris recognition algorithms, human examiners were asked to compare iris image pairs. These observations suggest that low temperatures increase the chances of recognizing an iris in a cadaver left outside for a longer period.

The studies listed above focused on the assessment of post-mortem iris recognition ending up with cumulative statistics over a dataset sample. Trokielewicz \etal offered the first detailed look at changes to both local and global iris features that occur during decay of a particular eye \cite{Trokielewicz_JFS_2020}. This paper also confirms what is known for live iris recognition: iris features observed in near infrared light are better (and for a longer time) visible than features acquired in visible light.

\subsection{Post-mortem Iris Recognition Methods}

The feasibility studies summarized in Sec.~\ref{sec:related:feasibility} deployed iris recognition approaches designed for live subjects. The decomposition process, and associated deformations encountered in human eyes, certainly call for forensic-specific algorithms.

Trokielewicz and Czajka \cite{trokielewicz2018data} proposed the first known to us data-driven method for post-mortem iris image segmentation, using a convolutional neural network based on SegNet  \cite{Badrinarayanan_ArXiv_2015}, fine-tuned with 1,300 manually-annotated post-mortem iris images from 37 subjects (extended to experiments with 79 deceased subjects in \cite{trokielewicz2020post}). The model could recognize specific distortions that exist in post-mortem samples but are absent in live irises. 
Later, Trokielewicz \etal \cite{trokielewicz2019iris, trokielewicz2020eyedecay} demonstrated that the post-mortem iris recognition pipeline also benefits from feature extraction approaches that are more robust to the decay of iris tissue. The authors used Siamese networks to learn filter kernels that describe iris patterns affected by post-mortem changes and proposed a hybrid filter bank, which included a combination of Gabor and Siamese networks-learned kernels. This approach resulted in lower error rates for all post-mortem intervals investigated, compared to a state-of-the-art commercial iris matcher.

Application of post-mortem iris recognition to forensic examinations may require incorporating human examiners into the image comparison process. Trokielewicz \etal \cite{trokielewicz2019perception} conducted the first study comparing features used by human subjects and an example algorithm based on deep convolutional neural network, when comparing forensic iris images. The authors compared the fixations obtained in eye tracking-based experiments with the neural networks' saliency maps, and found that the machine- and human-sourced cues about matching and non-matching regions rarely overlap, and that merging human and machine responses allowed for more accurate decisions than either source alone.

Encouraged by the above conclusions about potential complementarity of human- and machine-picked cues, Kuehlkamp \etal \cite{kuehlkamp2022interpretable} developed a deep learning-based end-to-end iris recognition system designed to provide interpretable results for forensic iris matching. Owing to this approach, the examiners are supported by the machine while judging about the samples. The reverse information flow is also possible, in that algorithms can benefit from the human operator's expertise. Indeed, Boyd \etal proposed the first human-interpretable post-mortem iris recognition algorithm designed with taking human perceptual capabilities into account. Their method first learned what features (or ``patches'') are most useful for humans, without attributing them to anatomical structures (e.g., iris crypts), and then detecting and matching such features automatically for new image pairs. The method uses many fewer ``keypoints'' than other general-purpose keypoint-based approaches (such as SIFT \cite{Lowe_JCV_2004}) to achieve the same accuracy level, making this method more useful for human experts.

\subsection{Identified Research Gaps Addressed by This Study}

The achievements in post-mortem iris recognition summarized above demonstrate the feasibility of identification of deceased subjects, and suggest potential forensic applications. The most obvious research gaps identified are: (a) datasets of post-mortem iris images that are very small and sparse, in terms of demographic and environmental factors, (b) a limited number of approaches and ready-to-use tools that would support the work of forensic examiners, (c) a lack of analyses of how post-mortem iris recognition performance depends on selected demographic and health factors, and (d) a lack of larger-scale experiments with modern approaches demonstrating the ability to automatically detect attempts to use cadaver eyes in presentation attacks. This paper presents results aimed to reduce these gaps.

\section{\datasetname~Dataset}

\begin{table*}[!thb]
\centering
\caption{Main novel characteristics of the newly-collected NIJ-2018-DU-BX-0215 post-mortem iris dataset compared with all existing publicly-available research datasets.}
\label{tab:datasets}
\begingroup
\footnotesize
\begin{tabular}{lccccccc}
\toprule
 {\bf Dataset name} & {\bf Number} & {\bf Number} & {\bf Number of cases} & {\bf Sensors used} & {\bf Min $\rightarrow$ Max} & {\bf Metadata}\\
& {\bf of subjects} & {\bf of images} & {\bf with ante-} &  & {\bf PMI} & \\
& {\bf (eyes)} &{\bf (NIR / RGB)} & {\bf and post-mortem}& & {\bf (hours)} & \\
& &  & {\bf image pairs} & & & \\
 \midrule
 Warsaw-BioBase-Post-Mortem-Iris & 17 (34) & 574 / 626 & 0 & IriShield M2120U & 5 $\rightarrow$ 814 & {\bf For all:} \\
  v1.0 \cite{trokielewicz2016human} & & & & Olympus TG-3 & & PMI \\
 Warsaw-BioBase-Post-Mortem-Iris& 20 (40) & 1,094 / 1,023 & 0 & IriShield M2120U & 7 $\rightarrow$ 453 & gender \\
  v2.0 \cite{trokielewicz2018iris} & & & & Olympus TG-3 & & age \\
 Warsaw-BioBase-Post-Mortem-Iris& 42 (84) & 764 / 785 & 0 & IriShield M2120U & 7 $\rightarrow$ 369 & reason of death \\
  v3.0 \cite{trokielewicz2020post} & & & & Olympus TG-3 & & use of eye drops\\\cline{1-6}
 \multicolumn{1}{l}{{\it Overall for existing datasets:}} & {\it 79 (158)} & {\it 2,294} / {\it 2,572} & {\it 0} & & {\it 5} $\rightarrow$ {\it 814} & \\
 \midrule
 NIJ-2018-DU-BX-0215 & 259 (518) & 5,770 / 4,643 & 1 (both eyes) & IriShield M2120U & 0.5 $\rightarrow$ 1,674 & PMI \\
 (offered with this paper) & & & & OmniVision OV8865 & & gender \\
 & & & & (Microsoft Surface) & & age\\
 \bottomrule
 \end{tabular}
 \endgroup
 \end{table*}

\subsection{Collection Site and Equipment}

The dataset offered with this paper was collected in the Dutchess County Medical Examiner's Office (DCMEO) in New York State. Two sensors were used: (a) IriTech Irishield iris recognition camera \cite{IriShield}, capturing ISO-compliant \cite{ISO_19794_6_2011} iris images in near-infrared spectrum, and (b) the OmniVision OV8865 sensor, embedded into a Microsoft tablet, capturing regular images in visible-light. Both are handheld sensors and were used by a technician \added{during} \removed{mimicking} a routine investigation. \added{Each capture session took up to 15 minutes. On occasion iris scans did not capture and this typically happened at longer postmortem time intervals, and when cornea became cloudy despite re-hydration attempts. All scans were done indoors under similar lighting and environmental conditions.}

\subsection{Data statistics}

\begin{figure}[hbt!] 
    \centering
    \includegraphics[width=\linewidth]{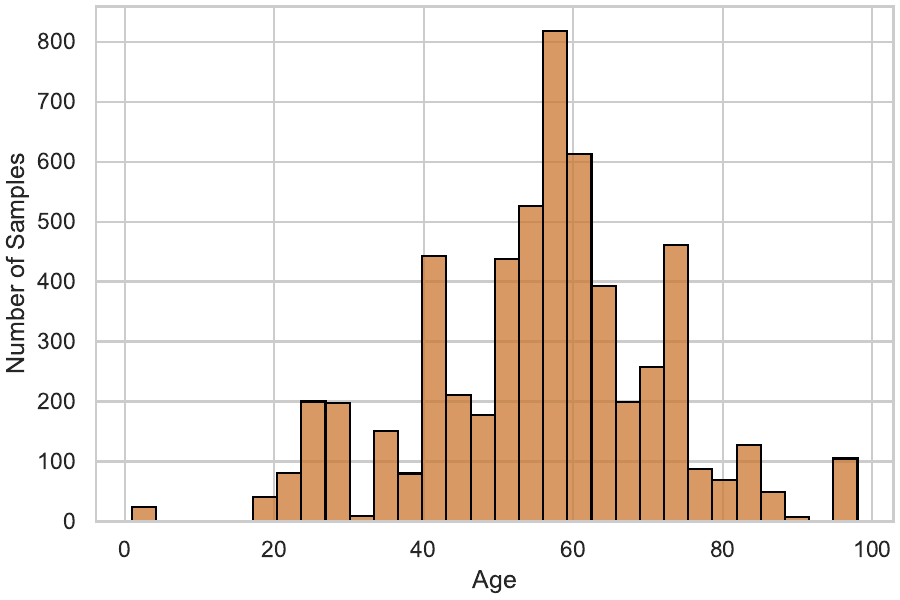}
    \caption{\added{Number of images as a function of subject's age in the \datasetname~dataset.}}
    \label{fig:age-dist}
\end{figure}

\begin{figure}[hbt!] 
    \centering
    \includegraphics[width=\linewidth]{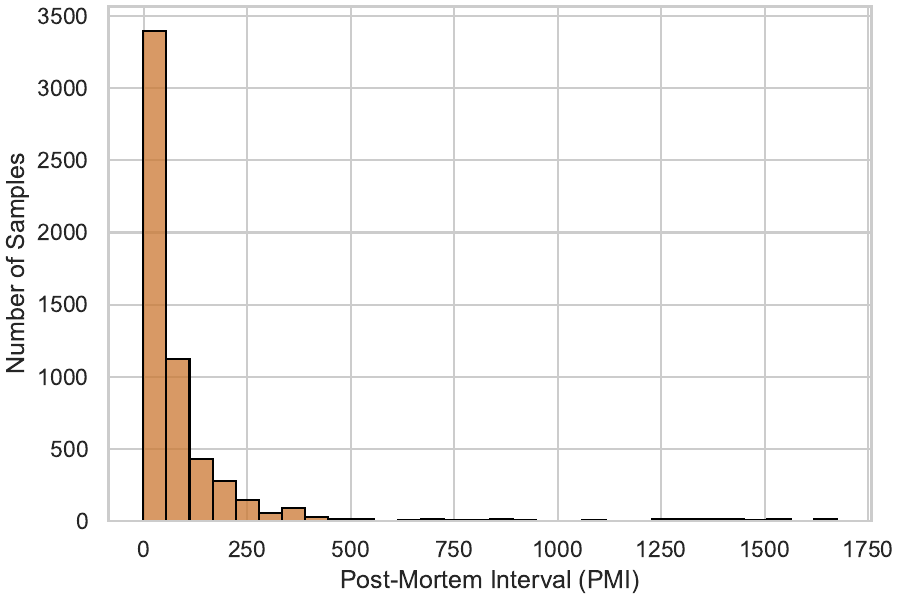}
    \caption{\added{Number of images as a function of the PMI in the \datasetname~dataset.}}
    \label{fig:PMI-dist}
\end{figure}

\begin{figure*}[hbt!]
    \includegraphics[width=\linewidth]{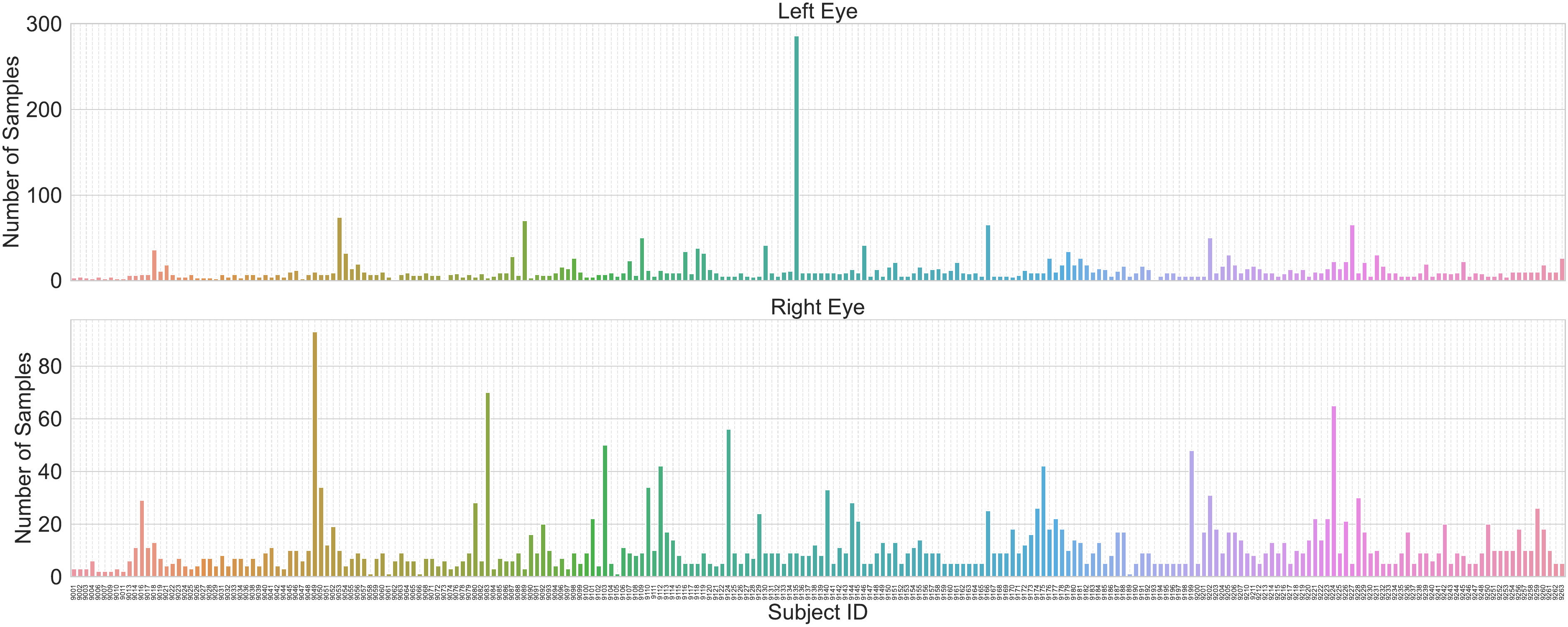}
    \caption{\added{Number of samples for each subject in all acquisition sessions in the \datasetname~dataset, shown separately for the left and right eyes.}}
    \label{fig:eye-dist}
\end{figure*}

The data was collected from 259 deceased subjects in 53 acquisition sessions. Sessions occurred approximately 12 hours apart. Both left and right irises of each subject were photographed, if possible. \added{To adhere to privacy-related regulations implemented at DCMEO, all samples were anonymized before the data left DCMEO premises.} After careful curation (described in the next subsection), this new dataset consists of 5,770 near-infrared and 4,643 visible-light images. \added{The number of images as a function of subject's age and PMI is shown in Fig. \ref{fig:age-dist} and \ref{fig:PMI-dist}, respectively. Fig. \ref{fig:eye-dist} shows the exact number of samples collected for each subject and each eye.}

\subsection{Dataset Curation}
\label{sec:DataCuration}

The NIR images were lightly curated in that images that did not contain iris texture at all (\eg due to closed eyes, or out-of-frame capture) were removed. The rotation of the images was not corrected.

The visible-light (RGB) images observed more substantial curation due to less-controlled (in a biometric sense) acquisition. In many cases, the RGB raw images contained part of the face of the deceased subjects, and often the camera positioning caused the line connecting eye canthi to deviate from horizontal. Thus, the RGB images were first rotated to put eye canthi on a horizontal line, and then such rotated images were cropped and (if needed) scaled to $640 \times 480$ pixel resolution, making sure that the margins between the iris outer boundary and the image frames are as those recommended by ISO/IEC 29794-6. The published dataset contains both raw and curated RGB samples.

\subsection{Unique Properties of NIJ-2018-DU-BX-0215}

\begin{figure*}[!htb] 
    \includegraphics[width=\linewidth]{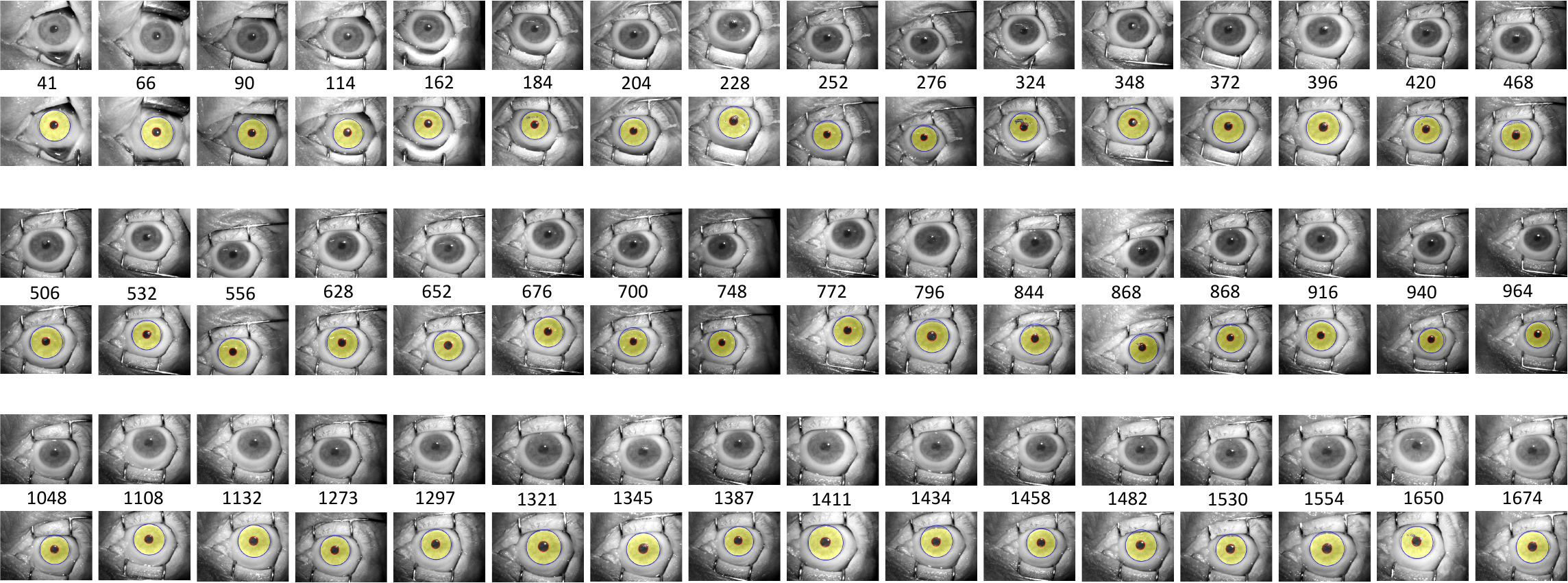}
    \caption{\added{Visualization of the progression of post-mortem deformations in time for the case with the longest PMI (1,674 hours) in the \datasetname~dataset. Corresponding pictures in lower rows illustrate the iris segmentation results. Numbers below pictures indicate the PMI for a given sample.}}
    \label{fig:progression} 
\end{figure*}

This dataset is unique in several respects. Table~\ref{tab:datasets} compares key differences between this new set and all post-mortem iris datasets published to date. The key novel components are:

\begin{itemize}[leftmargin=.12in]

\item the maximum PMI observed in this data is 1,674h (almost 70 days) for one case (\added{Fig.~\ref{fig:progression} illustrates this case by showing images from all different-day acquisitions}),

\item one case with peri-mortem and post-mortem images,

\item one iris photographed when the eye globe was completely dislocated posteriorly out of the bony eye socket,

\item one case with massive destruction to the head following a gunshot (rifle) wound, which produced fracturing of the bony orbits; the globes were however intact and irises were able to be scanned (and even matched by the IriTech algorithm). 
\end{itemize}

\section{Experimental Methodology}

\subsection{Dataset Composition}
\label{sec:DatasetComposition}

To authoritatively assess the post-mortem iris recognition performance, robustness of detection post-mortem samples (simulating a presentation attack detection scenario), as well as its dependency on the selected demographic factors, we have combined the newly-collected \datasetname~with all other  available forensic iris datasets (Warsaw BioBase Post Mortem Iris v2.0  \cite{trokielewicz2018iris} and Warsaw BioBase Post Mortem Iris v3.0  \cite{trokielewicz2020post}, both briefly characterized in Sec. \ref{sec:related:datasets}), to create the largest-possible corpus of forensic iris samples. 

In this study, we only use NIR iris images from all sets. After combining these datasets, the total number of subjects and the number of iris images were 338 and 8,064, respectively. All datasets offer metadata, such as gender, age, and PMI, and Warsaw subsets also offer the information about cause of death. From further analyses presented in this paper, we have removed subjects and corresponding data with missing metadata (\eg unknown PMI), ending up with the total number of subjects and images to be 329 and 7,993, respectively. To the best of our knowledge, this is the largest publicly-available forensic iris recognition dataset available at the time of writing.

\subsection{Data Balancing for Demographic Study}

While PMI is an obvious factor affecting post-mortem iris recognition, age of the decedent subject can also be correlated with the speed of tissue degradation. Thus, to make a fair assessment whether there are any differences in the iris recognition performance between gender and age, %
we balanced both the average PMI and the age of subjects. 

To do this, we first split the metadata into demographic groups to balance the PMI, sorted each subgroup's metadata in terms of PMI in ascending order, and kept removing samples from a group with a larger average PMI to make the average PMI equal. Owing to a diverse subject pool, this process also left us with a good balance of the age between male and female groups. Table \ref{tab:age-gender-data-stats} summarizes the PMI and age statistics, along with the numbers of genuine and impostor scores generated for experiments, in gender- and age-related studies.

\begin{table}[!htb]
\centering
\caption{Data statistics after balancing along the subject age and gender dimensions.}
\label{tab:age-gender-data-stats}
\begin{tabular}{cccccc}
\toprule
 & & \textbf{Age (years)} & & \multicolumn{2}{c}{\textbf{Gender}}\\ 
 & \textbf{1-33} & \textbf{34-66} & \textbf{67-99} & \textbf{Male} & \textbf{Female} \\ 
\midrule
Total Subjects & 37 & 210 & 81 & 238 & 91 \\ 
Impostor Pairs & 8,450 & 257,319 & 29,443 & 455,486 & 52,691\\ 
Genuine Pairs & 35,88 & 29,198 & 8,743 & 57,687 & 21,208 \\
Average PMI & 50.7 & 50.7 & 50.7 & 85.3 & 85.3 \\ 
Average Age &  26.15    &  53.53    &   75.56   & 55.8 & 57.7 \\
\bottomrule
\end{tabular}
\end{table}

\subsection{Post-mortem Iris Detection}
\label{sec:PostmortemIrisDetection}

In addition to the desired capabilities of post-mortem iris recognition, there is a need to detect iris images captured after demise as ``spoofs", to prevent using post-mortem cases in biometric presentation attacks. 

This problem, studied for the first time by Trokielewicz \etal \cite{trokielewicz2018presentation}, is revisited in this paper with (a) comprehensive experiments on a larger dataset than in \cite{trokielewicz2018presentation}, and (b) application of a state-of-the-art deep learning-based iris presentation attack detection model (D-NetPAD~\cite{sharma2020d}), demonstrating effectiveness of few-shot learning in post-mortem iris case detection. The choice of D-NetPAD for this study of post-mortem iris detection stems from its proven efficacy in detecting a range of presentation attacks, including those involving paper printouts, cosmetic contact lenses, or artificial eyes, and demonstrated by its success in the LivDet-Iris 2020 competition \cite{das2020iris}. The D-NetPAD model consists of 121 convolutional layers in a series of four dense blocks and three transition layers. It accepts a cropped and resized iris region and produces a presentation attack (PA) score between 0.0 (corresponding to a {\it bona fide} sample) and 1.0 (denoting a presentation attack).

We have designed and conducted eight distinct experiments of detecting presentation attacks with post-mortem iris images to demonstrate the feasibility of deep learning-based cadaver eye detection across various scenarios. These experiments utilized the newly collected \datasetname~and various combinations of the Warsaw datasets (described in Sec.~\ref{sec:DatasetComposition}). Given the necessity for cropped iris inputs in D-NetPAD, we segmented the iris images using VeriEye~\cite{verieye}.

In the first phase of our study, we explored the effectiveness of cadaver eye detection offered by the D-NetPAD model pre-trained on ImageNet dataset. \textbf{Experiment 1} involved testing the model with a blend of \added{256} live images from the \added{PAD-for-Cadaver-Iris-Live-Portion (PAD-CILP)} \cite{Trokielewicz_BTAS_2018} and 2,294 images from the combined Warsaw (v1.0, v2.0, and v3.0) datasets, while \textbf{Experiment 2} utilized the same live iris images alongside 5,770 images from the \datasetname~dataset. \added{We used live samples originating only from the PAD-CILP dataset since it was collected with the same sensor (IriTech IriShield) as the Warsaw and the \datasetname~datasets. This is to avoid bias and prevent the model from relying on sensor differences rather than biological eye features during training and testing}.

Subsequently, in \textbf{Experiments 3 and 4}, we fine-tuned the D-NetPAD model with post-mortem images \added{alongside the same set of 256 live images to enhance its detection capabilities. \textbf{Experiment 3} involved fine-tuning the model with 2,294 combined post-mortem images from the {Warsaw} dataset and 80\% of the live images (201 samples from 15 identities), and testing it on the full set of post-mortem images from {\datasetname} and the remaining 55 live images from 4 identities to evaluate cross-dataset performance. \textbf{Experiment 4} followed the reverse setup: the model was trained on post-mortem images from {\datasetname} along with the same 80\% of live images (201 samples), and tested on post-mortem samples from {Warsaw} and the remaining 55 live images to assess generalization in the opposite direction.}

\added{In \textbf{Experiments 5 and 6}, the D-NetPAD model was fine-tuned with 50 images from the {Warsaw} and {\datasetname}, respectively, and tested on the opposite dataset to evaluate its ability to detect presentation attacks in a cross-dataset scenario, assuming limited availability of forensic samples. \textbf{Experiment 5} involved fine-tuning the model with 50 post-mortem images from the {Warsaw} dataset along with 50 live images (excluding 5 live samples belonging to one identity), and testing it on the full {\datasetname} set consisting of 5,770 images, as well as the remaining 201 live images from 15 unique identities. \textbf{Experiment 6} followed the reverse setup: the model was fine-tuned using 50 post-mortem images from {\datasetname} and the same 50 live images, and tested on the full {Warsaw} dataset and the remaining 201 live images to evaluate generalization under limited data conditions.}

\added{Our final \textbf{Experiments, 7 and 8}, focused on the model’s robustness when fine-tuned with a minimal set of post-mortem images. \textbf{Experiment 7} used only 5 post-mortem images from the combined {Warsaw} dataset and 5 live images for fine-tuning and tested on {\datasetname}, while \textbf{Experiment 8} followed the same setup but used 5 post-mortem images from {\datasetname} instead and tested on {Warsaw}. In both cases, the live test set included the remaining 201 live images from 15 unique identities to evaluate performance under extremely limited training data conditions.}

\subsection{Iris Recognition Methods}
\label{sec:IrisRecognitionMethods}

We have employed five iris recognition methods representing five different families of approaches: (1) a method that approximates Daugman's Gabor wavelet-based approach (OSIRIS), (2) an additional iris code-based matcher employing non-Gabor kernels (USIT) (3) an iris code-based method employing kernels sourced from human perception studies (HDBIF), (4) a deep learning-based method (DGR), and (5) a commercial approach (VeriEye). All methods are briefly described in the following paragraphs.

\paragraph{OSIRIS \cite{othman2016osiris}} Open Source for IRIS v4.1 is an open-source academic solution developed under the BioSecure EU project. As in Daugman's original approach \cite{Daugman_PAMI_1993} the phase quantization of the Gabor filter outcomes has been utilized to calculate the iris code. A fractional Hamming distance is then used to calculate a comparison score between iris codes.

\paragraph{USIT \cite{USIT3}} The University of Salzburg Iris Toolkit v3.0 is an open-source iris recognition method implementing several iris image pre-processing, segmentation, feature extraction, and comparison algorithms, possible to be mixed-and-matched to experiment with optimal combinations of approaches for a given scenario.  
In this study we have followed the recommendations of the USIT authors on the optimal combination of methods and applied the Contrast-adjusted Hough Transform \texttt{CAHT} \cite{rathgeb2013iris} for iris segmentation, a 1D Log-Gabor-based encoding \texttt{LG} \cite{masek2003recognition}, and the \texttt{TripleA} algorithm to compare iris codes.

\paragraph{HDBIF \cite{czajka2019domain}} Human-Driven Binarized Image Feature open-source method is %
one of a few iris recognition algorithms designed with post-mortem iris recognition taken into account. HDBIF combines deep learning-based iris segmentation~\cite{trokielewicz2020post} and human perception-sourced iris-domain-specific feature extraction. To find filtering kernels, first an eye tracking-based experiments were concluded to locate salient features (iris image patches) used by humans when solving the iris recognition task. Next, Independent Component Analysis was applied to the localized patches to find convolution kernels that maximize the statistical independence among the filtering results, following \cite{kannala2012bsif}. 
Finally, the comparison score is calculated using fractional Hamming distance between non-occluded iris portions. A version of the IREX X-ready \cite{IREX_X_URL} implementation of the HDBIF matcher is open-sourced \cite{HDBIF_GitHub}.

\paragraph{DGR \cite{ren2020dynamic}} Dynamic Graph Representation uses a hybrid framework of cascaded Convolutional Neural Networks (CNNs) and graph models to learn dynamic graph representations of iris features. With the help of a graph generator that establishes connections among image parts, the convolutional features are extracte and used to generate feature graphs based on node representations. Each node of the graph contains a feature vector, and the weights of edges express the relationships between nodes. Then, the feature graph is sent to a hierarchical feature extractor for the graph, known as SE-GAT, a novel structure based on Graph Attention Networks (GAT) \cite{velivckovic2017graph}. The loss function introduced by the DGR authors combines cross-entropy and triplet loss functions, which provides a new measure of calculating similarity between two graphs. Owing to the use of a graph-based feature representations, the proposed algorithm does not perform any iris image segmentation and instead makes an attempt to learn the locations of salient iris features within the model itself.

\paragraph{VeriEye \cite{verieye}} Neurotechnology's VeriEye matcher is a commercial solution implementing an unpublished algorithm. VeriEye was evaluated by NIST in two of their programs, ICE \cite{phillips2008iris} and IREX \cite{irex_general}, always being placed at top positions in the leaderboard. VeriEye calculates a similarity between iris images, that is, the higher the score, the better the match between samples. The similarity measure is recommended to be at least 40 to judge two iris images as representing the same eye.

\subsection{Biological and Demographic Factors Impacting Post-mortem Iris Recognition}

Post-mortem iris recognition is a new field that applies iris recognition technology in a forensic setting, specifically in situations when deceased subjects need to be identified. Such situations  are far less controlled than typical research laboratory or consumer setups in which good subject positioning and lighting can be ensured. Thus, many factors beyond the PMI may affect post-mortem iris recognition, such as ambient conditions, studied by Sauerwein \etal \cite{sauerwein2017effect} and Bolme \etal \cite{bolme2016impact}, who concluded that irises, when bodies are kept outdoors, are unusable after a few days in warmer temperatures, but longer in the winter time.

There were no, however, previous studies reporting potential differences in post-mortem recognition performance across selected demographic factors, such as subject's age and gender. Gorodnichy and Chumakov \cite{gorodnichy2019analysis} showed that iris recognition performs worse for specific age groups and the best performance is achieved for subjects between 30 and 60 years old. Quinn \etal \cite{quinn2018irex} performed an  analysis of a potential impact of gender on iris recognition as part of the IREX IX program, and concluded that the recognition performance varied inconsistently between male and female subjects. It is unknown, however, if the detected correlations between the gender group and the performance is causal, since iris texture itself was found to be not dependent on genotype features \cite{Daugman_TCSVT_2004} and there may exist other, non-biological reasons to observe these correlations, such as presence of cosmetics \cite{Kuehlkamp_WACV_2017}. Thus, in this study, which focuses solely on post-mortem iris recognition, we are verifying a hypothesis that there are possible differences in the performance between gender groups, and among the groups gathering subjects of similar age at the time of death.

\section{Experimental Results}

\subsection{General Performance of Post-mortem Iris Recognition}
\label{subsec:general-performance}

After \datasetname~dataset curation (see Sec.~\ref{sec:DatasetComposition}), we have performed all possible genuine and impostor pair matching for \datasetname~ and Warsaw datasets separately, what yielded 23,836 genuine and 2,606,235 impostor pairs for the combined Warsaw datasets, and 98,612 genuine and 29,732,970 impostor pairs for the \datasetname~corpus. Hence, this study is based on 122,448 genuine and 32,339,205 impostor comparison scores.

\begin{figure*}[hbt!]
    \centering
       \subfloat[]{\includegraphics[width=0.20\linewidth]{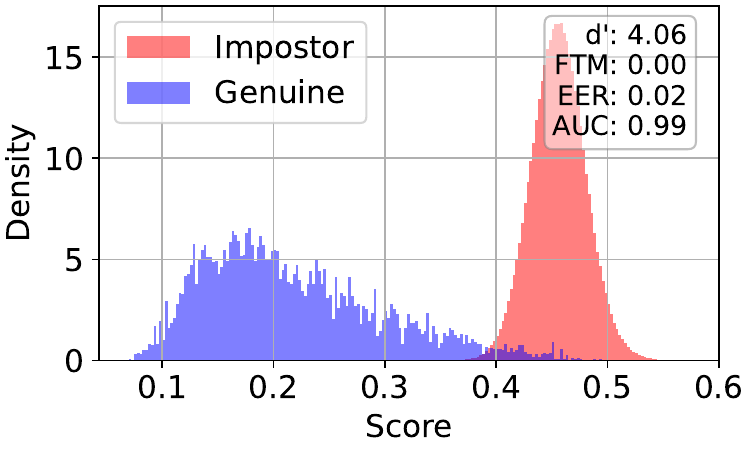}}
       \subfloat[]{\includegraphics[width=0.20\linewidth]{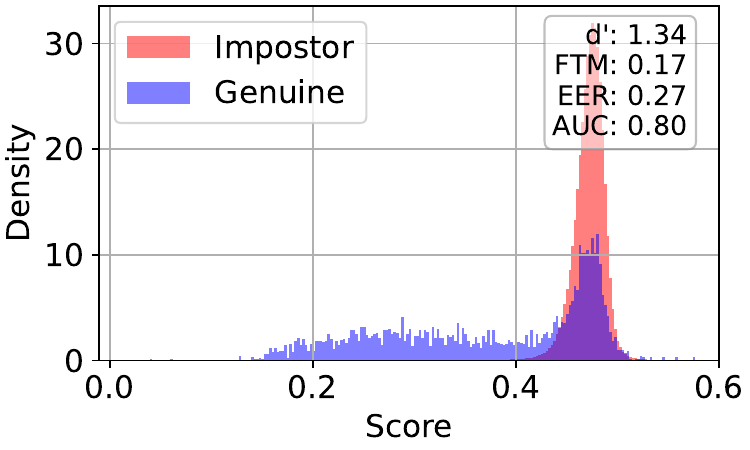}}
       \subfloat[]{\includegraphics[width=0.20\linewidth]{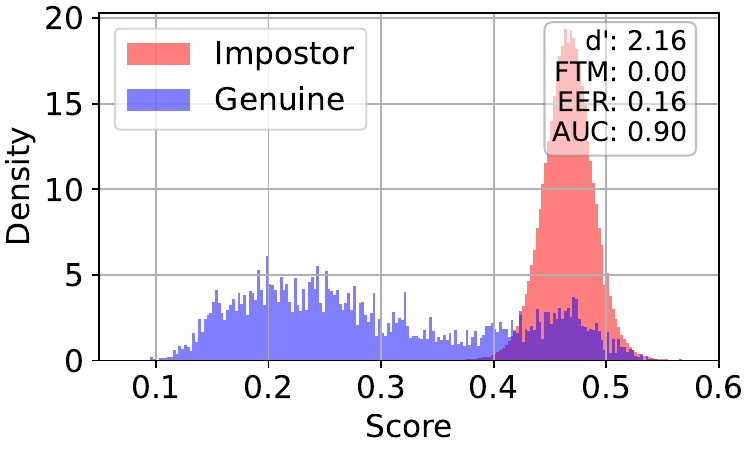}}
       \subfloat[]{\includegraphics[width=0.20\linewidth]{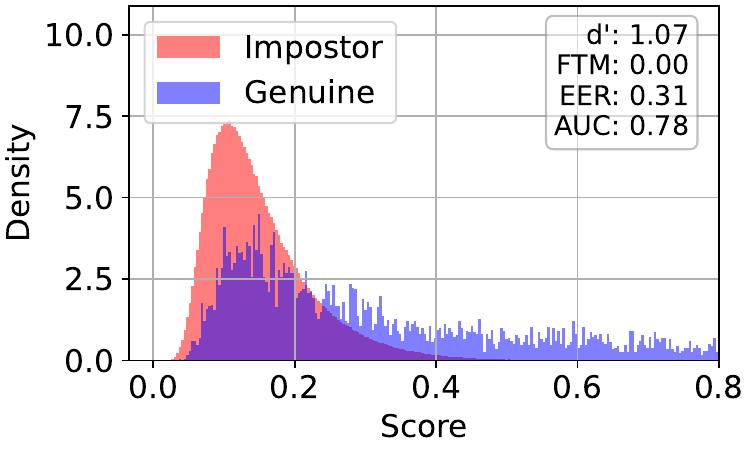}}
       \subfloat[]{\includegraphics[width=0.20\linewidth]{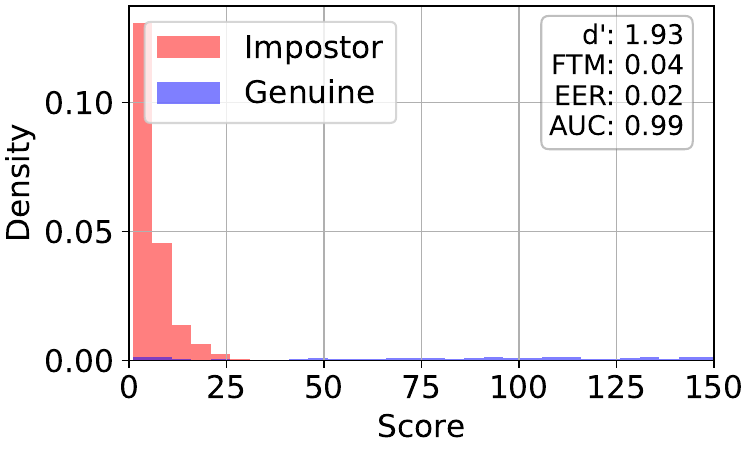}}
       \vskip-6mm
       \subfloat[]{\includegraphics[width=0.20\linewidth]{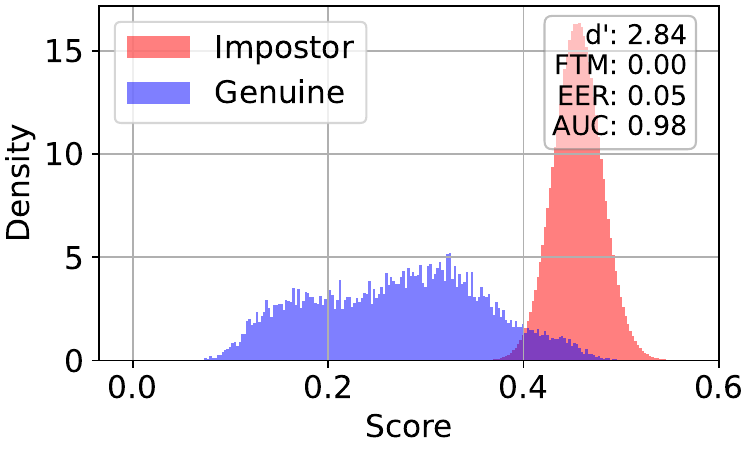}}
       \subfloat[]{\includegraphics[width=0.20\linewidth]{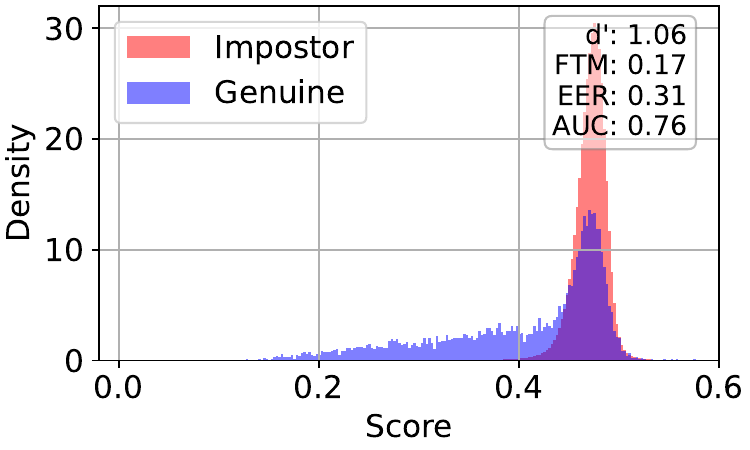}}
       \subfloat[]{\includegraphics[width=0.20\linewidth]{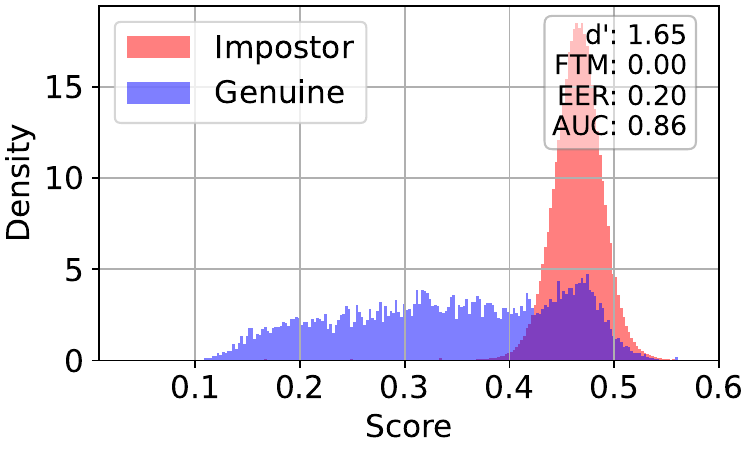}}
       \subfloat[]{\includegraphics[width=0.20\linewidth]{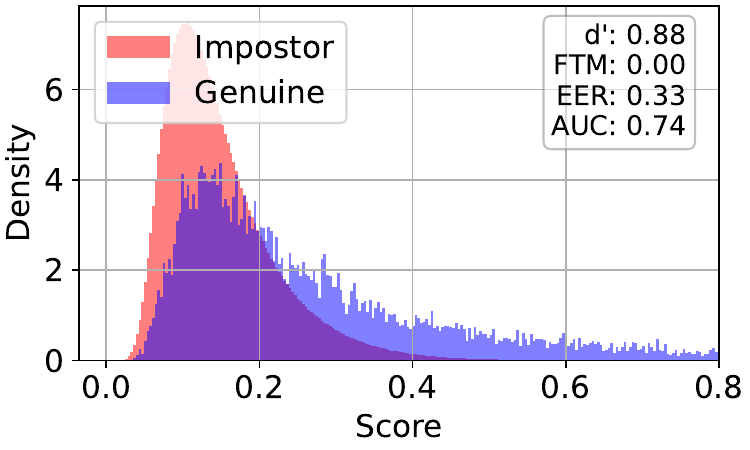}}
       \subfloat[]{\includegraphics[width=0.20\linewidth]{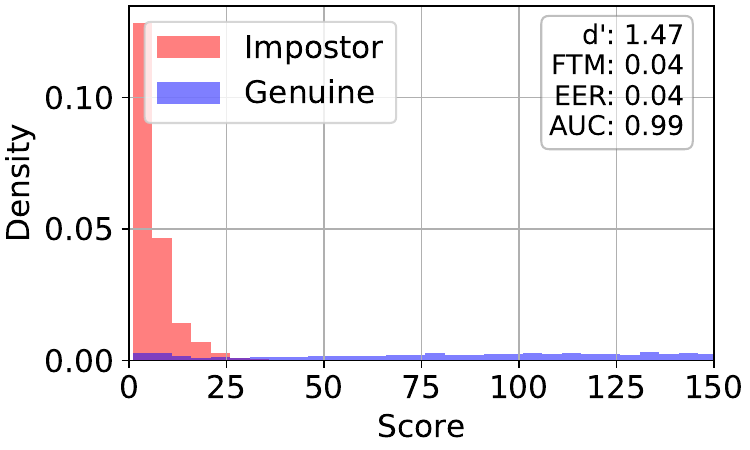}}
       \vskip-6mm
       \subfloat[]{\includegraphics[width=0.20\linewidth]{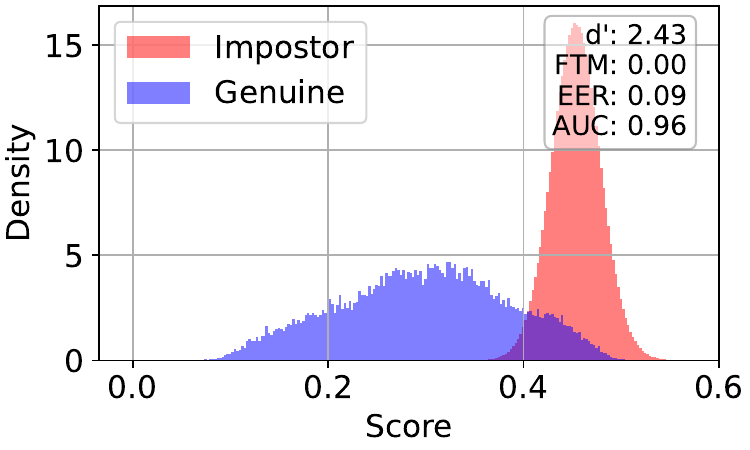}}
       \subfloat[]{\includegraphics[width=0.20\linewidth]{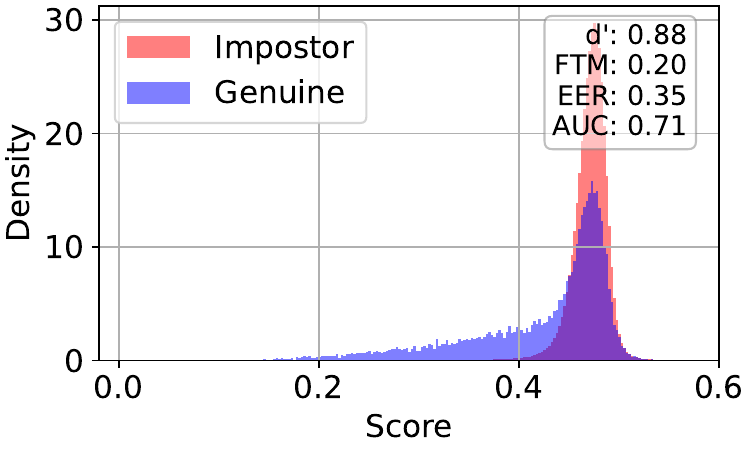}}
       \subfloat[]{\includegraphics[width=0.20\linewidth]{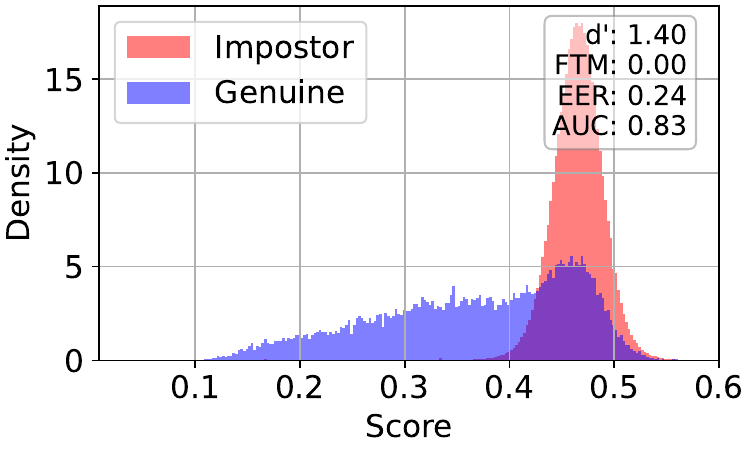}}
       \subfloat[]{\includegraphics[width=0.20\linewidth]{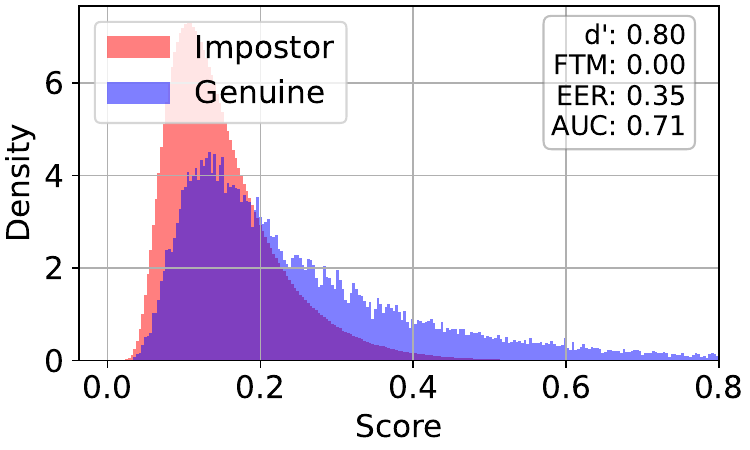}}
       \subfloat[]{\includegraphics[width=0.20\linewidth]{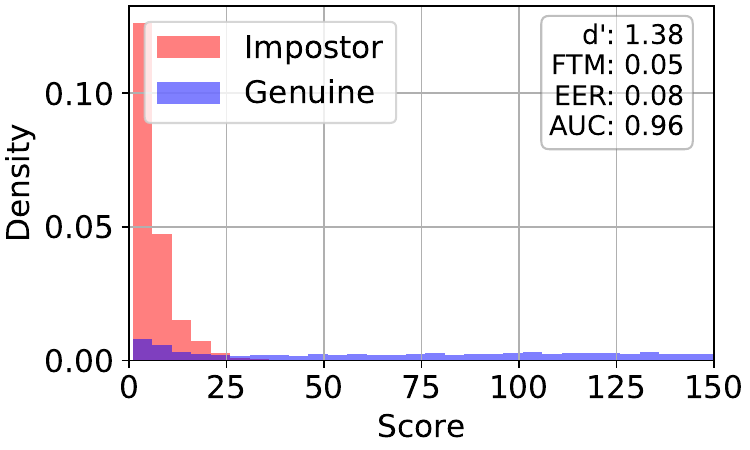}}
       \vskip-6mm
       \subfloat[HDBIF]{\includegraphics[width=0.20\linewidth]{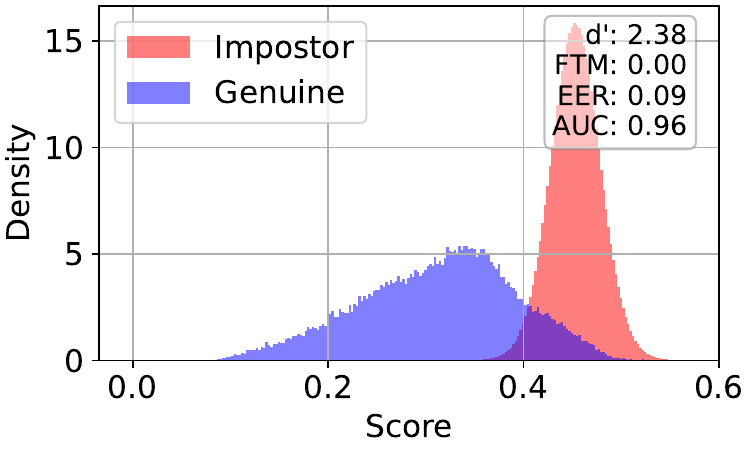}}
       \subfloat[USIT]{\includegraphics[width=0.20\linewidth]{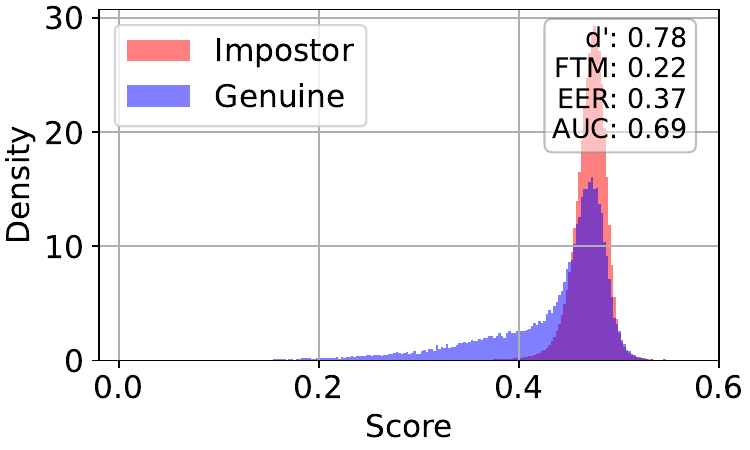}}
       \subfloat[OSIRIS]{\includegraphics[width=0.20\linewidth]{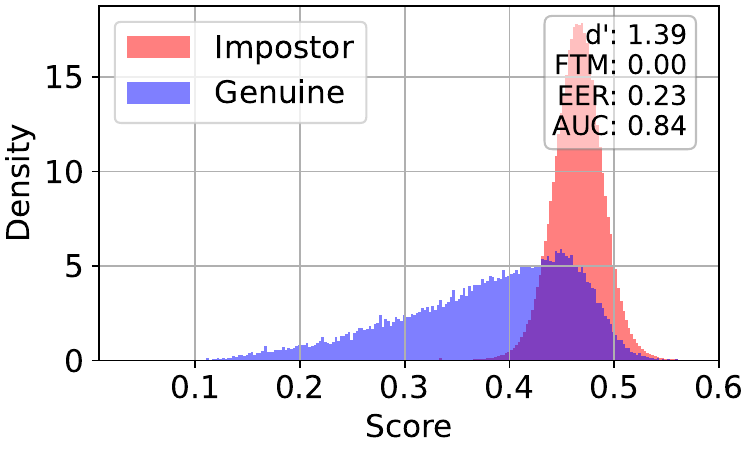}}
       \subfloat[DGR]{\includegraphics[width=0.20\linewidth]{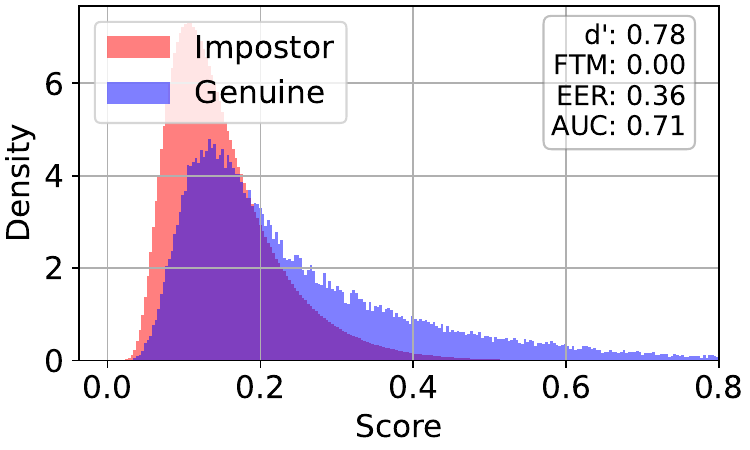}}
       \subfloat[VeriEye]{\includegraphics[width=0.20\linewidth]{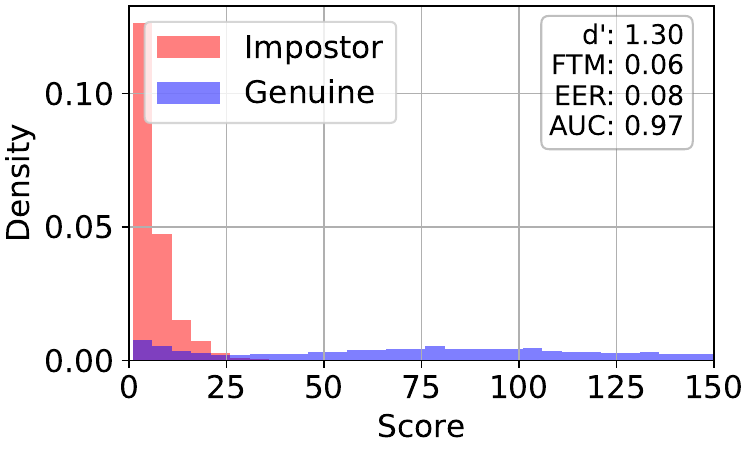}}
       
    \caption{\added{The comparison score distributions for the {\bf combined \datasetname~+~Warsaw NIR datasets}, obtained for five different iris matching algorithms. The main performance metrics (d', Failure to Match, Equal Error Rate and Area Under ROC curve) are also shown, and results are {\bf split by the PMI range}: 0-24h (top row), 0-72h (second row), 0-240h (third row), and all PMIs combined (bottom row).}}
    \label{fig:combined-score-dist} 
\end{figure*}

\begin{figure*}[hbt!]
    \centering
    \subfloat[]{\includegraphics[width=0.20\linewidth]{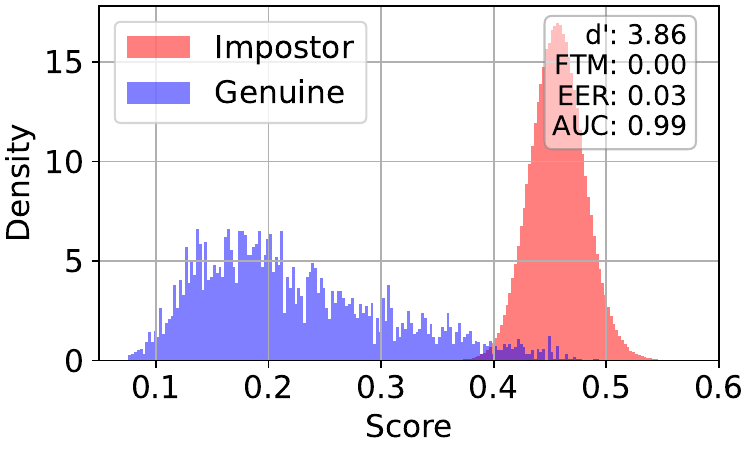}}
    \subfloat[]{\includegraphics[width=0.20\linewidth]{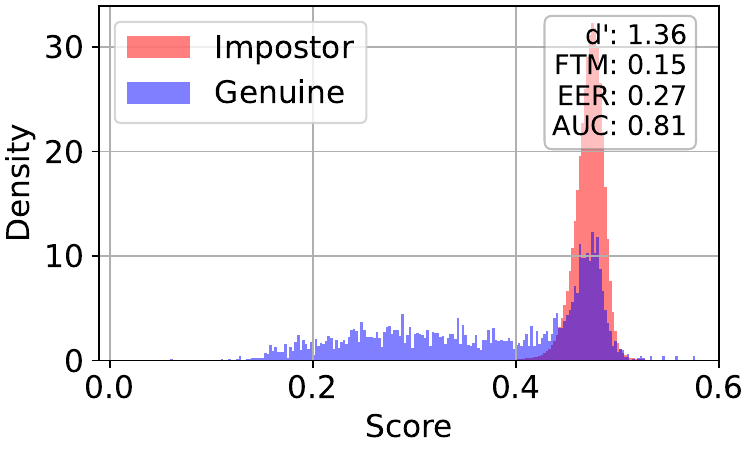}}
    \subfloat[]{\includegraphics[width=0.20\linewidth]{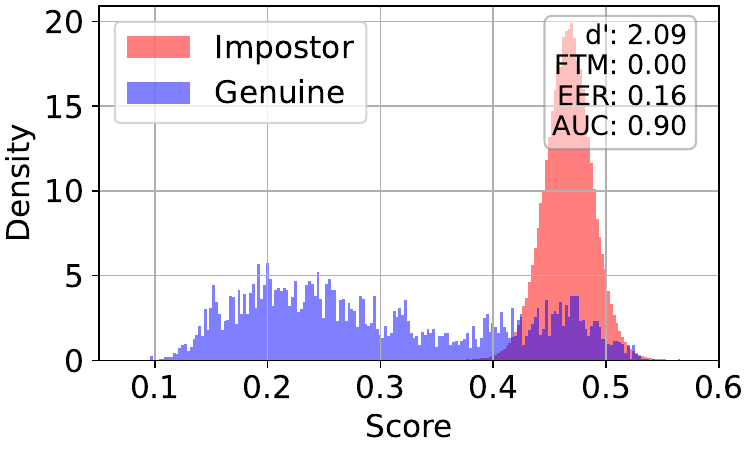}}
    \subfloat[]{\includegraphics[width=0.20\linewidth]{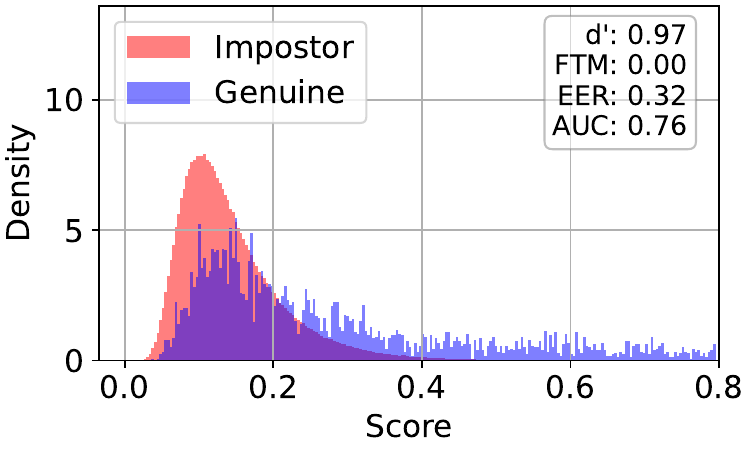}}
    \subfloat[]{\includegraphics[width=0.20\linewidth]{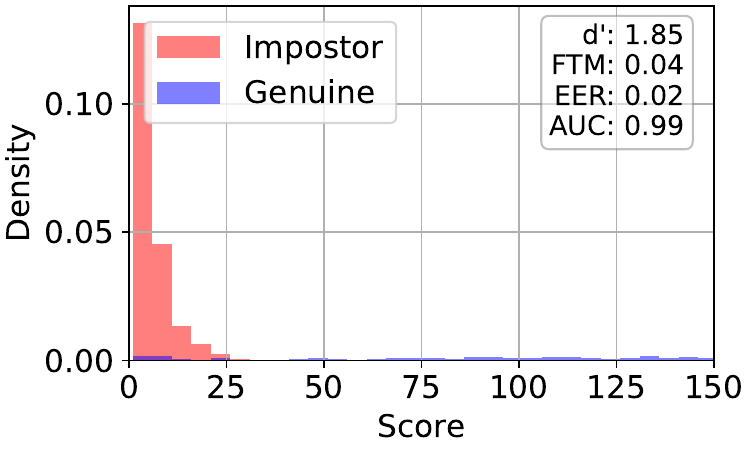}}
    \vskip-6mm
    \subfloat[]{\includegraphics[width=0.20\linewidth]{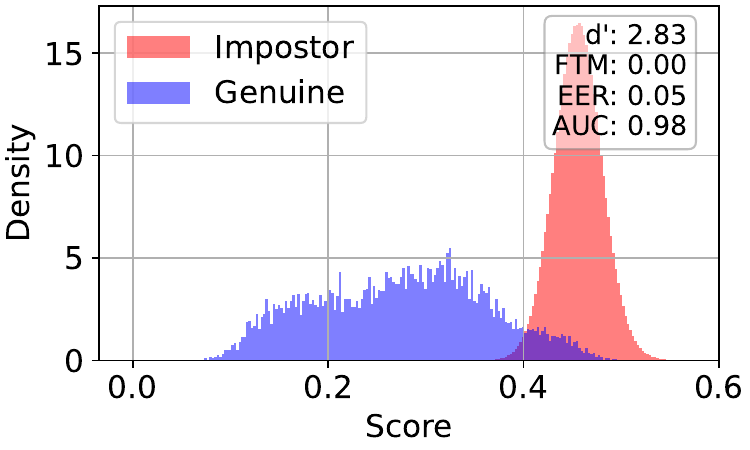}}
    \subfloat[]{\includegraphics[width=0.20\linewidth]{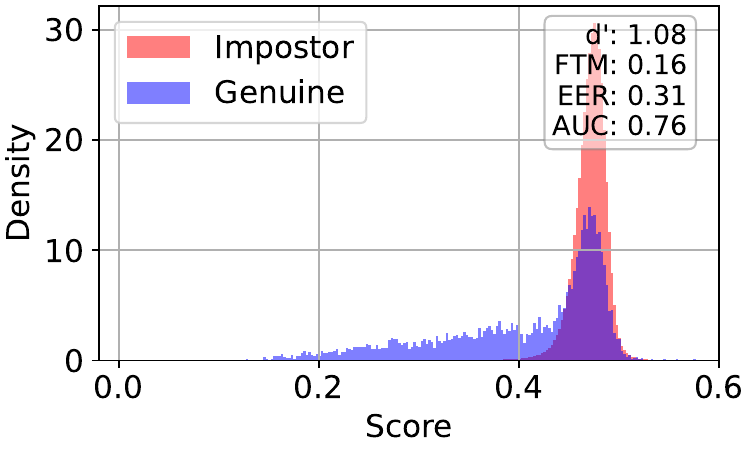}}
    \subfloat[]{\includegraphics[width=0.20\linewidth]{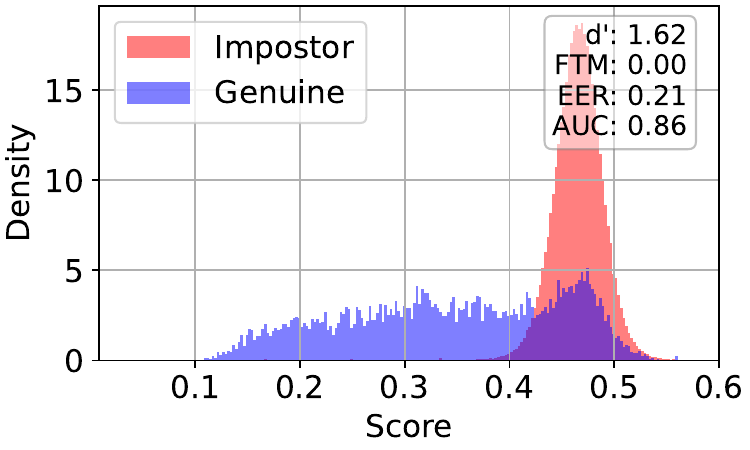}}
    \subfloat[]{\includegraphics[width=0.20\linewidth]{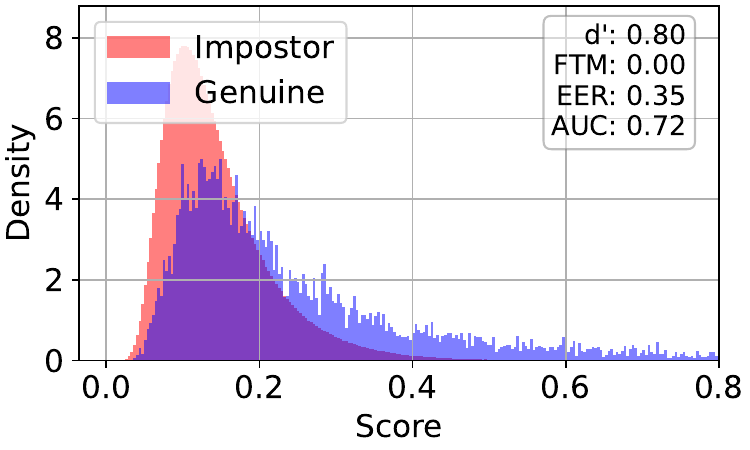}}
    \subfloat[]{\includegraphics[width=0.20\linewidth]{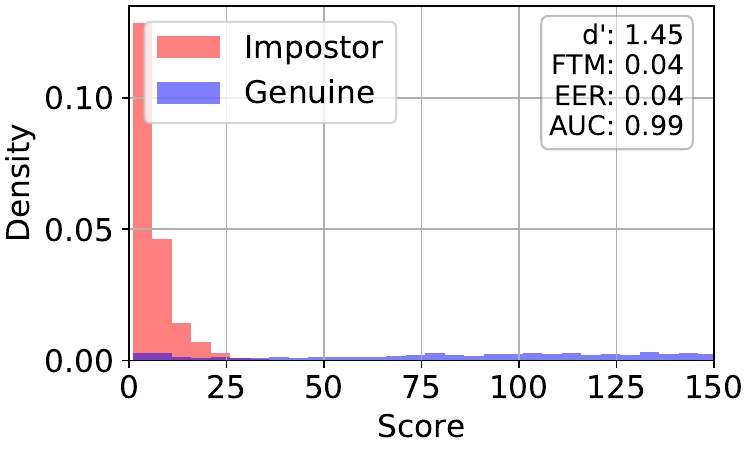}}
    \vskip-6mm
    \subfloat[]{\includegraphics[width=0.20\linewidth]{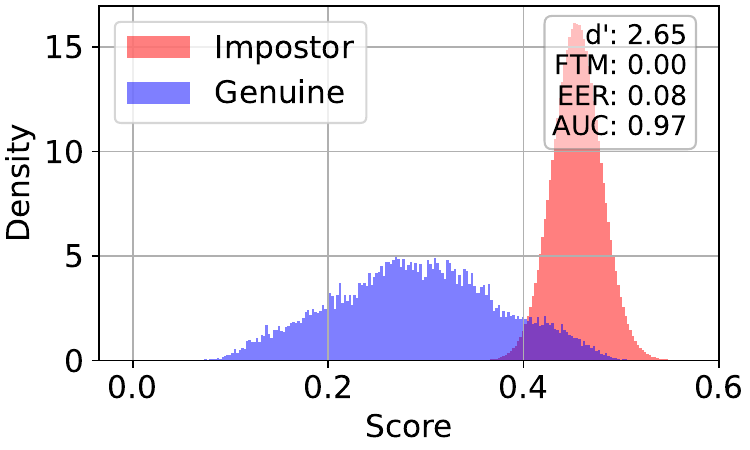}}
    \subfloat[]{\includegraphics[width=0.20\linewidth]{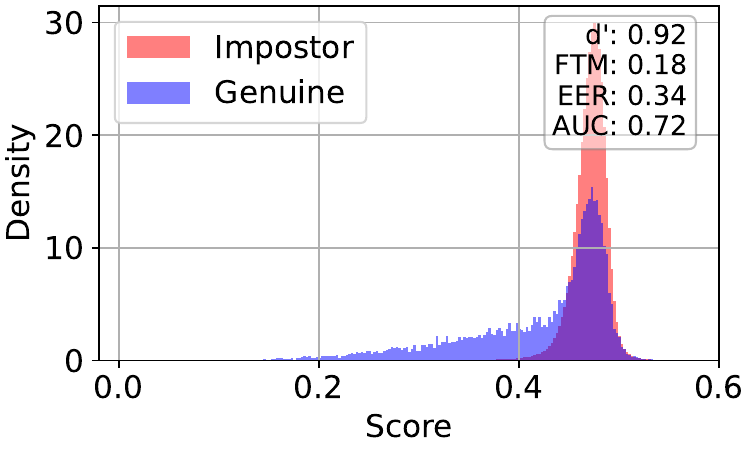}}
    \subfloat[]{\includegraphics[width=0.20\linewidth]{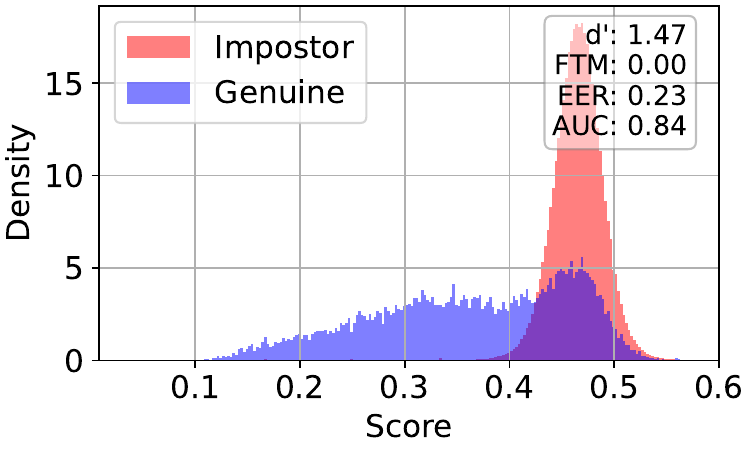}}
    \subfloat[]{\includegraphics[width=0.20\linewidth]{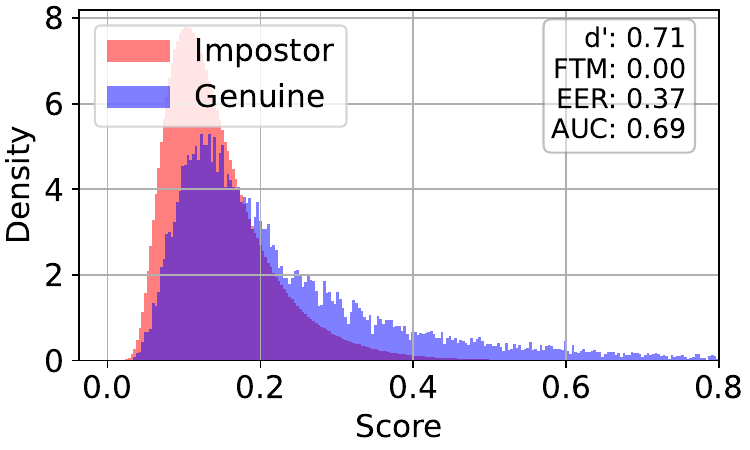}}
    \subfloat[]{\includegraphics[width=0.20\linewidth]{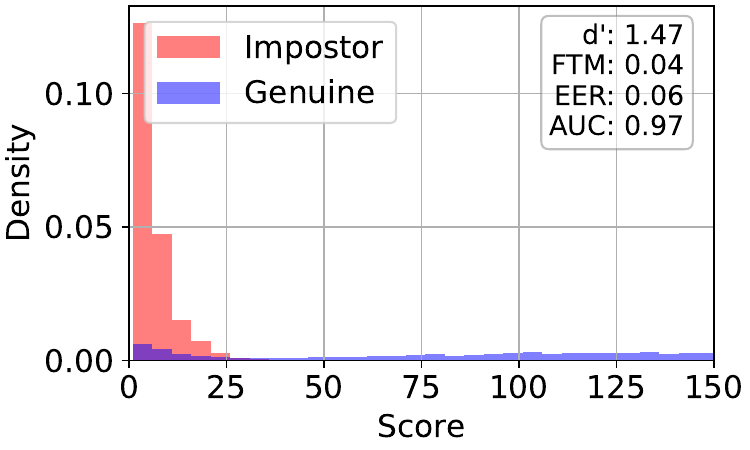}}
    \vskip-6mm
    \subfloat[HDBIF]{\includegraphics[width=0.20\linewidth]{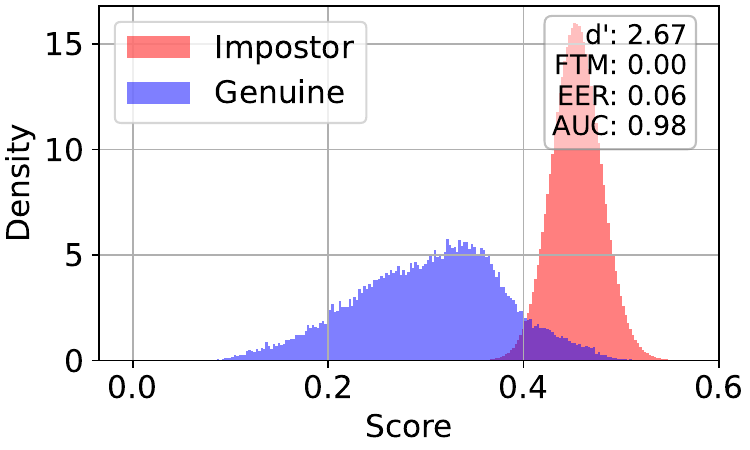}}
    \subfloat[USIT]{\includegraphics[width=0.20\linewidth]{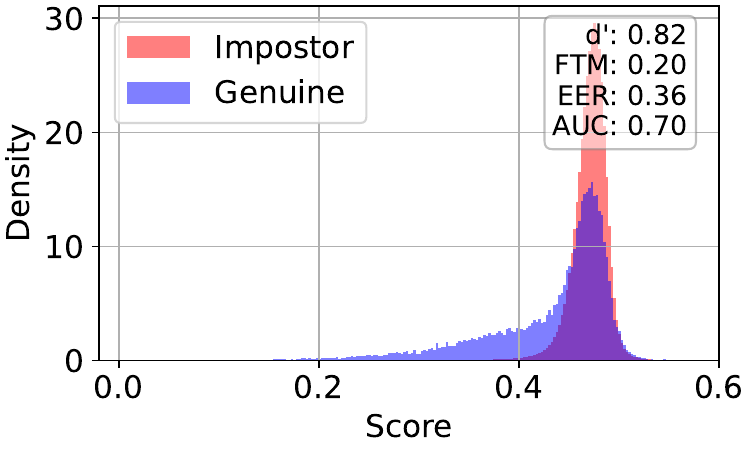}}
    \subfloat[OSIRIS]{\includegraphics[width=0.20\linewidth]{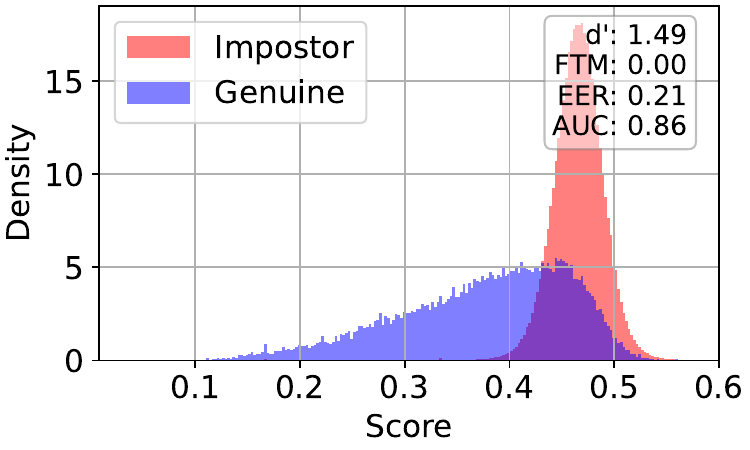}}
    \subfloat[DGR]{\includegraphics[width=0.20\linewidth]{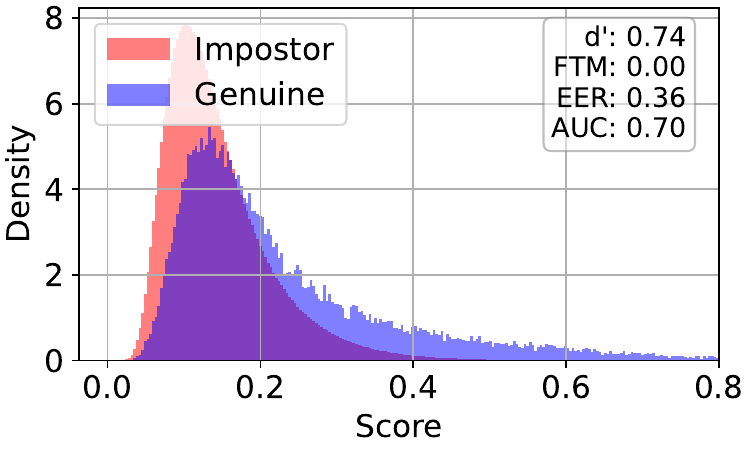}}
    \subfloat[VeriEye]{\includegraphics[width=0.20\linewidth]{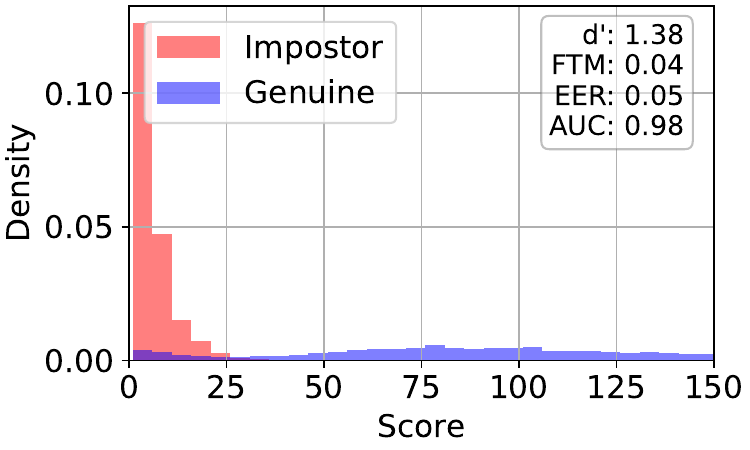}}        
  \caption{Same as in Fig. \ref{fig:combined-score-dist} in the main paper (the comparison score distributions obtained for five different iris matching algorithms) except that {\bf only scored obtained for \datasetname~NIR samples} are shown, and results are {\bf split by the PMI range}: 0-24h (top row), 0-72h (second row), 0-240h (third row), and all PMIs combined (bottom row). }
  \label{fig:dcmeo2-score-dist} 
\end{figure*}

\begin{figure*}[hbt!]
    \centering
    \subfloat[]{\includegraphics[width=0.20\linewidth]{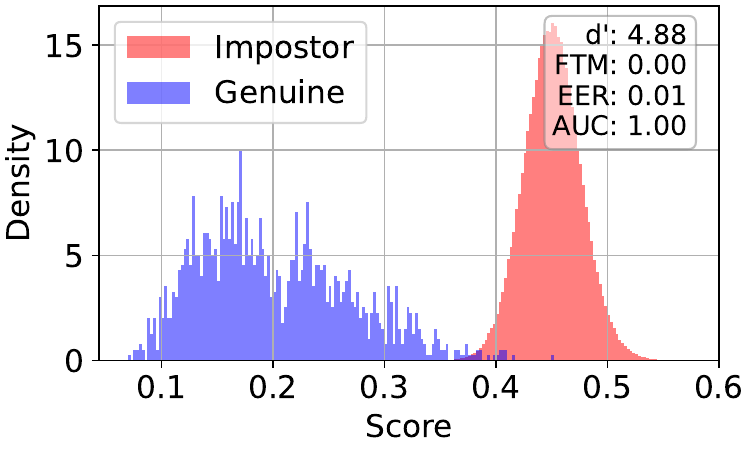}}
    \subfloat[]{\includegraphics[width=0.20\linewidth]{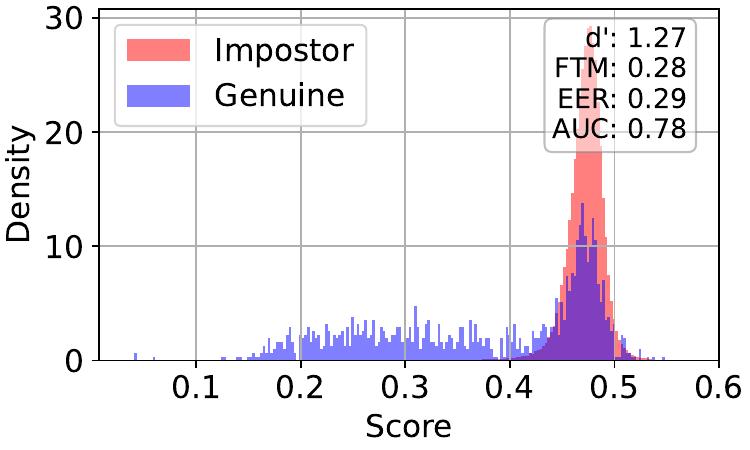}}
    \subfloat[]{\includegraphics[width=0.20\linewidth]{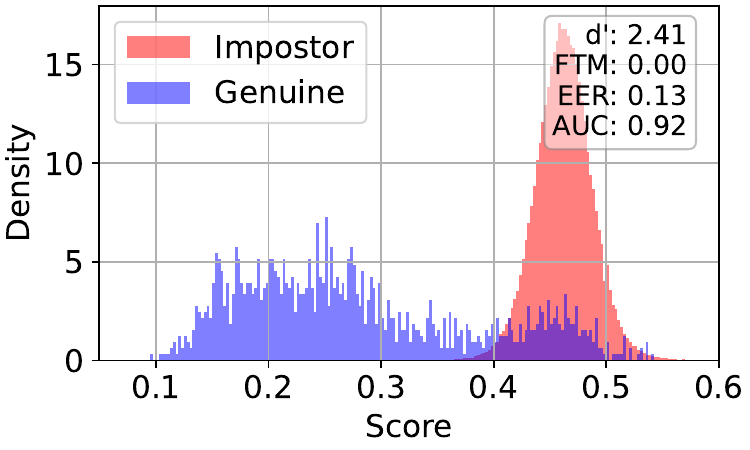}}
    \subfloat[]{\includegraphics[width=0.20\linewidth]{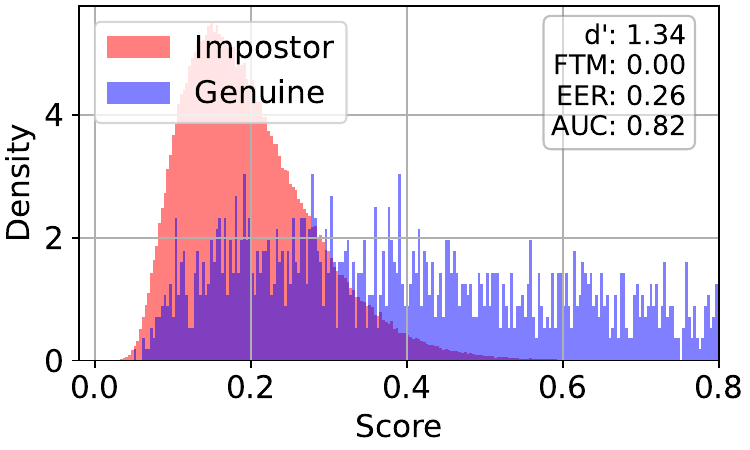}}
    \subfloat[]{\includegraphics[width=0.20\linewidth]{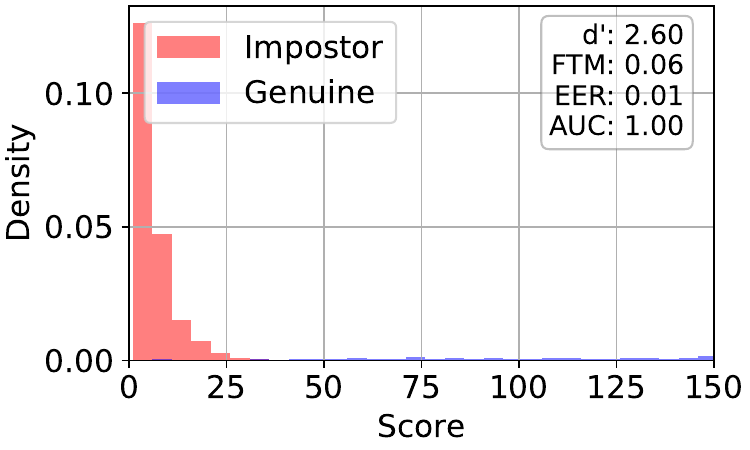}}
    \vskip-6mm
    \subfloat[]{\includegraphics[width=0.20\linewidth]{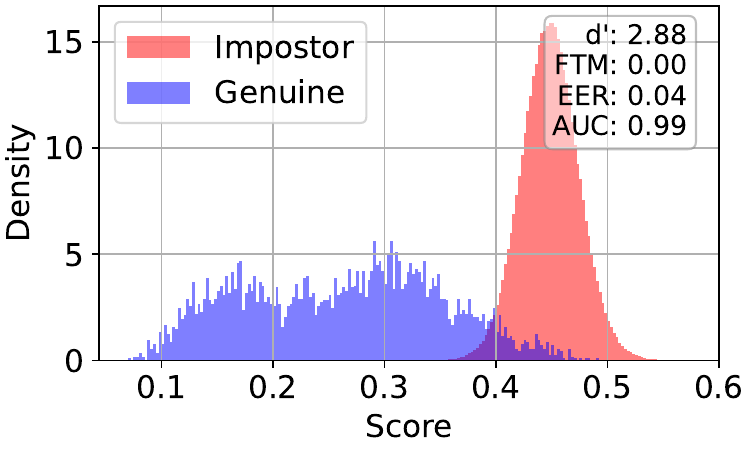}}
    \subfloat[]{\includegraphics[width=0.20\linewidth]{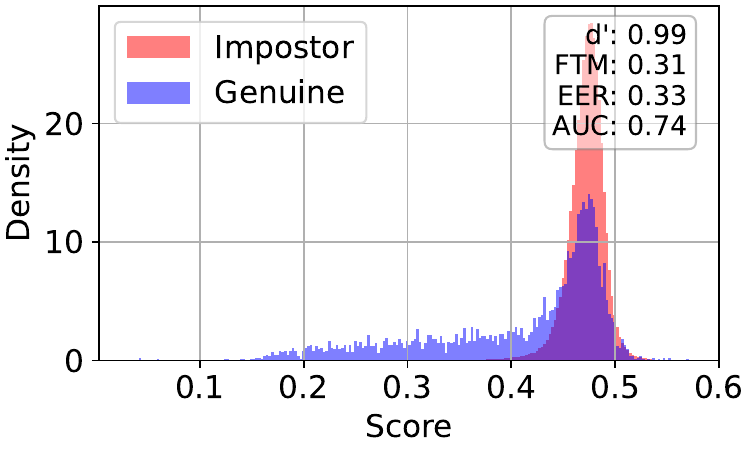}}
    \subfloat[]{\includegraphics[width=0.20\linewidth]{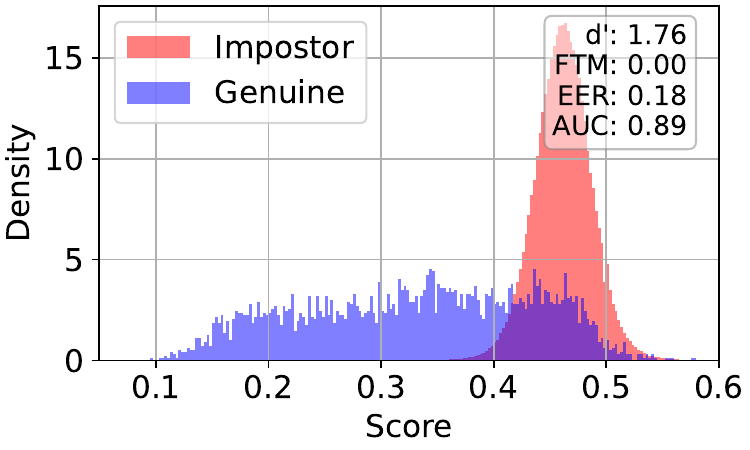}}
    \subfloat[]{\includegraphics[width=0.20\linewidth]{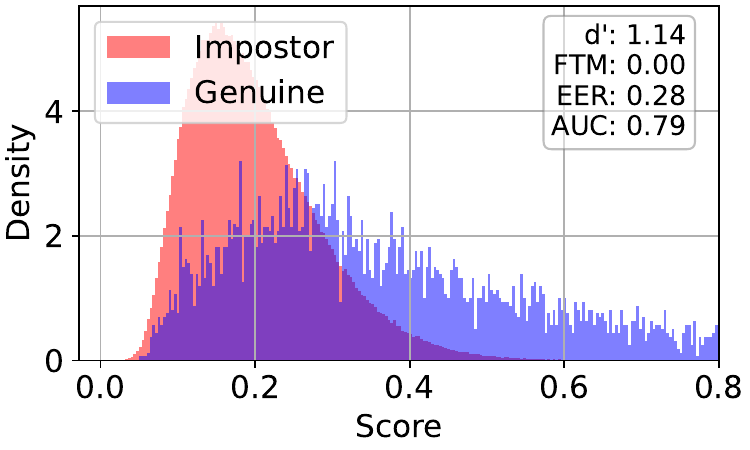}}
    \subfloat[]{\includegraphics[width=0.20\linewidth]{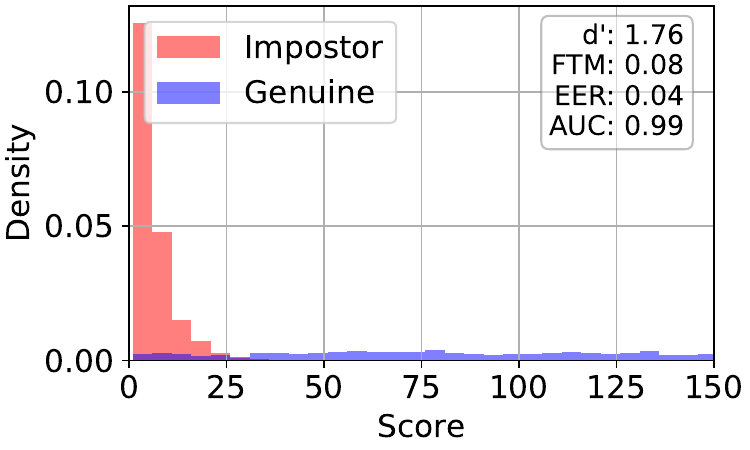}}
    \vskip-6mm
    \subfloat[]{\includegraphics[width=0.20\linewidth]{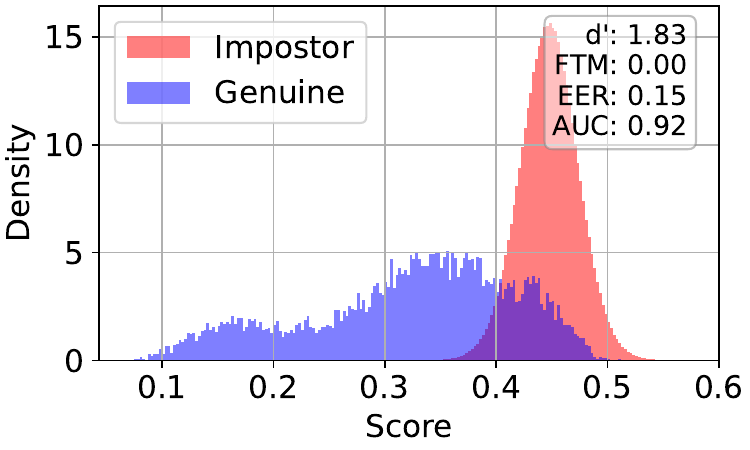}}
    \subfloat[]{\includegraphics[width=0.20\linewidth]{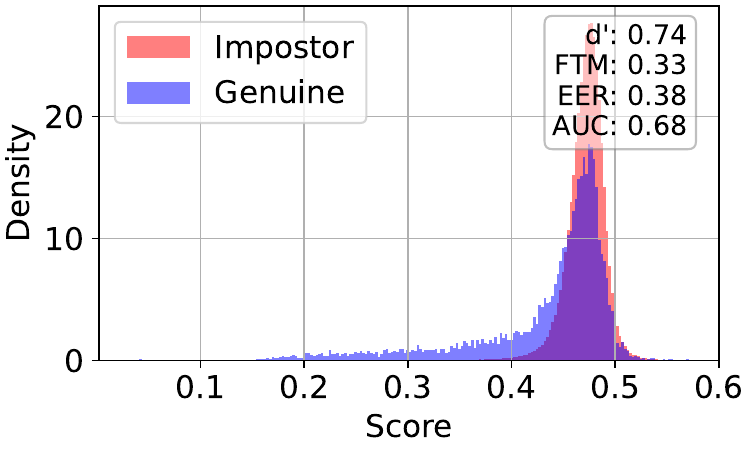}}
    \subfloat[]{\includegraphics[width=0.20\linewidth]{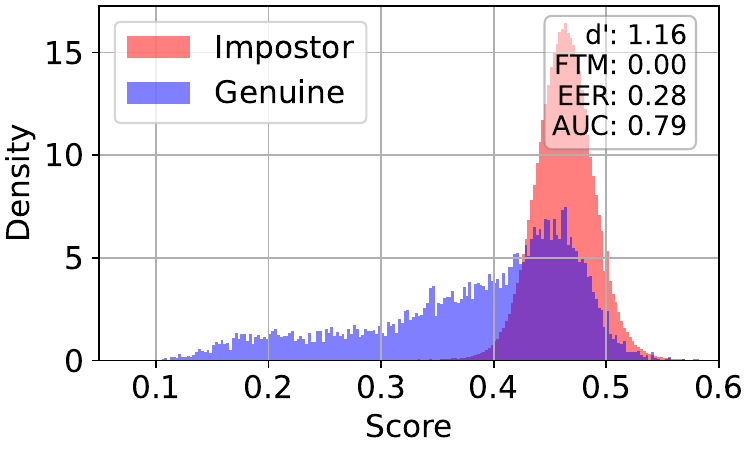}}
    \subfloat[]{\includegraphics[width=0.20\linewidth]{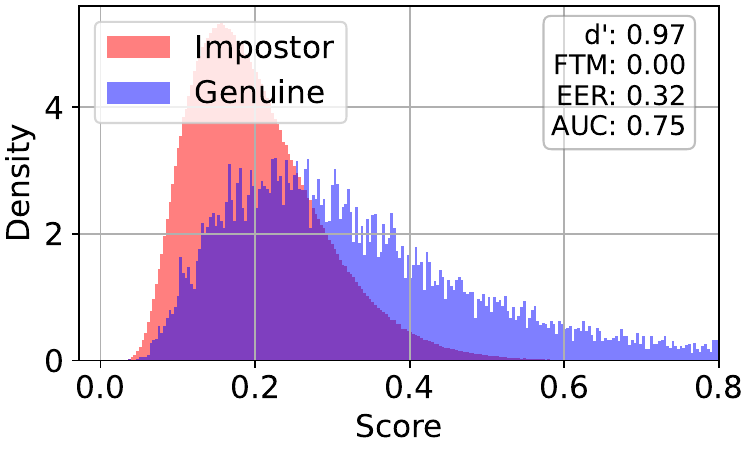}}
    \subfloat[]{\includegraphics[width=0.20\linewidth]{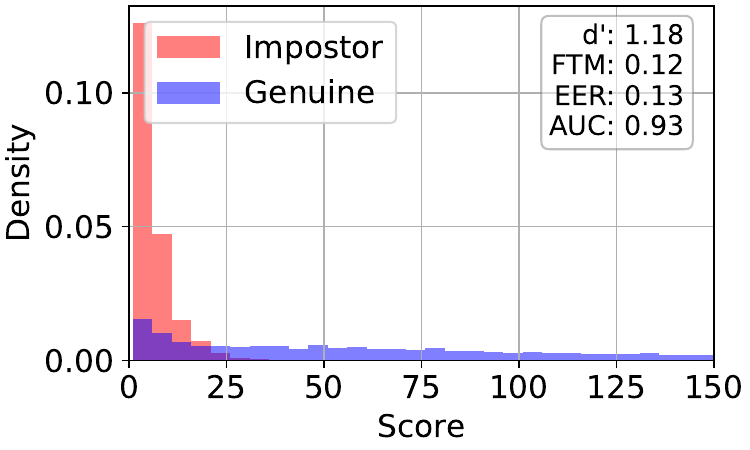}}
    \vskip-6mm
    \subfloat[HDBIF]{\includegraphics[width=0.20\linewidth]{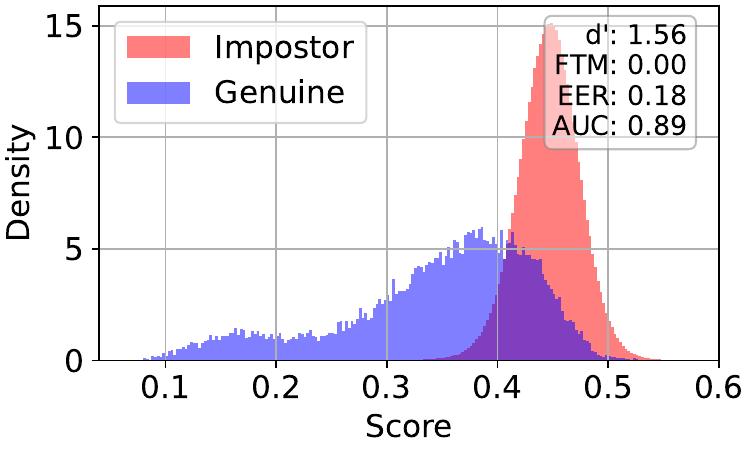}}
    \subfloat[USIT]{\includegraphics[width=0.20\linewidth]{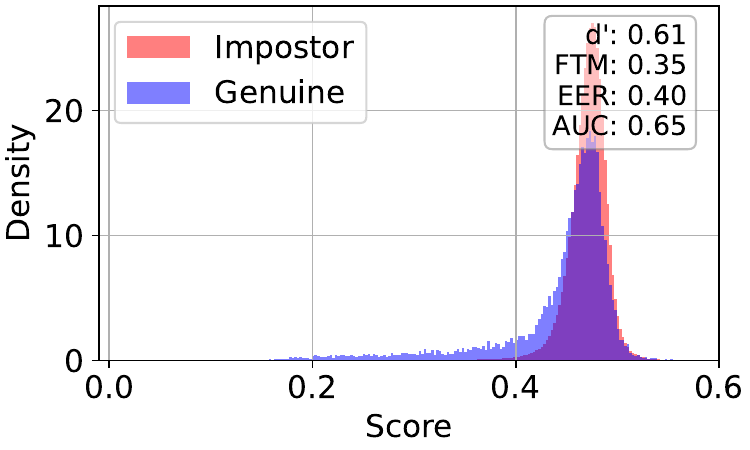}}
    \subfloat[OSIRIS]{\includegraphics[width=0.20\linewidth]{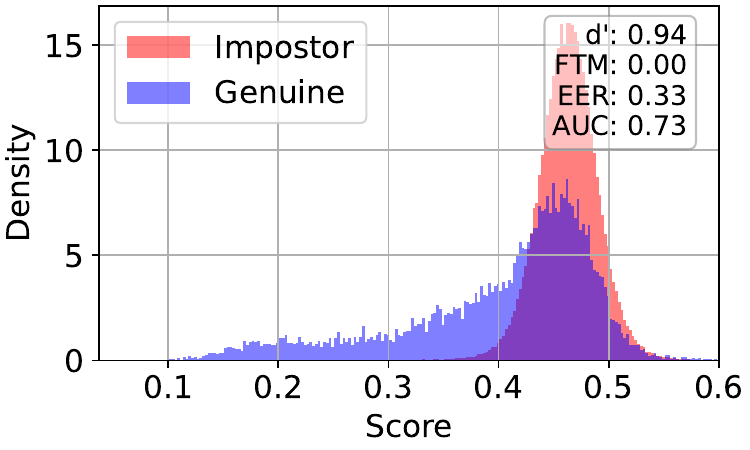}}
    \subfloat[DGR]{\includegraphics[width=0.20\linewidth]{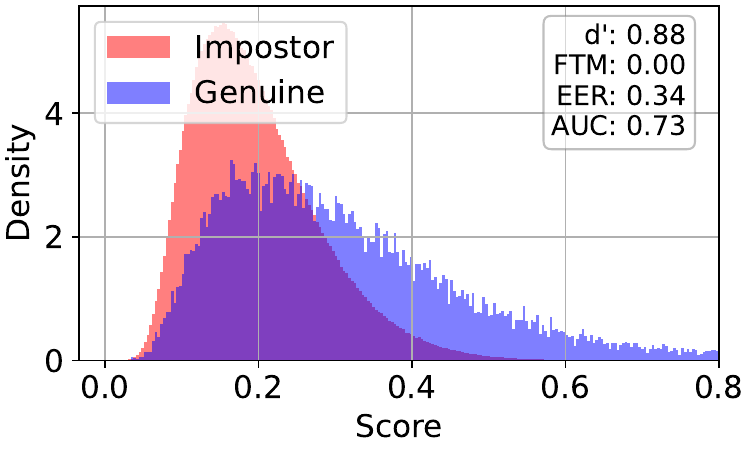}}
    \subfloat[VeriEye]{\includegraphics[width=0.20\linewidth]{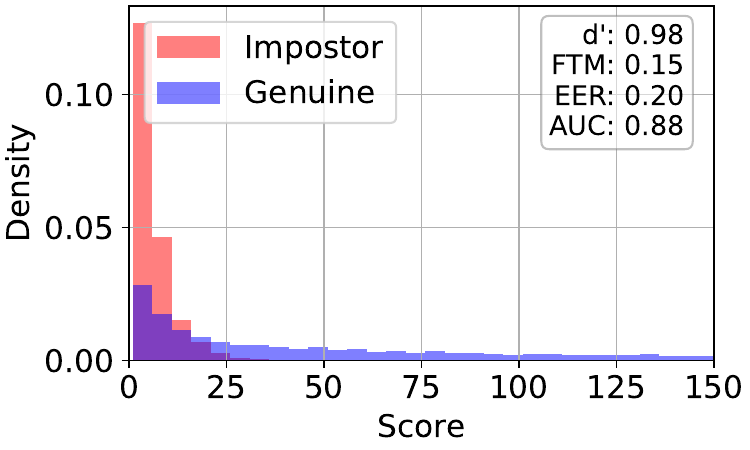}}        
  \caption{Same as in Fig. \ref{fig:combined-score-dist} in the main paper (the comparison score distributions obtained for five different iris matching algorithms) except that {\bf only scored obtained for Warsaw NIR samples} are shown, and results are {\bf split by the PMI range}: up to 24h (top row), up to 72h (second row), up to 240h (third row), and all PMIs combined (bottom row). }
  \label{fig:warsaw-score-dist} 
\end{figure*}

The distributions of scores for NIR data for five iris recognition methods (OSIRIS, USIT, HDBIF, DGR, and VeriEye; see Sec.~\ref{sec:IrisRecognitionMethods}), combined for both datasets (\datasetname~and Warsaw) and all PMIs are shown in Fig. \ref{fig:combined-score-dist} (\added{bottom row}). Selected performance metrics, such as $d'$ \cite{daugman2007new}, Equal Error Rate (EER), Failure to Match (FTM), and Area Under the Receiver Operating Characteristic (ROC) curve (AUC), are included in the plots. (Similar plots, but illustrating comparison score distributions separately for \datasetname~and Warsaw datasets, broken by the PMI range, are included in Figs ~\ref{fig:dcmeo2-score-dist} and \ref{fig:warsaw-score-dist}). 

The performance metrics indicate that the HDBIF matcher outperformed others. Conversely, an example deep learning-based method, DGR, showed the worst performance along with the non deep learning-based USIT approach. However, the HDBIF matcher, contrary to other methods, was the only approach whose segmentation model was trained with a small set of post-mortem iris images, what demonstrates a typical advantage of domain-specific adaptation.

The Failure-to-Match (FTM) rate (indicating the percentage of matching pairs, in which at least one sample was not usable for matching), shown in the plots, is very uneven across the matchers. That is, the USIT and VeriEye matchers implement relatively fastidious image quality checks before executing the matching. However, this rigorous filtering does not necessarily result in superior performance on the remaining high-quality samples, suggesting that quality control mechanisms may need to be tailored specifically for post-mortem iris recognition. Interestingly, the HDBIF matcher, which achieved the best performance in this study, had minimal FTM rates. Instead of relying on extensive filtering, HDBIF utilized a segmentation model trained on post-mortem iris data, demonstrating effective adaptation to the forensic iris domain.

\begin{figure*}[hbt!]
    \centering
       \subfloat[HDBIF]{\includegraphics[width=0.20\linewidth]{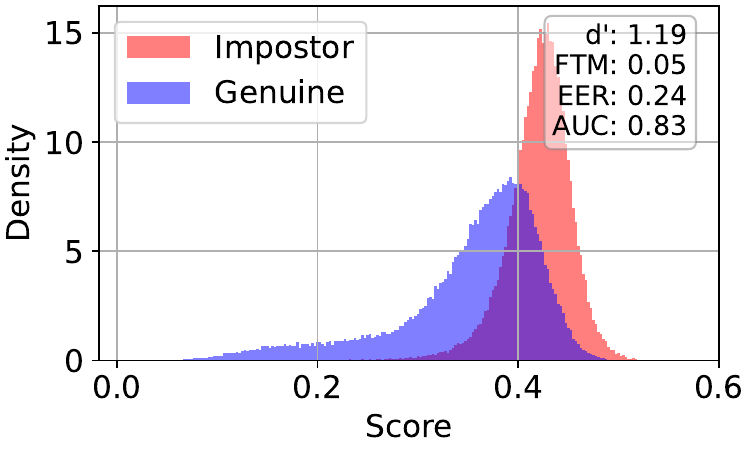}}
       \subfloat[USIT]{\includegraphics[width=0.20\linewidth]{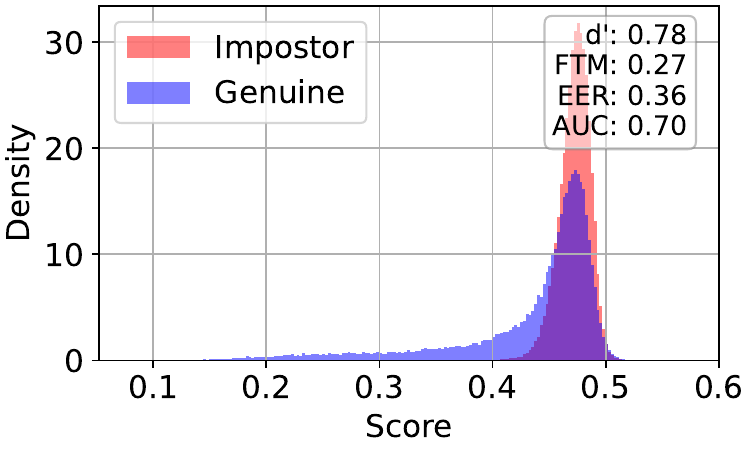}}
       \subfloat[OSIRIS]{\includegraphics[width=0.20\linewidth]{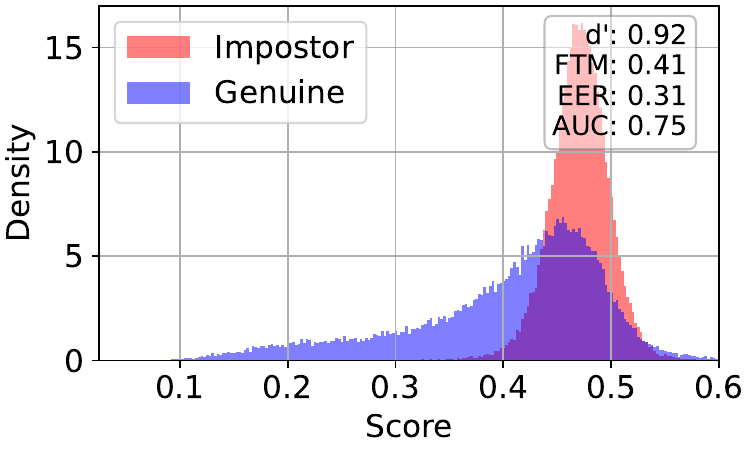}}
       \subfloat[DGR]{\includegraphics[width=0.20\linewidth]{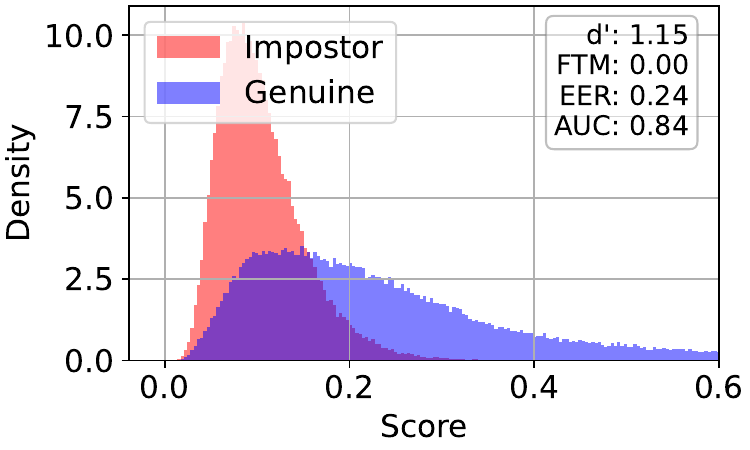}}
       \subfloat[VeriEye]{\includegraphics[width=0.20\linewidth]{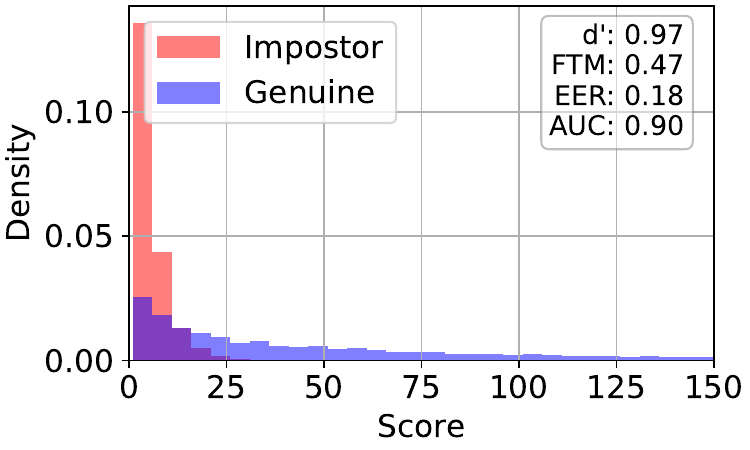}}
  \caption{\added{The comparison score distributionss for the {\bf combined \datasetname~+~Warsaw RGB datasets}, obtained for five different iris matchers.}}
  \label{fig:rgb-score-dist} 
\end{figure*}

\added{Fig. \ref{fig:rgb-score-dist} shows the comparison score distributions and performance metrics for RGB images. To make samples as close as possible to those collected in the NIR spectrum, we used only the red channel from the RGB images. The obtained results suggest that RGB images give much worse performance compared to the NIR samples. Also, the Failed-To-Match (FTM) rates are higher for all matchers when using RGB iris images. This analysis suggests that RGB images are not as effective for post-mortem iris recognition as NIR samples.}

\begin{figure*}[hbt!]
    \centering
       \subfloat[HDBIF]{\includegraphics[width=0.20\linewidth]{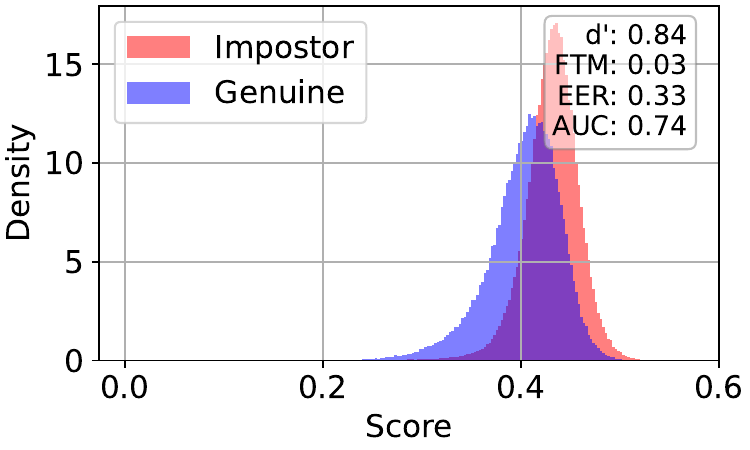}}
       \subfloat[USIT]{\includegraphics[width=0.20\linewidth]{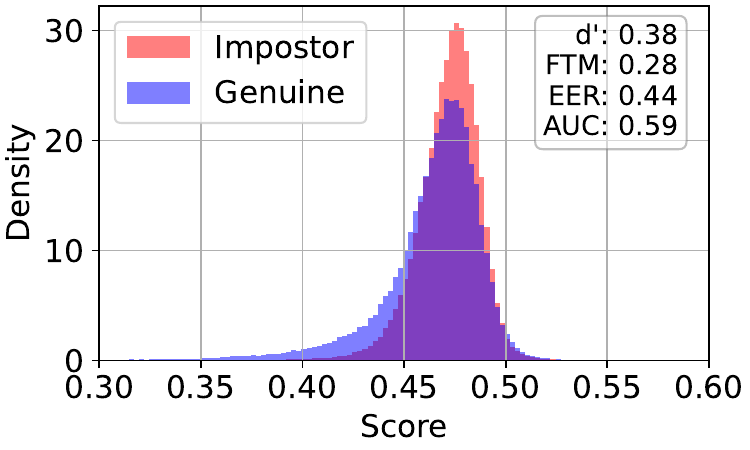}}
       \subfloat[OSIRIS]{\includegraphics[width=0.20\linewidth]{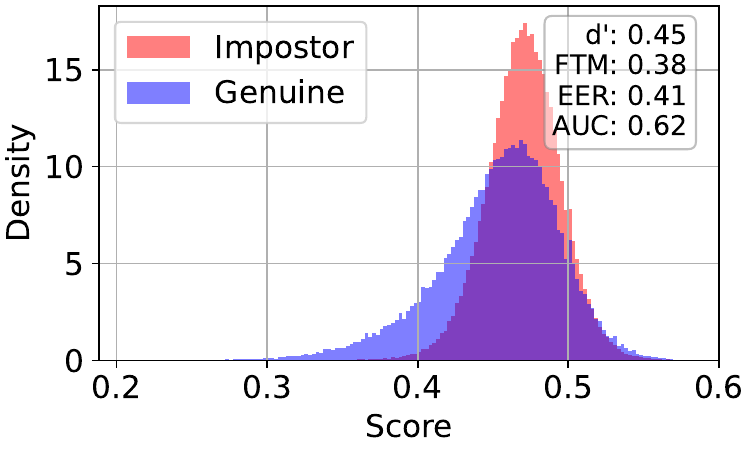}}
       \subfloat[DGR]{\includegraphics[width=0.20\linewidth]{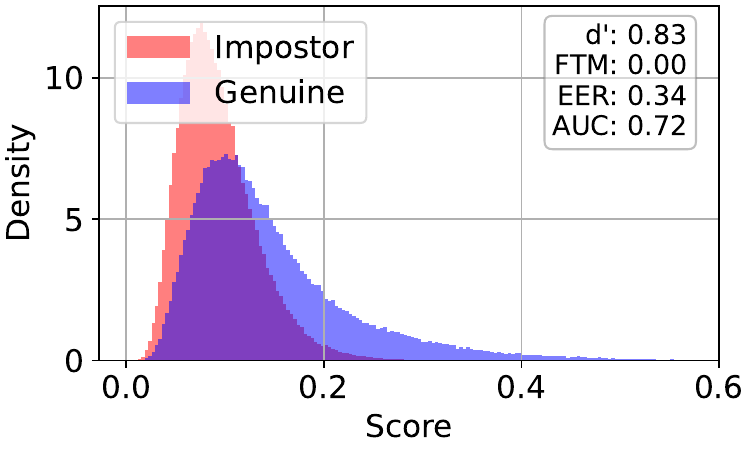}}
       \subfloat[VeriEye]{\includegraphics[width=0.20\linewidth]{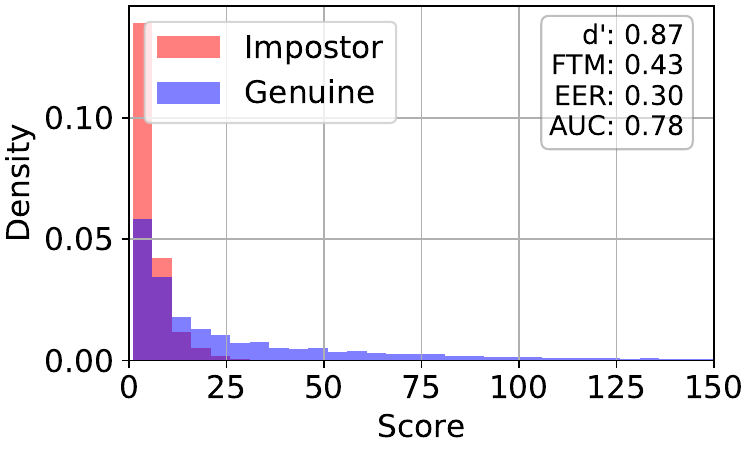}}
  \caption{\added{The comparison score distributions for the {\bf combined \datasetname~+~Warsaw} datasets obtained in a cross-wavelength experiment.}}
  \label{fig:cross-score-dist} 
\end{figure*}

\added{Fig. \ref{fig:cross-score-dist} shows the comparison score distributions obtained in the cross-wavelength post-mortem iris recognition, in which corresponding NIR and RGB images were compared. This simulates a scenario where, for instance, NIR gallery samples are available but probe samples are only captured in visible light. The cross-wavelength post-mortem iris recognition performance is even worse than when using either NIR or RGB samples alone.}

\begin{figure*}[hbt!]
    \centering
       \subfloat[HDBIF]{\includegraphics[width=0.20\linewidth]{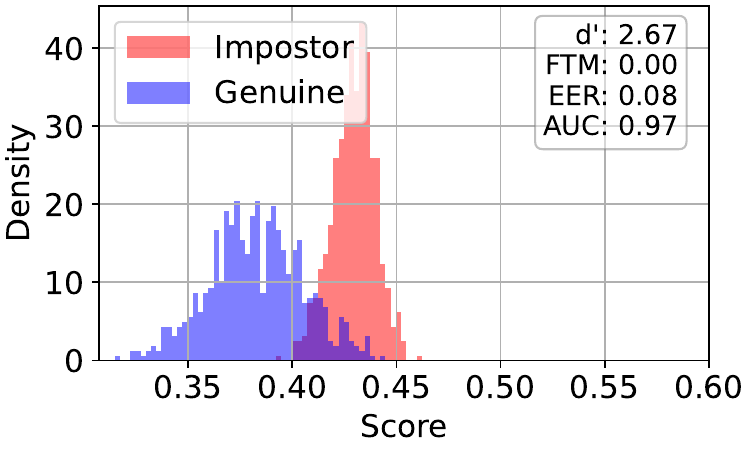}}
       \subfloat[USIT]{\includegraphics[width=0.20\linewidth]{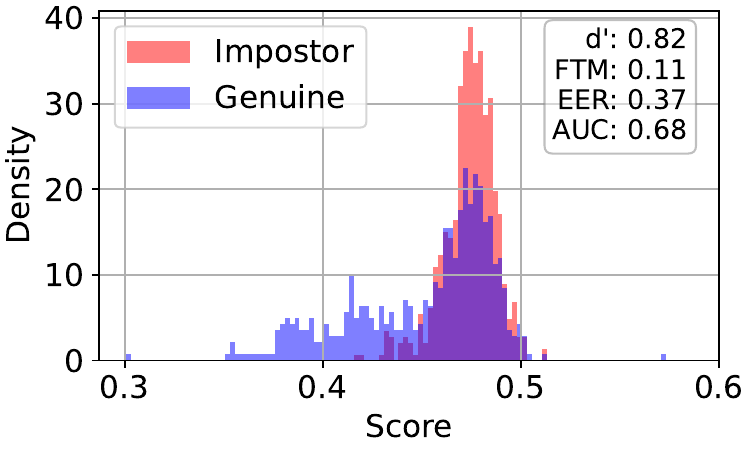}}
       \subfloat[OSIRIS]{\includegraphics[width=0.20\linewidth]{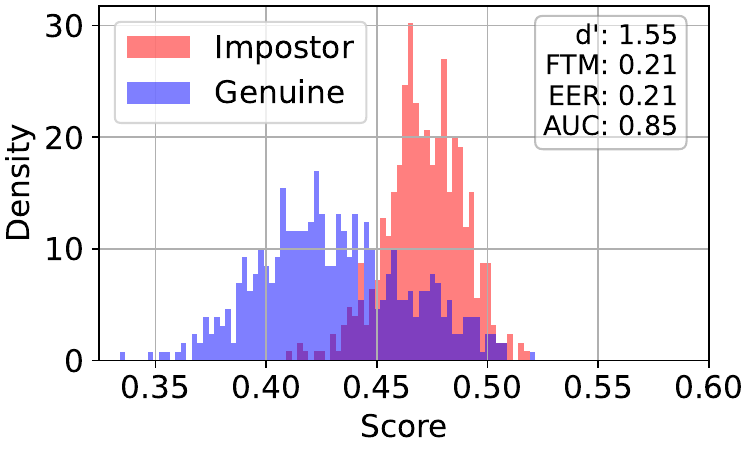}}
       \subfloat[DGR]{\includegraphics[width=0.20\linewidth]{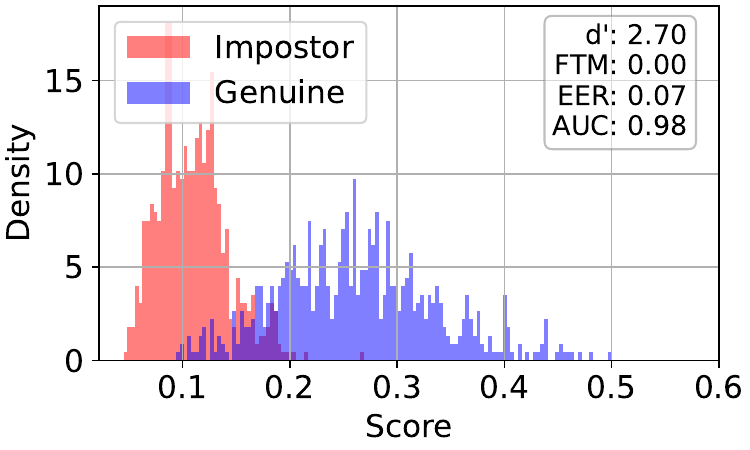}}
       \subfloat[VeriEye]{\includegraphics[width=0.20\linewidth]{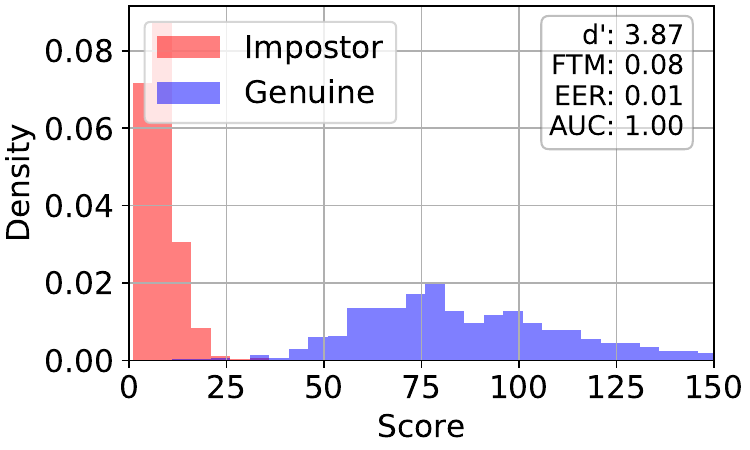}}
  \caption{\added{The comparison score distributions for the {\bf ante-mortem vs post-mortem samples}, obtained for five different iris matchers.}}
  \label{fig:ante-vs-post-score-dist} 
\end{figure*}

\added{Finally, Fig. \ref{fig:ante-vs-post-score-dist} illustrates the iris recognition performance when ante-mortem and corresponding post-mortem images are compared. Among all the matchers, the VeriEye achieved the best performance, with EER=1\% and an FTM=8\%, suggesting that such recognition is viable.}

\subsection{Impact of Biological Factors}

\subsubsection{Post-mortem Interval}

\added{Fig. \ref{fig:combined-score-dist}, \ref{fig:dcmeo2-score-dist}, and \ref{fig:warsaw-score-dist} show the selected performance metrics calculated for shorter ranges of the PMI of samples included in the matching pairs.} This is to illustrate the dynamics of the performance deterioration as a function of the PMI. It can be seen that the drop in separation between genuine and impostor score distributions is more significant when results for samples collected up to 24 hours post-mortem are juxtaposed to those obtained for samples collected up to 72 hours post-mortem. This observation is true for all five iris recognition methods. Comparing the latter set of scores with those obtained for samples collected up to 240 hours post mortem, it can be seen that the deterioration of identity verification performance is less significant. This suggests that the loss of identity information is not a linear function of time, and the largest degradation of the identification power may be expected in the first 72 hours after demise.

\added{The additional factor, beyond high PMI (which is correlated with tissue decomposition), contributing to matching failures is the excessive eye rotation due to less controlled acquisition done by a handheld sensor and not compensated by iris matchers.}

\subsubsection{Gender} 
\label{sec:gender}
We performed an experiment to check whether there is a statistically significant difference in the post-mortem iris recognition performance observed between gender groups. Since PMI is an important factor impacting performance, we balanced the PMI in both groups by sub-sampling the data, resulting in images representing 238 male and 91 female subjects, as shown earlier in Table \ref{tab:age-gender-data-stats}. While the average PMI is precisely balanced between males and females through this approach, the average age is also nearly balanced, with males averaging 55.8 years and females 57.7 years.

\begin{figure*}[hbt!]
    \centering
    \subfloat[]{%
       \includegraphics[width=0.20\linewidth]{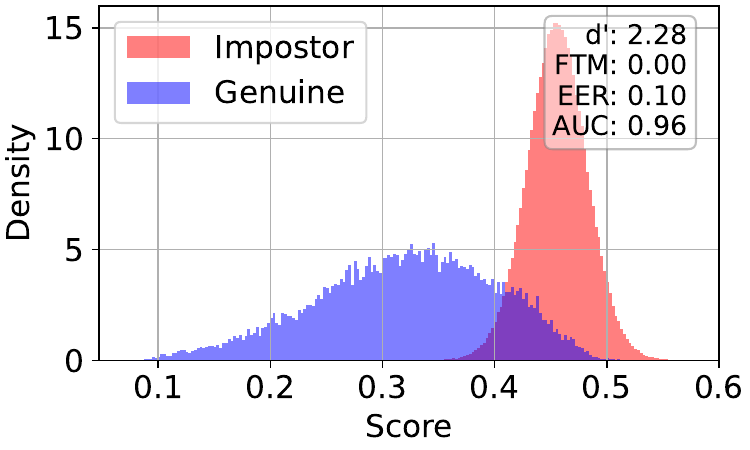}}
    \subfloat[]{%
        \includegraphics[width=0.20\linewidth]{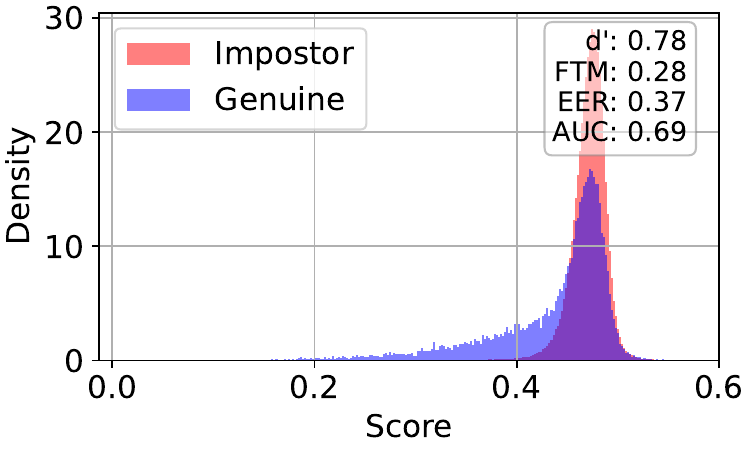}}
    \subfloat[]{%
        \includegraphics[width=0.20\linewidth]{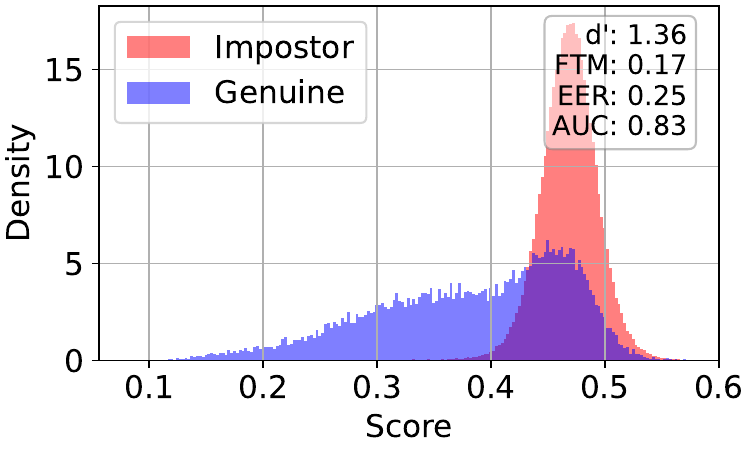}}
    \subfloat[]{%
        \includegraphics[width=0.20\linewidth]{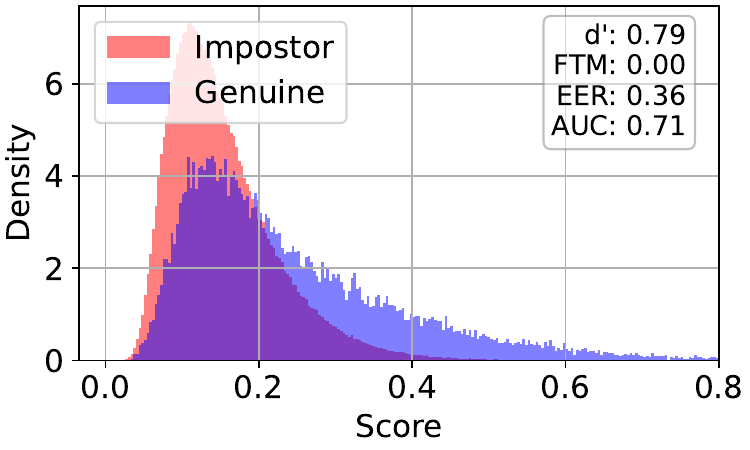}}
    \subfloat[]{%
        \includegraphics[width=0.20\linewidth]{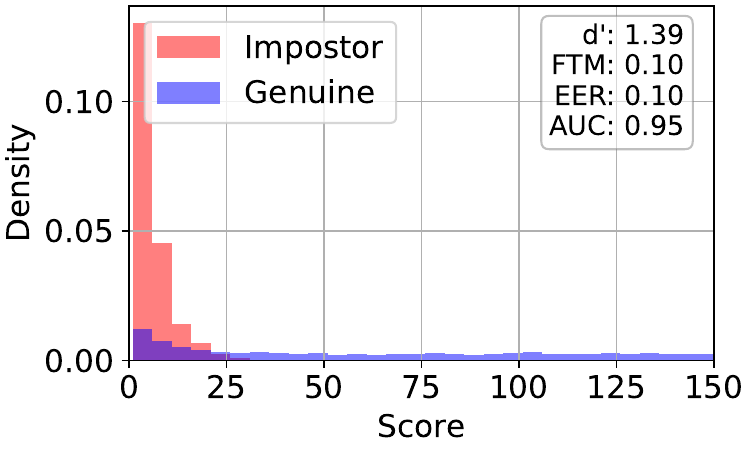}}
    \vskip-6mm
    \subfloat[HDBIF\label{1a:sex-score-dist}]{%
       \includegraphics[width=0.20\linewidth]{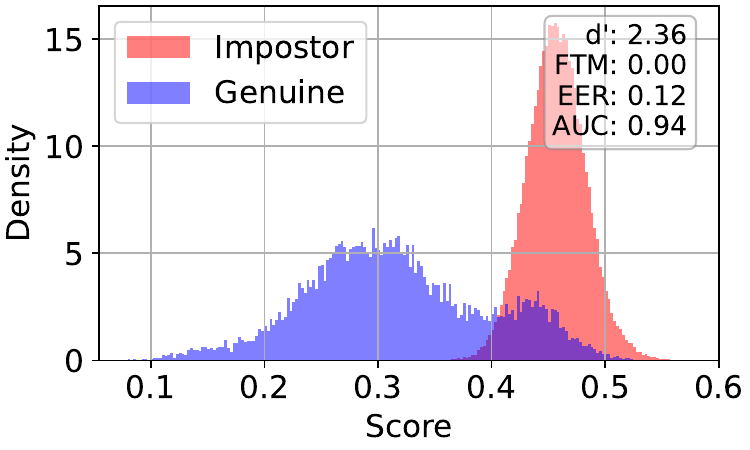}}
    \subfloat[USIT\label{1b:sex-score-dist}]{%
       \includegraphics[width=0.20\linewidth]{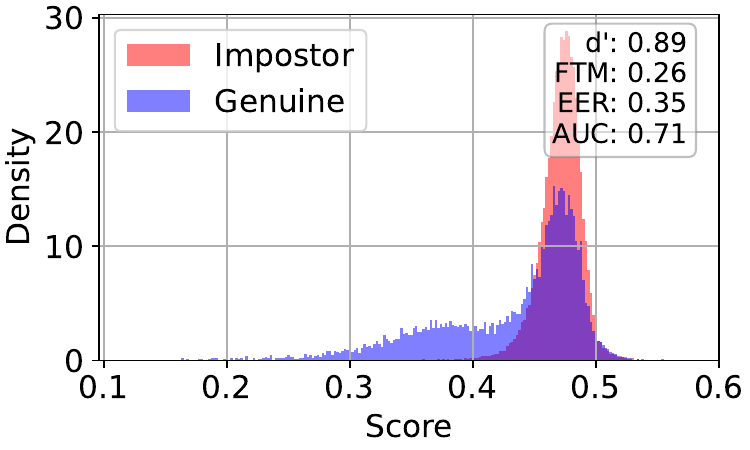}}
    \subfloat[OSIRIS\label{1c:sex-score-dist}]{%
       \includegraphics[width=0.20\linewidth]{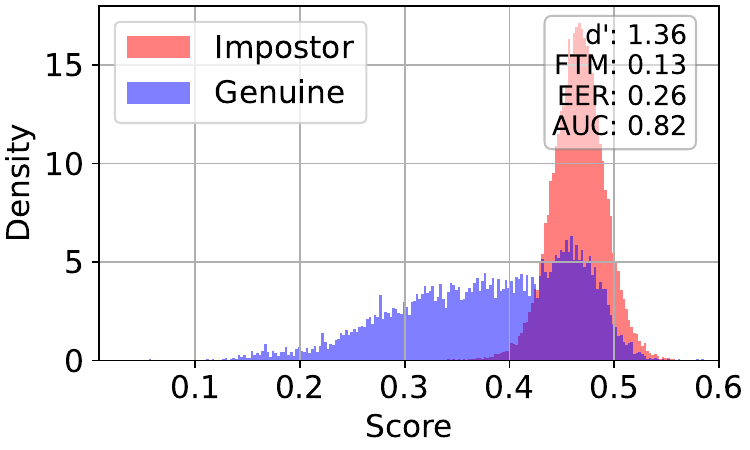}}
    \subfloat[DGR\label{1d:sex-score-dist}]{%
       \includegraphics[width=0.20\linewidth]{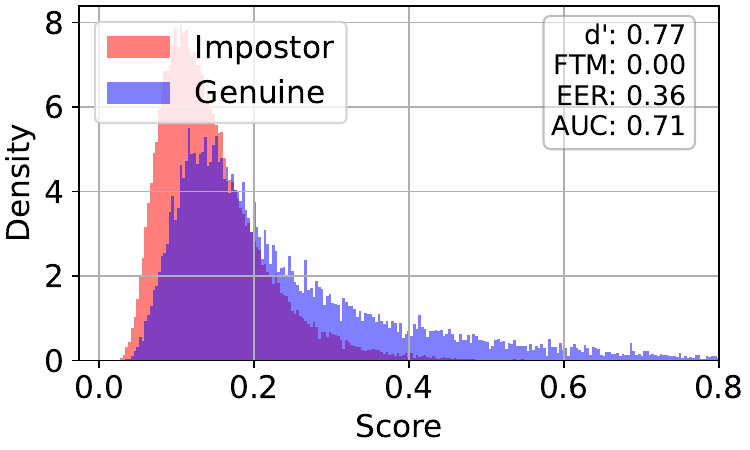}}
    \subfloat[VeriEye\label{1e:sex-score-dist}]{%
       \includegraphics[width=0.20\linewidth]{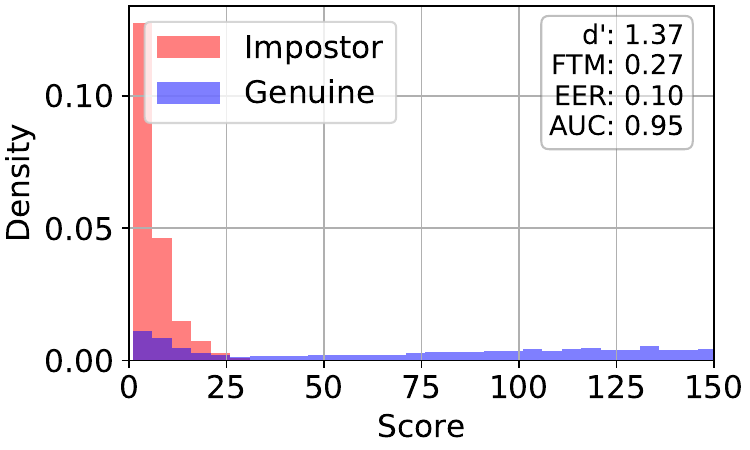}}
    
  \caption{The comparison score distributions for the {\bf combined \datasetname~and Warsaw datasets}, obtained for five different iris matching algorithms, and {\bf split by gender}: results for male and female subjects are shown in the top and bottom rows, respectively.}
  \label{fig:sex-score-dist} 
\end{figure*}

Figure \ref{fig:sex-score-dist} illustrates score distributions and performance metrics for independently for male and female subjects. HDBIF and USIT show better $d'$ values for females, whereas DGR and VeriEye favor males, with OSIRIS exhibiting neutral performance. However, other metrics show some discrepancies. Notably, HDBIF outperformed other matchers, achieving $d'$ of 2.28, EER of 10\%, and AUC of 96\% for males, compared to $d'$ of 2.36, EER of 12\%, and AUC of 94\% for females. Conversely, DGR performed the worst across both genders, with $d'$ of 0.79, EER of 36\%, and AUC of 71\% for males and nearly identical metrics for females ($d'$ of 0.77, EER of 36\%, and AUC of 71\%). The FTM rates vary across matchers and genders, with the highest for males observed in USIT (27.54\%) and for females in VeriEye (27.00\%), and reveal trends consistent with Sec. \ref{subsec:general-performance}.

To check the significance of the performance difference across gender, we randomly chose $50\%$ genuine pairs and $50\%$ impostor pairs thirty times and calculated the $d'$ value each time for both genders. Then, we performed ANOVA and Kruskal statistical hypothesis tests on these thirty $d'$ values of both groups. These steps are performed for all of the matchers. The $p-values$ obtained from these tests for all the matchers are below 0.05 (\added{see Table \ref{tab:gender}}), indicating a significant performance difference between males and females. However, due to the lack of consistency across all matchers in favoring one gender group over another, we cannot conclude, based on the dataset at hand, whether biological, population-wide factors contribute to better performance for either gender group. \added{Fig. \ref{fig:sex-heatmap} in the supplementary material presents the density heatmaps for ISO quality metrics broken by gender.}

\begin{table}[!htb]
\centering
\caption{Statistical hypothesis test results for male vs female average d-prime values.}
\begin{tabular}{ccccc}
\toprule
& \multicolumn{2}{c}{\textbf{ANOVA}}   & \multicolumn{2}{c}{\textbf{Kruskal-Wallis}} \\ 
\textbf{Method}  & \textbf{Statistics} & \textbf{$p$-value} & \textbf{Statistics} & \textbf{$p$-value} \\ \midrule
\textbf{HDBIF}   & 462.66              & 2.52e-29         & 44.26               & 2.87e-11         \\ 
\textbf{USIT}    & 4005.54             & 3.18e-55         & 44.26               & 2.87e-11         \\ 
\textbf{OSIRIS}  & 10.48               & 0.0019           & 8.83                & 0.0029           \\ 
\textbf{DGR}     & 214.83              & 3.68e-21         & 44.26               & 2.87e-11         \\ 
\textbf{VeriEye} & 22.49               & 1.41e-05         & 18.50               & 1.69e-05         \\ \bottomrule
\end{tabular}
\label{tab:gender}
\end{table}

\subsubsection{Age}
In order to check how the subject's age impacts the post-mortem iris recognition performance, we have split our data into three age groups (Group-1: 1 to 33 years old, Group-2: 34 to 66 years old, and Group-3: 67 to 99 years old) and balanced the PMI in each group, as presented in Table \ref{tab:age-gender-data-stats}.

\begin{figure*}[hbt!]
    \centering
    \subfloat[]{%
       \includegraphics[width=0.20\linewidth]{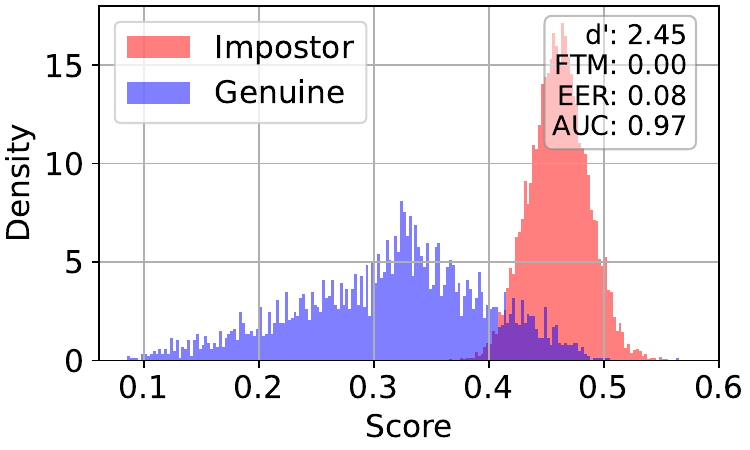}}
    \subfloat[]{%
        \includegraphics[width=0.20\linewidth]{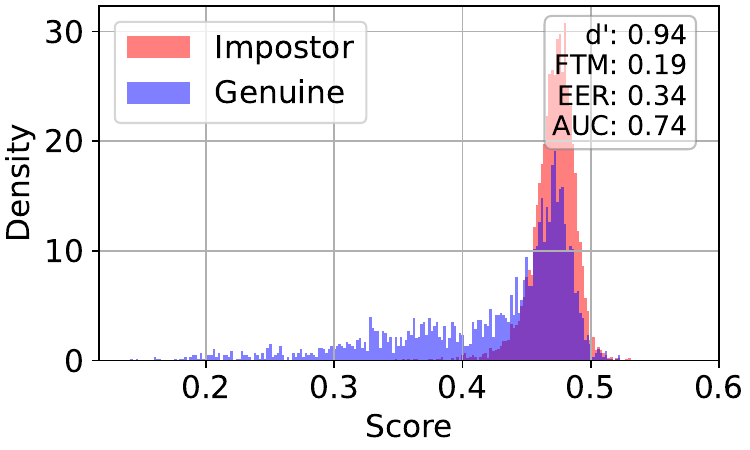}}
    \subfloat[]{%
       \includegraphics[width=0.20\linewidth]{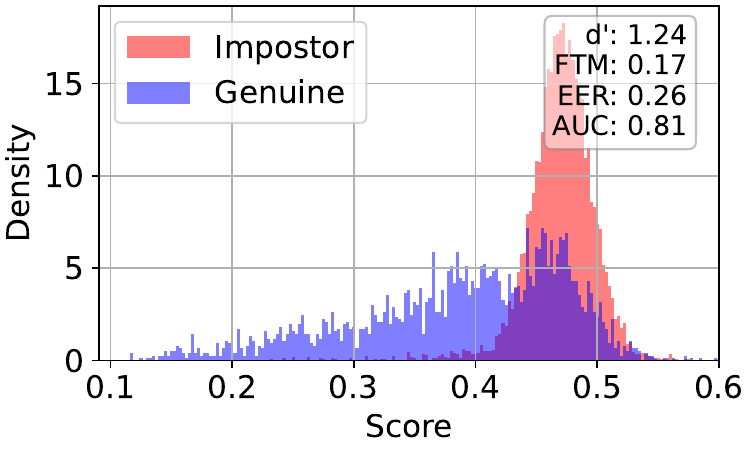}}
    \subfloat[]{%
        \includegraphics[width=0.20\linewidth]{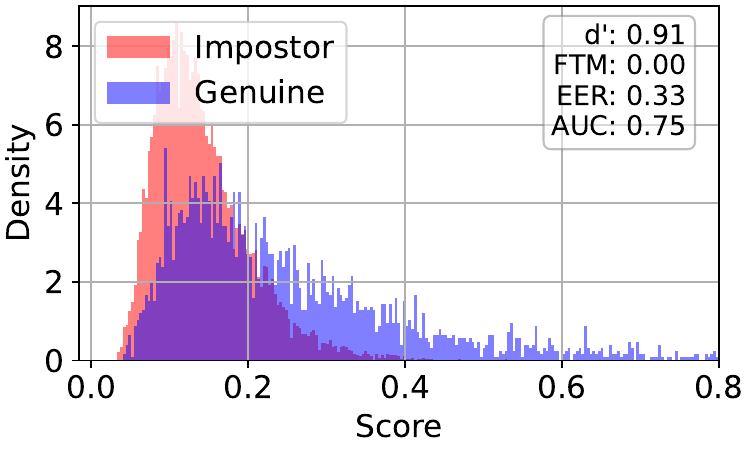}}
    \subfloat[]{%
       \includegraphics[width=0.20\linewidth]{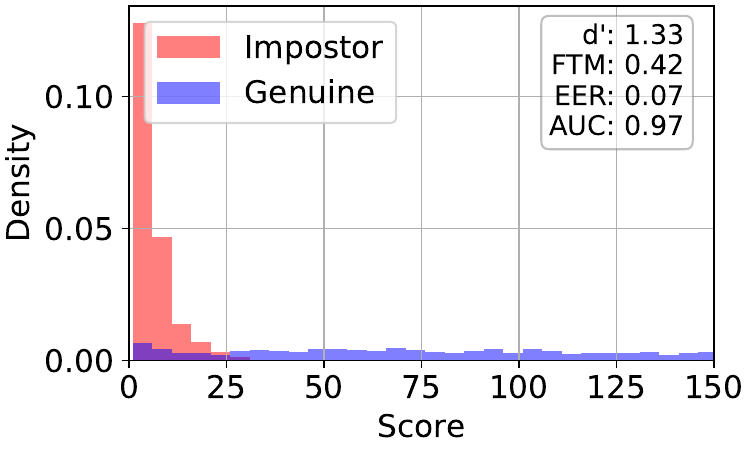}}
    \vskip-6mm
    \subfloat[]{%
       \includegraphics[width=0.20\linewidth]{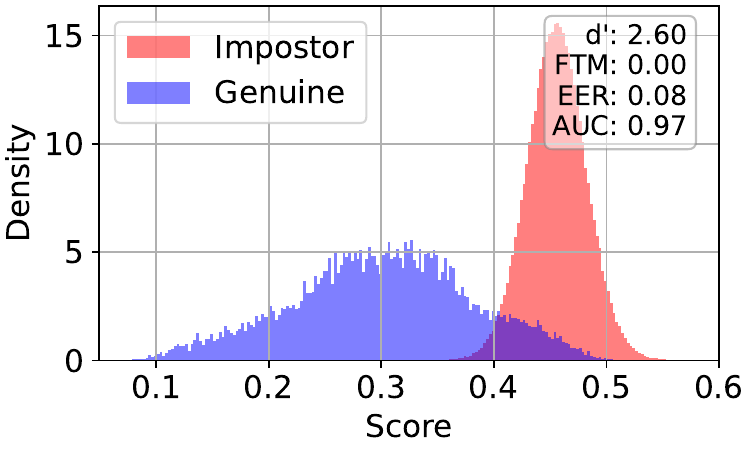}}
    \subfloat[]{%
       \includegraphics[width=0.20\linewidth]{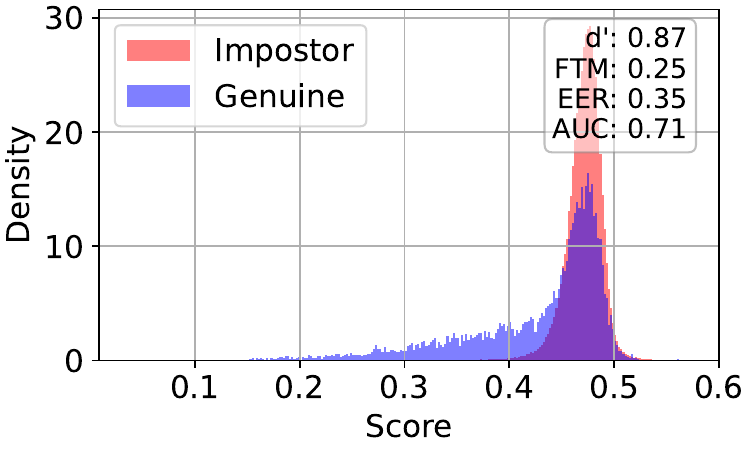}}
    \subfloat[]{%
       \includegraphics[width=0.20\linewidth]{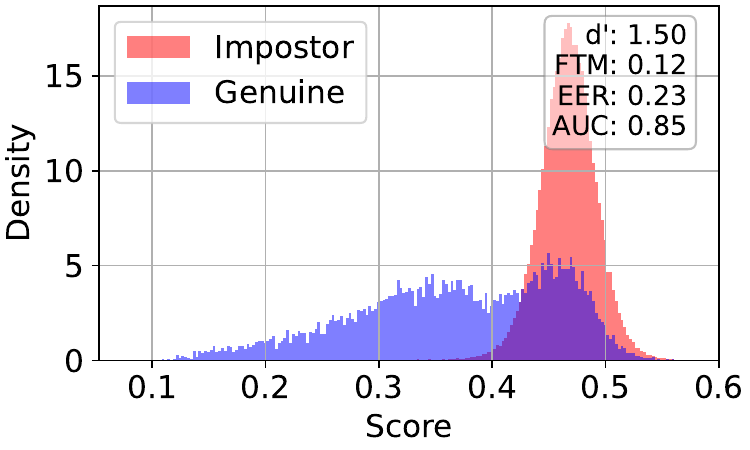}}
    \subfloat[]{%
       \includegraphics[width=0.20\linewidth]{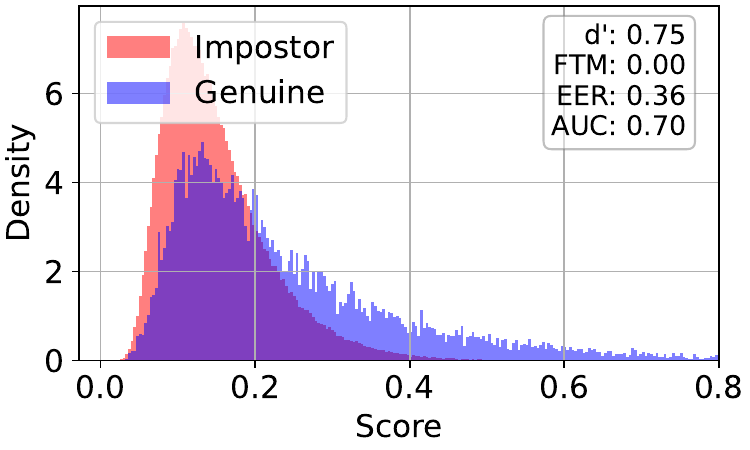}}
    \subfloat[]{%
       \includegraphics[width=0.20\linewidth]{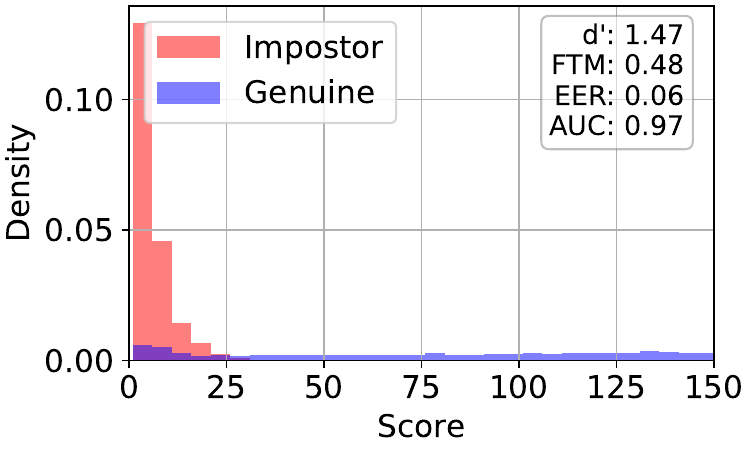}}
    \vskip-6mm
    \subfloat[HDBIF\label{1a:age-score-dist}]{%
       \includegraphics[width=0.20\linewidth]{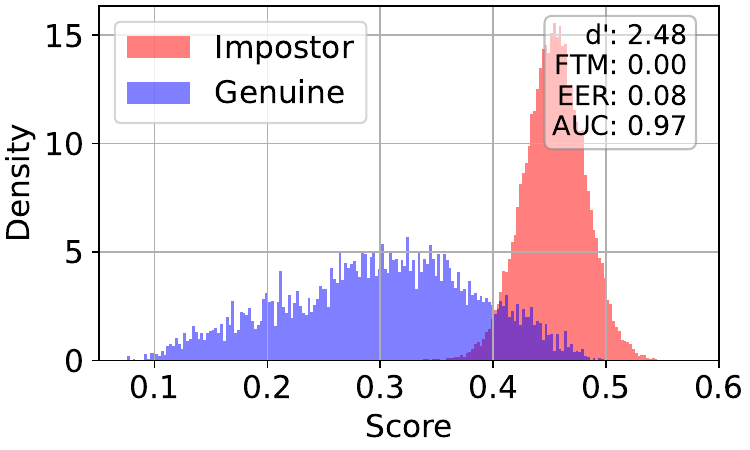}}
    \subfloat[USIT\label{1b:age-score-dist}]{%
       \includegraphics[width=0.20\linewidth]{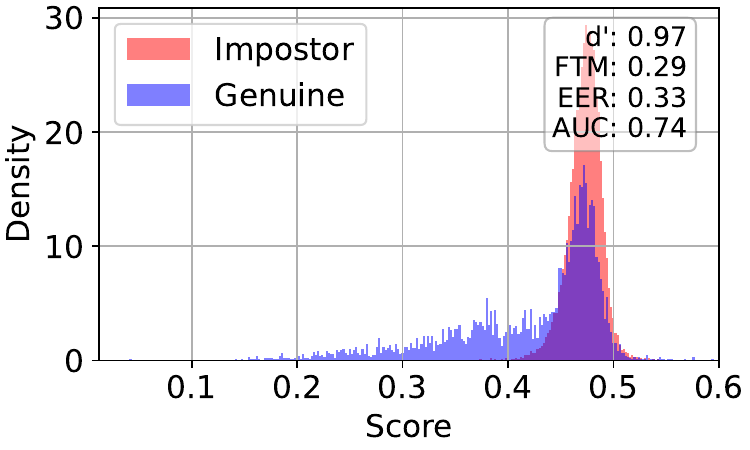}}
    \subfloat[OSIRIS\label{1c:age-score-dist}]{%
       \includegraphics[width=0.20\linewidth]{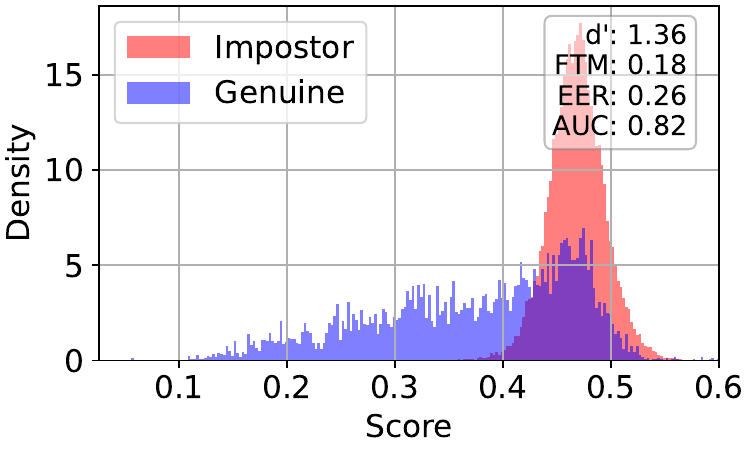}}
    \subfloat[DGR\label{1d:age-score-dist}]{%
       \includegraphics[width=0.20\linewidth]{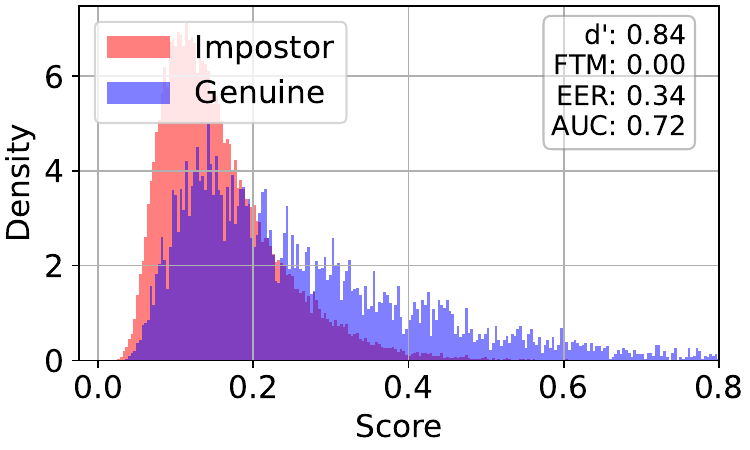}}
    \subfloat[VeriEye\label{1e:age-score-dist}]{%
       \includegraphics[width=0.20\linewidth]{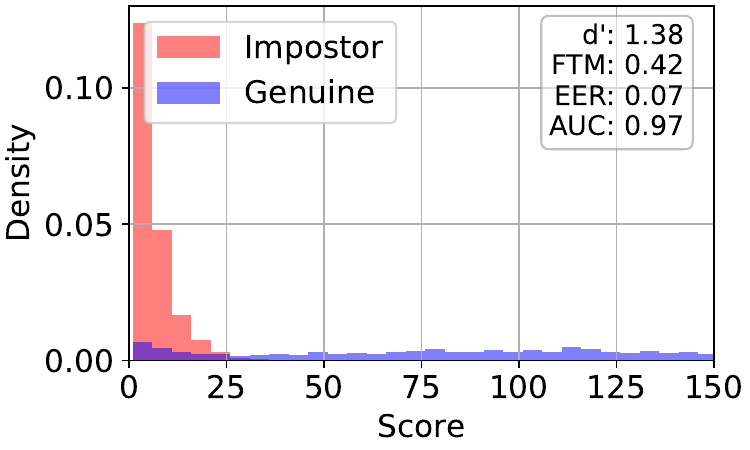}}
  \caption{The comparison score distributions for the {\bf combined \datasetname~and Warsaw datasets}, obtained for five different iris matching algorithms and shown separately for each {\bf subject's age} group in the rows, from top to bottom: 1-33 years old, 34-66 years old and 67-99 years old.}
  \label{fig:age-score-dist} 
\end{figure*}

Fig. \ref{fig:age-score-dist} presents matching score distributions and performance metrics highlighting performance inconsistency between age groups for different matchers. The $d'$, EER, and AUC scores indicate that the middle age group performs better than age groups 1 and 3 for HDBIF, OSIRIS, and VeriEye matchers. On the other hand, the older and younger age groups perform better than the middle age group for USIT and DGR, respectively. The FTM rates are higher for VeriEye, OSIRIS, and USIT, while HDBIF demonstrated minimal FTM rates and superior performance with $d'$ values of 2.45, 2.60, and 2.48, along with 8\% EER and 97\% AUC for all age groups. DGR had 0\% FTM rates but performed the worst. Since most matchers performed better for the middle age group, we assume that post-mortem iris recognition may be more effective for subjects aged 33 to 66 than for younger or older individuals. The statistical hypothesis tests (using the procedure described in Section~\ref{sec:gender}) imply that the performance difference between independent age groups is significant. \added{The exact values of the test statistics for both tests are included in Table \ref{tab:age-groups}. Fig. \ref{fig:age-heatmap} in the supplementary material presents the density heatmaps for ISO quality metrics broken by the age group.}

\begin{table}[!htb]
\centering
\caption{Statistical hypothesis test results for average d-prime values across three age groups.}
\begin{tabular}{ccccc}
\toprule
                 & \multicolumn{2}{c}{\textbf{ANOVA}}   & \multicolumn{2}{c}{\textbf{Kruskal-Wallis}} \\ 
\textbf{Method}  & \textbf{Statistics} & \textbf{$p$-value} & \textbf{Statistics} & \textbf{$p$-value} \\ \midrule
\textbf{HDBIF}   & 346.09              & 3.82e-42         & 65.44               & 6.14e-15         \\ 
\textbf{USIT}    & 316.21              & 1.23e-40         & 68.42               & 1.38e-15         \\ 
\textbf{OSIRIS}  & 1762.93             & 4.00e-71         & 79.12               & 6.59e-18         \\ 
\textbf{DGR}     & 1219.85             & 2.27e-64         & 79.12               & 6.59e-18         \\ 
\textbf{VeriEye} & 170.92              & 7.30e-31         & 68.83               & 1.13e-15         \\ 
\bottomrule
\end{tabular}
\label{tab:age-groups}
\end{table}

\subsection{Post-Mortem Iris Detection}
\label{sec:pad}
This section summarizes the results of the series of experiments evaluating the D-NetPAD model's capabilities in detecting post-mortem iris samples treated as presentation attacks. Each experiment (see Sec. \ref{sec:PostmortemIrisDetection} for detailed descriptions of all experimental scenarios) explores different dataset combinations for training and testing, ranging from no fine-tuning scenarios to cross-dataset validations and minimal training sets, highlighting the model's adaptability and effectiveness across varied conditions. 

\begin{figure*}[hbt!]
    \centering
    \begin{minipage}{\linewidth}
        \centering
        \includegraphics[width=0.24\linewidth]{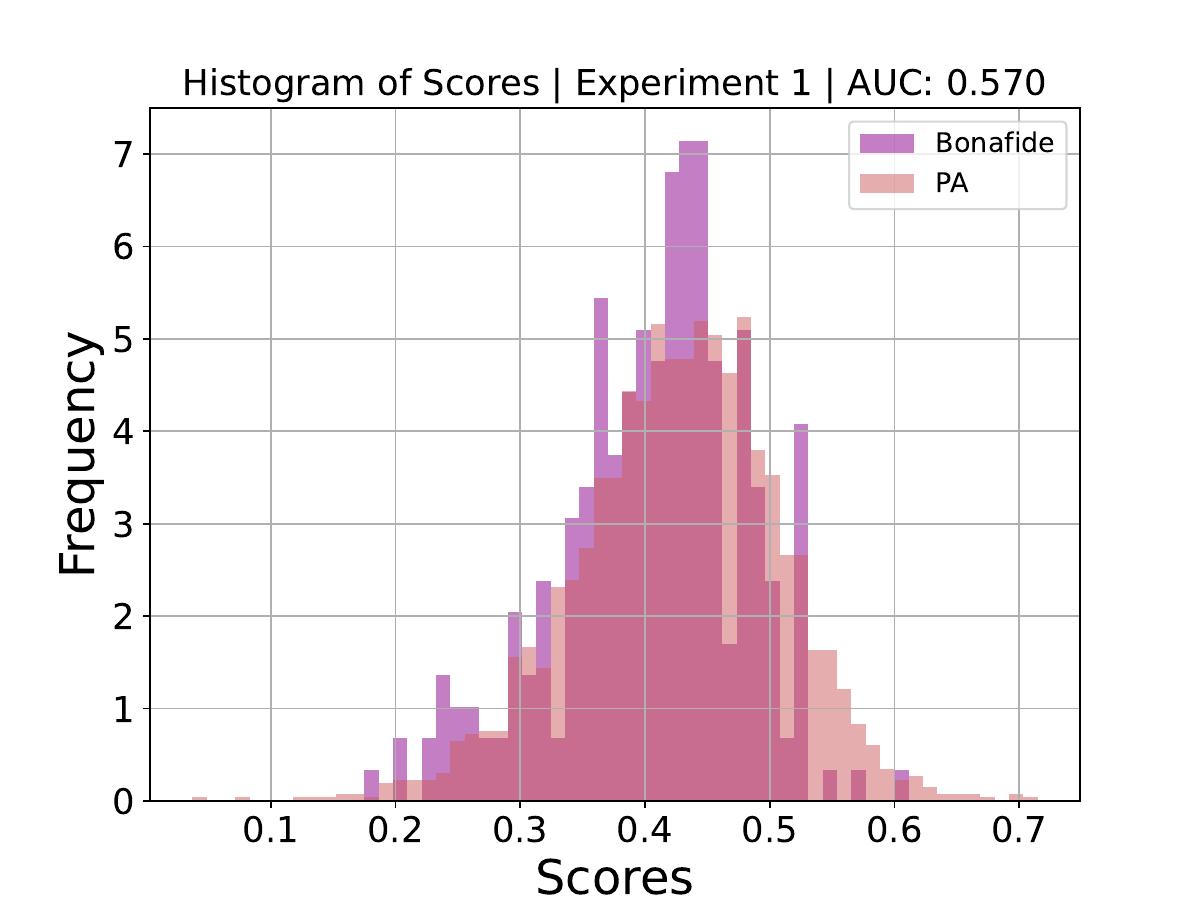}
        \includegraphics[width=0.24\linewidth]{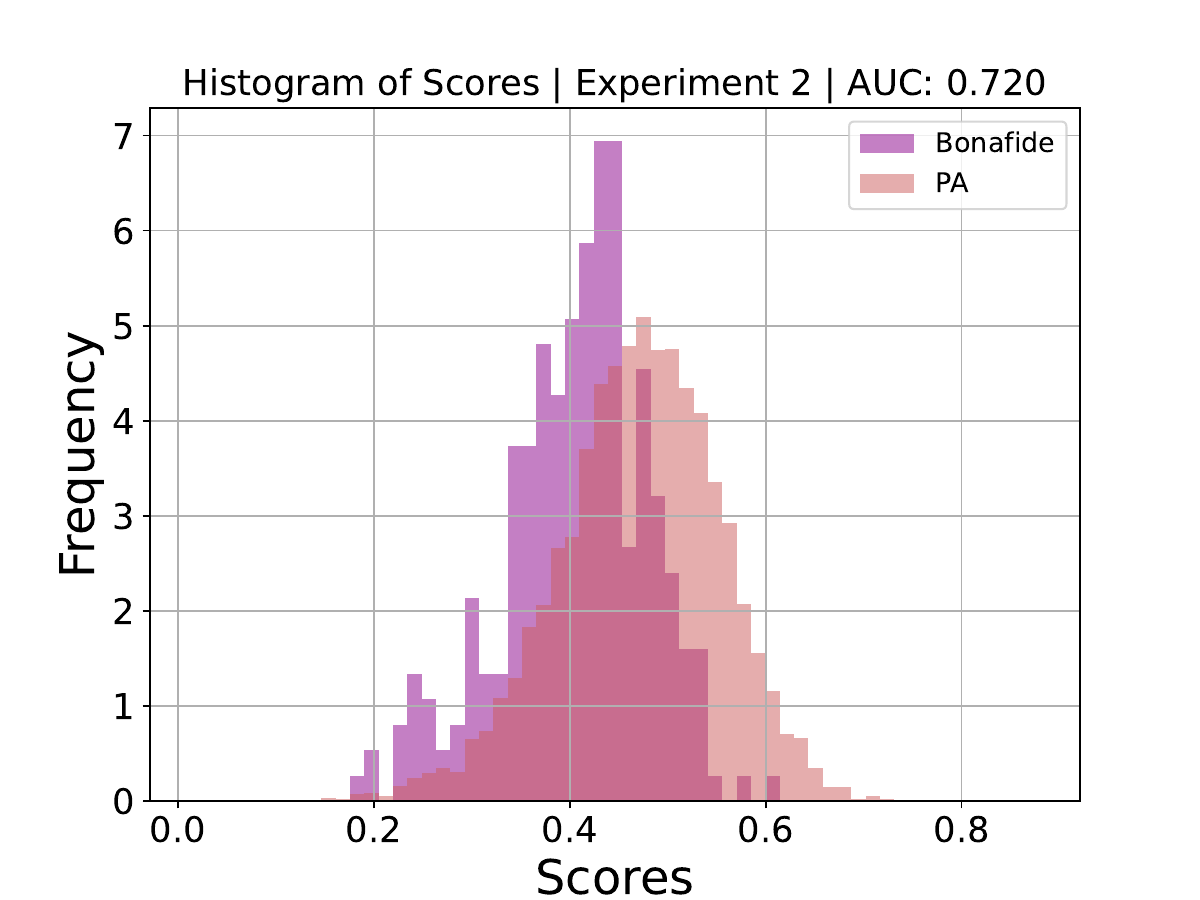}
        \hskip2mm
        \includegraphics[width=0.24\linewidth]{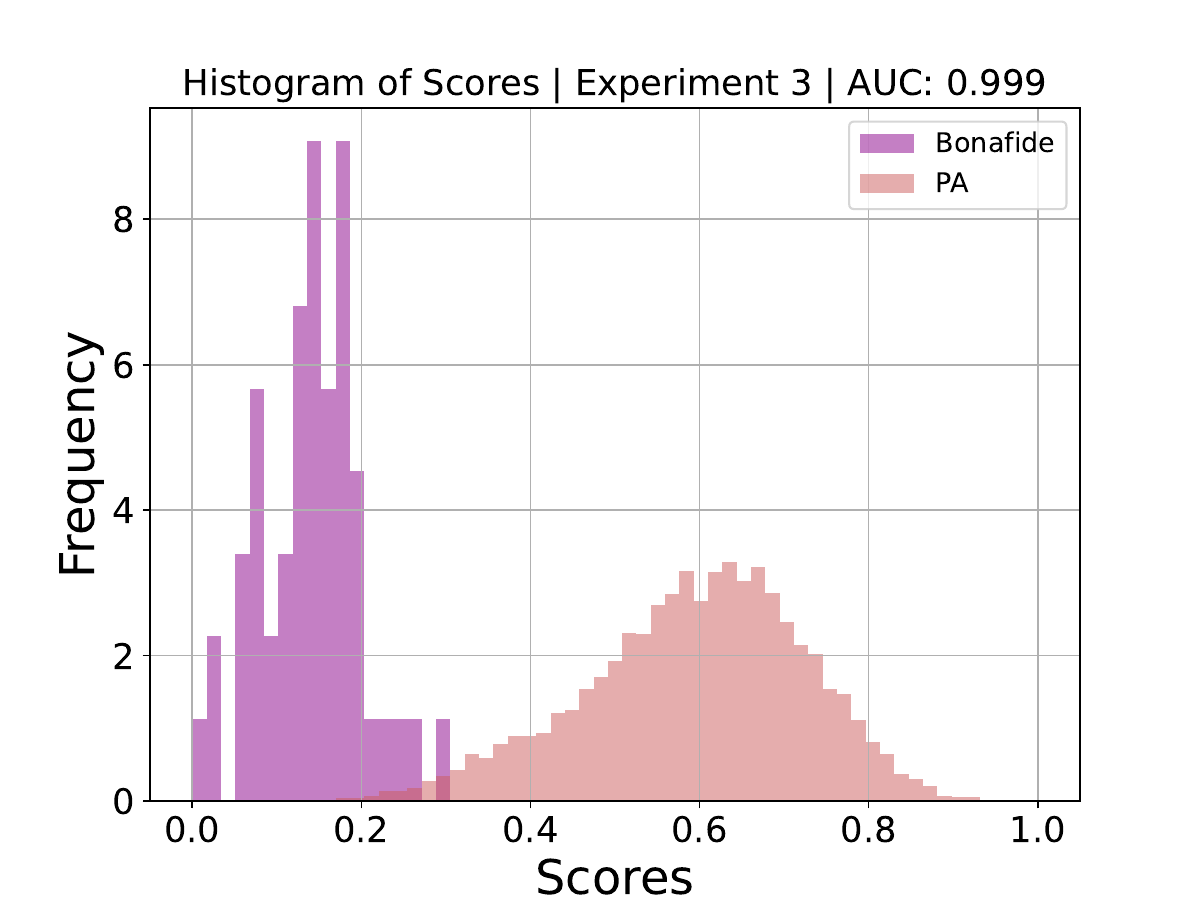}
        \includegraphics[width=0.24\linewidth]{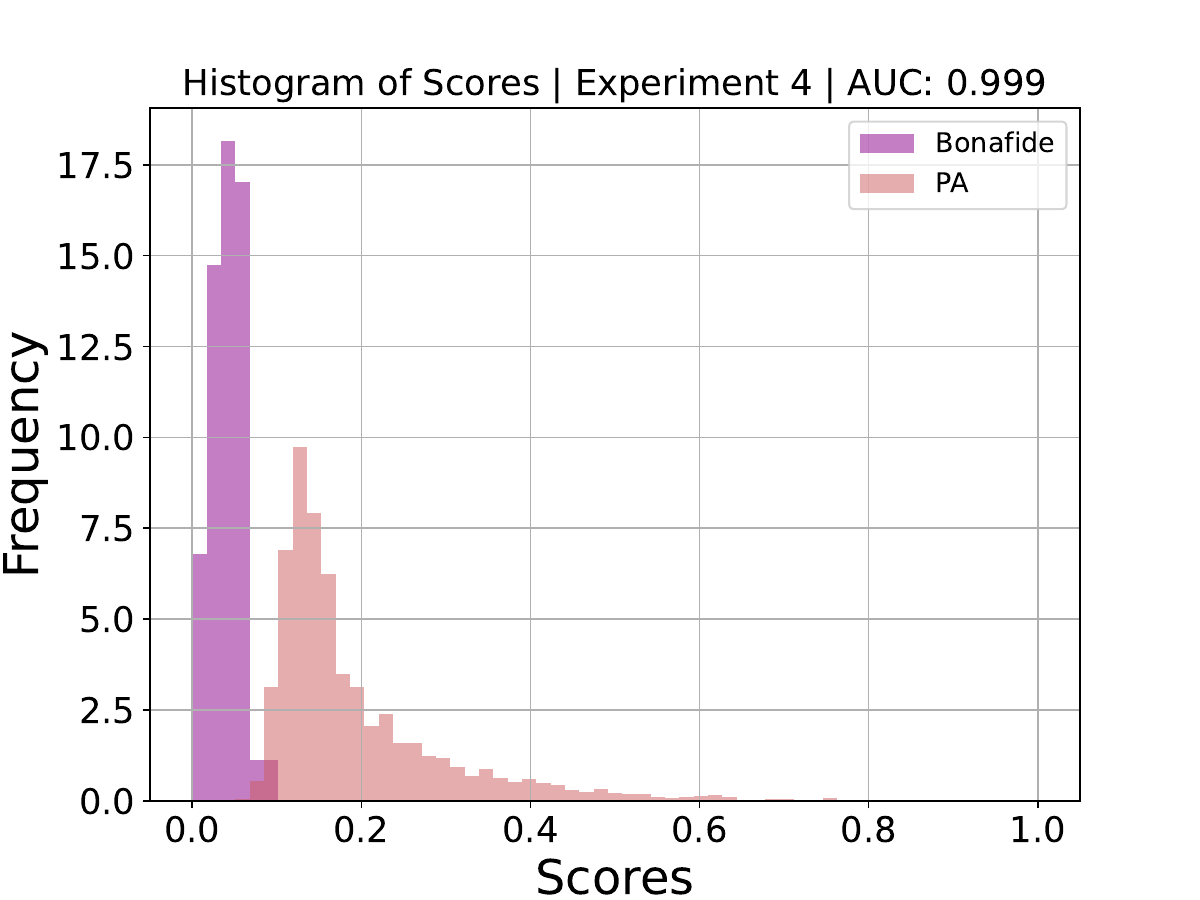}
    \end{minipage}
    \\
    \begin{minipage}{\linewidth}
        \centering
        \includegraphics[width=0.24\linewidth]{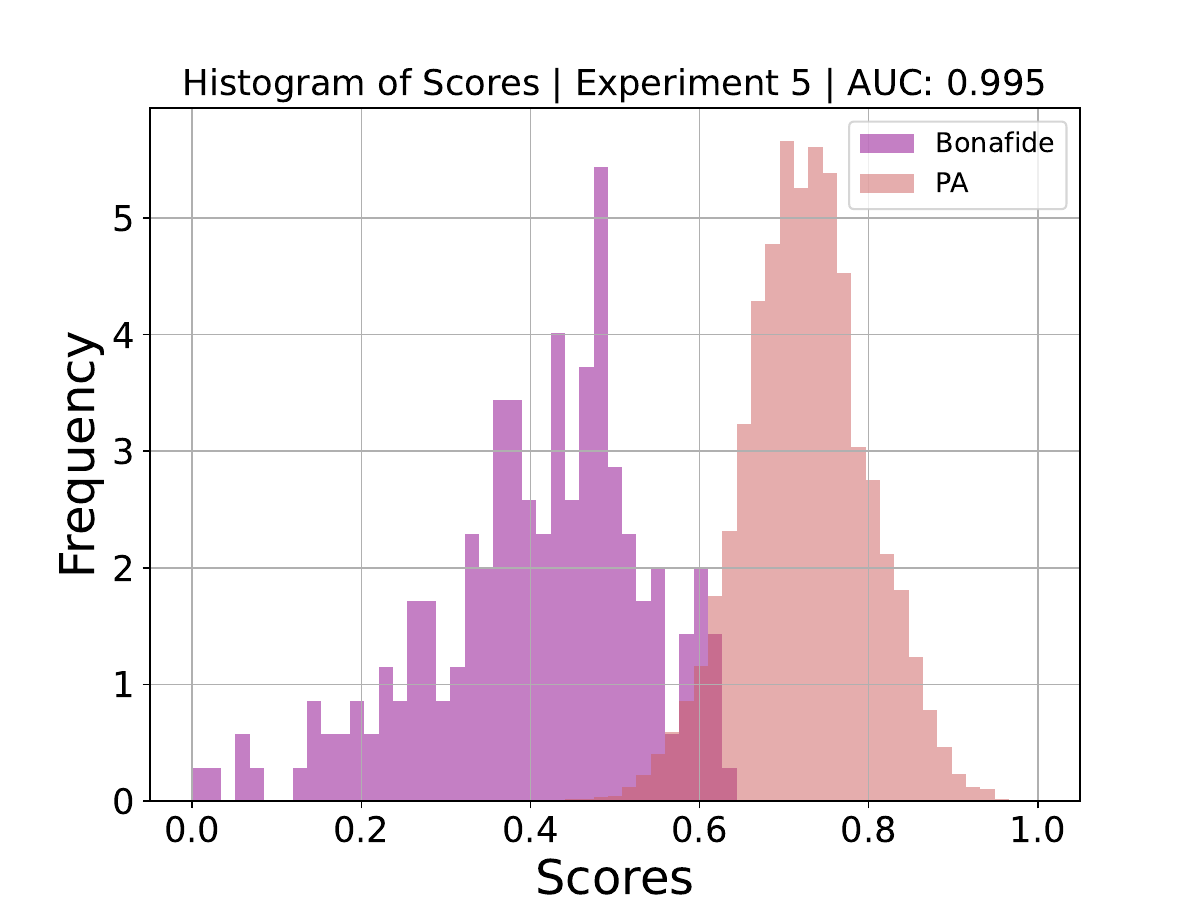}
        \includegraphics[width=0.24\linewidth]{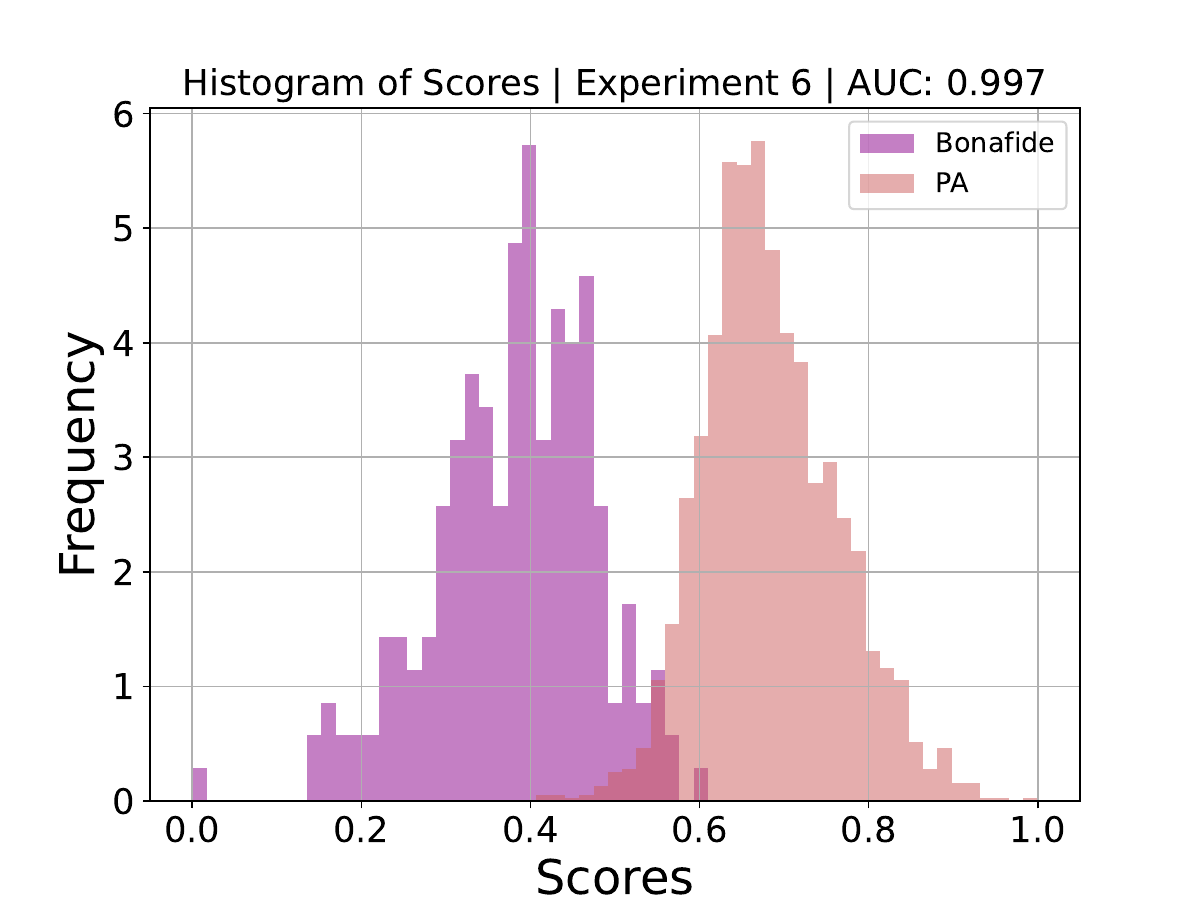}
        \hskip2mm
        \includegraphics[width=0.24\linewidth]{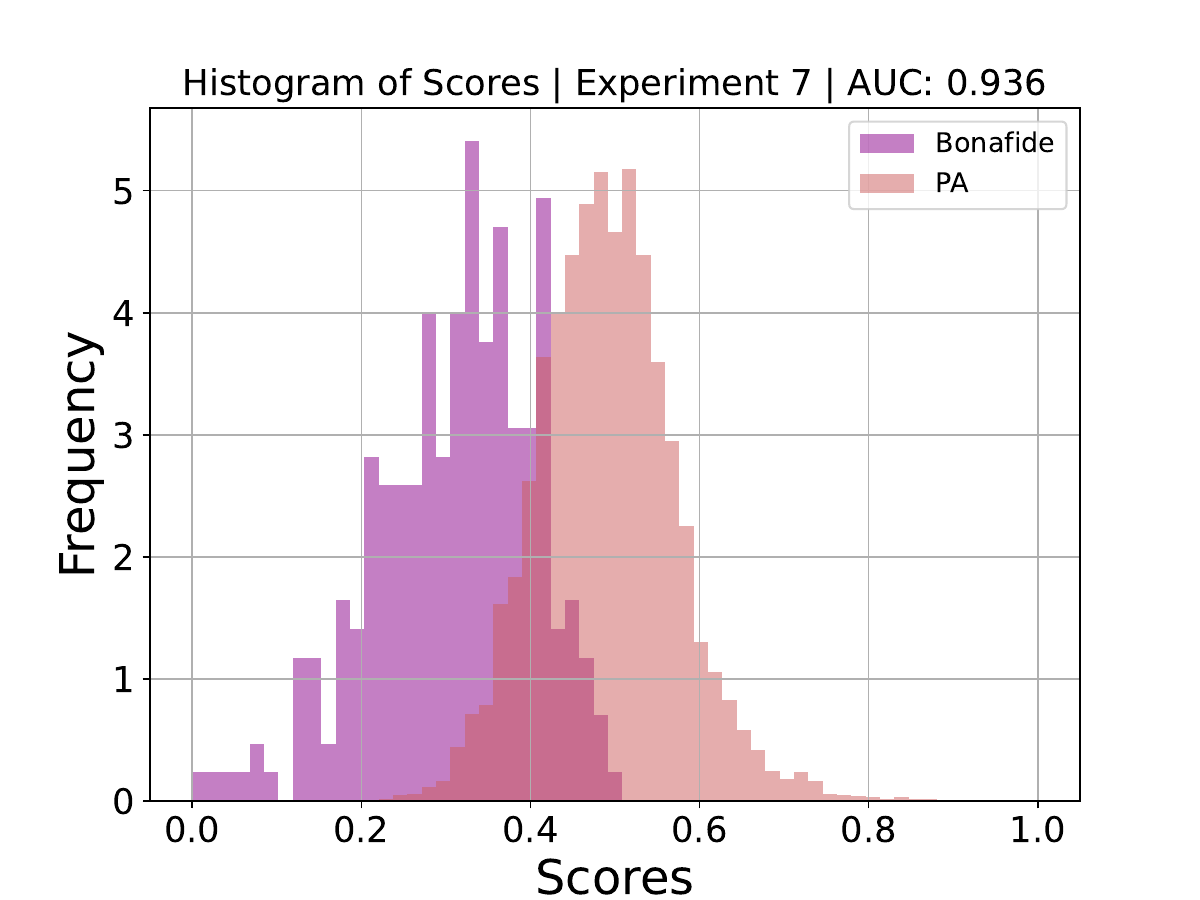}
        \includegraphics[width=0.24\linewidth]{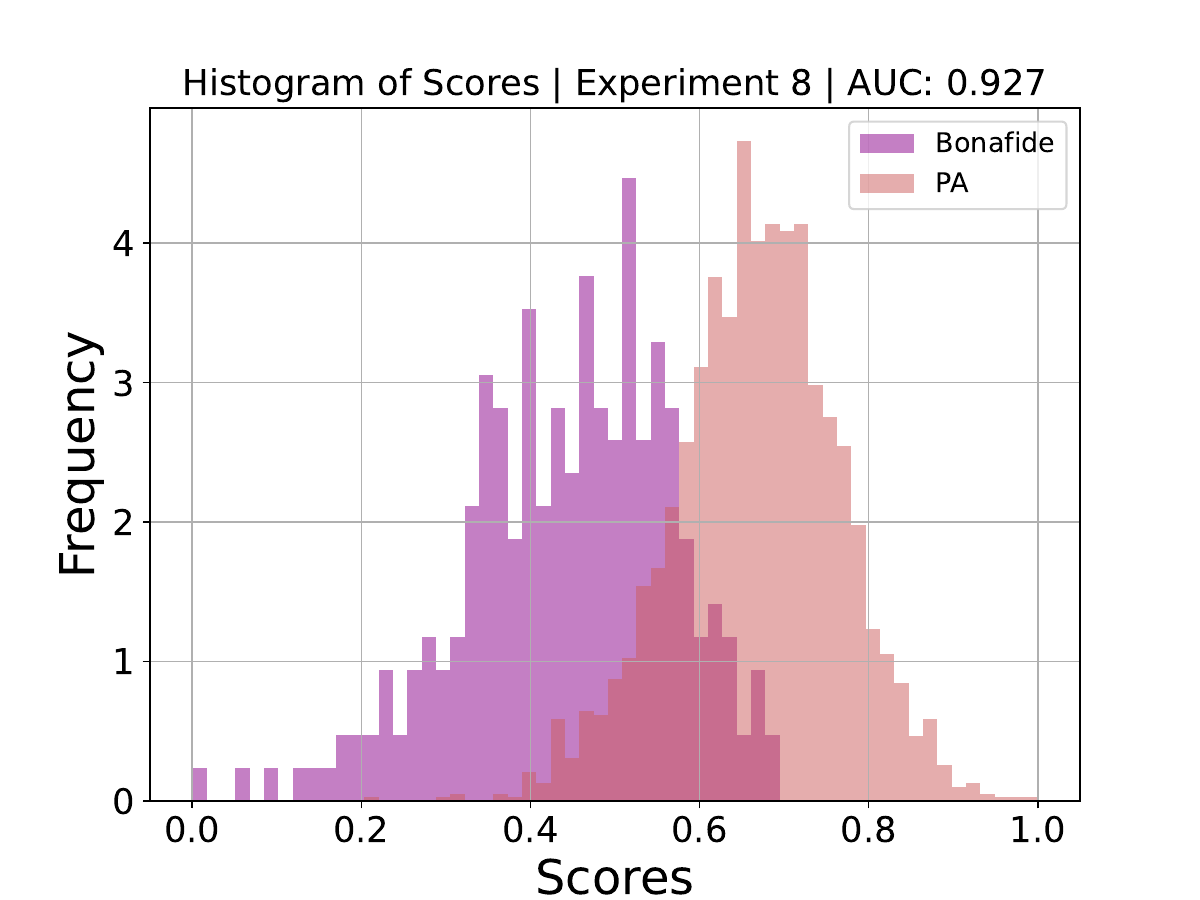}
    \end{minipage}
    \caption{\added{The presentation attack detection score distributions obtained for all eight {\bf post-mortem detection experiments} described in Section \ref{sec:pad}. Experiments 1 \& 2: Pre-trained model tested on Warsaw and \datasetname~without fine-tuning; Experiments 3 \& 4: Fine-tuned model tested cross-dataset (Warsaw and \datasetname); Experiments 5 \& 6: Model fine-tuned with 50 samples and tested cross-dataset (Warsaw and \datasetname); Experiments 7 \& 8: Model fine-tuned with 5 samples and tested cross-dataset (Warsaw and \datasetname)}.}
    \label{fig:all-8-experiments}
\end{figure*}

Fig. \ref{fig:all-8-experiments} contains the distributions of presentation attack detection scores obtained for the D-NetPAD model in all eight experiments. The {\bf first two experiments} revealed that the pre-trained D-NetPAD model, without prior exposure to post-mortem iris images, struggled to deliver favorable outcomes, underscoring the model’s initial lack of training on post-mortem images.

As can be seen in Fig. \ref{fig:all-8-experiments}, \textbf{Experiments 3 and 4}, when the model was trained on the Warsaw combined dataset and tested on the \datasetname~dataset, and vice versa, we observed significantly improved performance, as evidenced by a \added{good} separation between bona fide and presentation attack (PA) scores. Notably, the model demonstrated a more confident separation of scores when \added{fine-tuned using the Warsaw combined dataset, possibly due to higher-quality images, better consistency in capture conditions, or less noise/variation compared to \datasetname~dataset.}

\added{The histograms obtained in \textbf{Experiments 5 and 6}, where the D-NetPAD model was fine-tuned with only 50 post-mortem iris samples, continue to illustrate a clear separation between the distributions of {bona fide} and presentation attack scores, despite a slight drop in performance metrics. As shown in Figure~\ref{fig:all-8-experiments}, this separation remains pronounced, with AUC values of 0.995 and 0.997 for \textbf{Experiments 5 and 6}, respectively. These results demonstrate that even with limited training data, the model maintains a strong ability to distinguish genuine iris scans from post-mortem samples.}

\added{Finally, in \textbf{Experiments 7 and 8}, the D-NetPAD model was fine-tuned with only 5 post-mortem iris samples, aiming to evaluate the model's performance under extreme data scarcity. Despite the minimal training data, the model still demonstrated reasonable discriminative ability. In \textbf{Experiment 7}, where the model was trained on five post-mortem samples from {Warsaw} and tested on the full {NIJ-2018-DU-BX-0215} dataset, the AUC reached 0.936. In \textbf{Experiment 8}, where the model was trained on five post-mortem samples from {\datasetname} and tested on the {Warsaw} dataset, the AUC was 0.927. While the histograms indicate greater overlap between the bona fide and presentation attack score distributions compared to previous experiments, the model still showed a promising ability to detect presentation attacks with very limited training data, particularly in generalizing to unseen datasets.}

In an effort to explore the potential for further improvement, we extended the training to 100 epochs for the last two experiments. However, this adjustment did not yield any significant enhancement in the results, indicating that the model's performance stabilized after 50 epochs.

\begin{figure}[ht]
\centering
\small
\setlength{\tabcolsep}{2pt}
\begin{tabular}{cccc}
\parbox{0.11\textwidth}{\centering
  \includegraphics[width=\linewidth]{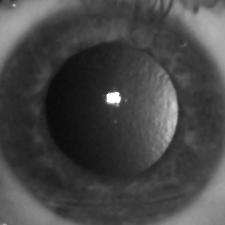}\\
  \scriptsize \added{PMI=7h \\PA=0.1008}
} &
\parbox{0.11\textwidth}{\centering
  \includegraphics[width=\linewidth]{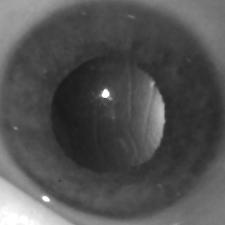}\\
  \scriptsize \added{PMI=5h\\ PA=0.1134}
} &
\parbox{0.11\textwidth}{\centering
  \includegraphics[width=\linewidth]{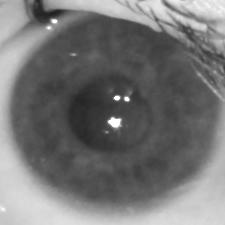}\\
  \scriptsize \added{PMI=7h \\PA=0.1159}
} &
\parbox{0.11\textwidth}{\centering
  \includegraphics[width=\linewidth]{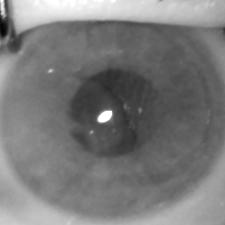}\\
  \scriptsize \added{PMI=24h\\PA=0.1867}
} \\
\parbox{0.11\textwidth}{\centering\vskip2mm
  \includegraphics[width=\linewidth]{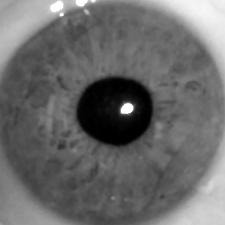}\\
  \scriptsize \added{PMI=4h\\PA=0.2030}
} &
\parbox{0.11\textwidth}{\centering\vskip2mm
  \includegraphics[width=\linewidth]{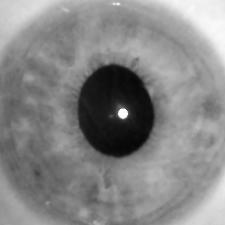}\\
  \scriptsize \added{PMI=32h\\PA=0.2215}
} &
\parbox{0.11\textwidth}{\centering\vskip2mm
  \includegraphics[width=\linewidth]{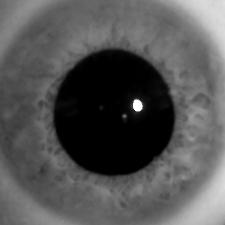}\\
  \scriptsize \added{PMI=20h\\PA=0.2579}
} &
\parbox{0.11\textwidth}{\centering\vskip2mm
  \includegraphics[width=\linewidth]{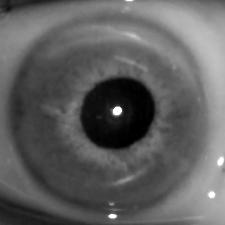}\\
  \scriptsize \added{PMI=16h\\PA=0.2866}
} \\
\end{tabular}
\caption{\added{Post-mortem iris images that were misclassified as live by the D-NetPAD model. The top row shows samples from Experiment 4 (tested on Warsaw), and the bottom row shows samples from Experiment 3 (tested on \datasetname).}}
\label{fig:pmi-pa-visualized}
\end{figure}

These findings highlight the potential of D-NetPAD, once fine-tuned, to effectively differentiate between live and post-mortem iris images across diverse datasets. The model's adaptability, evidenced by its performance in cross-dataset evaluations and minimal training sets, underscores its utility in enhancing the security measures of iris recognition systems against presentation attacks (Table \ref{tab:tdr_fdr-dnetpad}). \added{As shown in Table \ref{tab:pmi-pa-minmax}, we further explore the effect of PMI on PAD performance using Experiments 3 and 4, which involved cross-dataset evaluation on the full set of images. By comparing the PAD scores of samples with minimum and maximum PMI, we observe that higher PMI generally increases the likelihood of successful PA detection. Some failure cases can be visualized in Fig. \ref{fig:pmi-pa-visualized}.}

\begin{table}[ht]
\centering
\caption{True detection rate (1-BPCER) at APCER of 0.01\% and 1\% for Experiments 3 to 8 using D-NetPAD. BPCER stands for {\em Bona fide} Presentation Classification Error Rate, and APCER stands for Attack Presentation Classification Error Rate \cite{ISO_30107-1_2016}.}
\label{tab:tdr_fdr-dnetpad}
\begin{tabular}{l@{\hskip 6pt}c@{\hskip 6pt}c}
\toprule
\textbf{Exp.} & \textbf{1-BPCER @ 0.01\%} & \textbf{1-BPCER @ 1\%} \\
\midrule
3 & \added{98.29\%}   & \added{98.29\%} \\
4 & \added{95.29\%}   & \added{95.29\%} \\
5 & \added{90.89\%}   & \added{93.00\%} \\
6 & \added{83.74\%}   & \added{95.03\%} \\
7 & \added{43.04\%}   & \added{52.22\%} \\
8 & \added{41.84\%}   & \added{49.78\%} \\
\bottomrule
\end{tabular}
\end{table}

\begin{table}[ht]
\small
\centering
\caption{\added{Minimum and maximum PMI and the corresponding presentation attack (PA) scores for each dataset in Experiments 3 and 4.}}
\label{tab:pmi-pa-minmax}
\begin{tabular}{lcccc}
\toprule
\textbf{Dataset} & \textbf{Min PMI(h)} & \textbf{PA (Min)} & \textbf{Max PMI(h)} & \textbf{PA (Max)} \\
\midrule
Warsaw    & 5.0     & 0.3046  & 814.0 & 0.4247 \\
NIJ       & 1.0     & 0.6770    & 1674.0 & 0.7583 \\
\bottomrule
\end{tabular}
\end{table}

\section{Forensic Examiner-Supporting Tool}

One of the practical goals of this study was to create a tool to provide support to forensic iris examiners in the tasks of analysis, verification and identification of forensic iris images, which -- in addition to matching results and candidate lists -- also provides an explanation of the decisions made.

\begin{figure*}[!htb]
    \centering
    \includegraphics[height=0.21\linewidth]{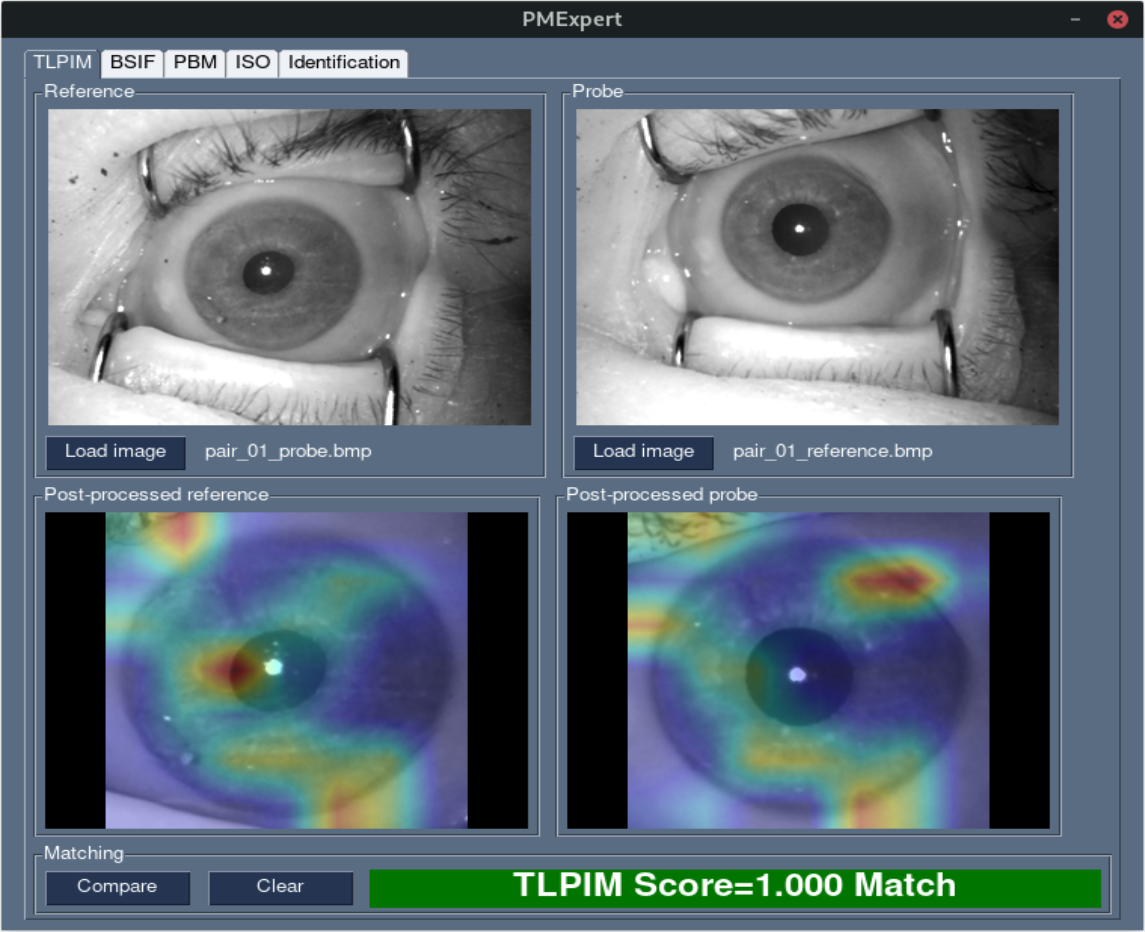}\hskip1mm
    \includegraphics[height=0.21\linewidth]{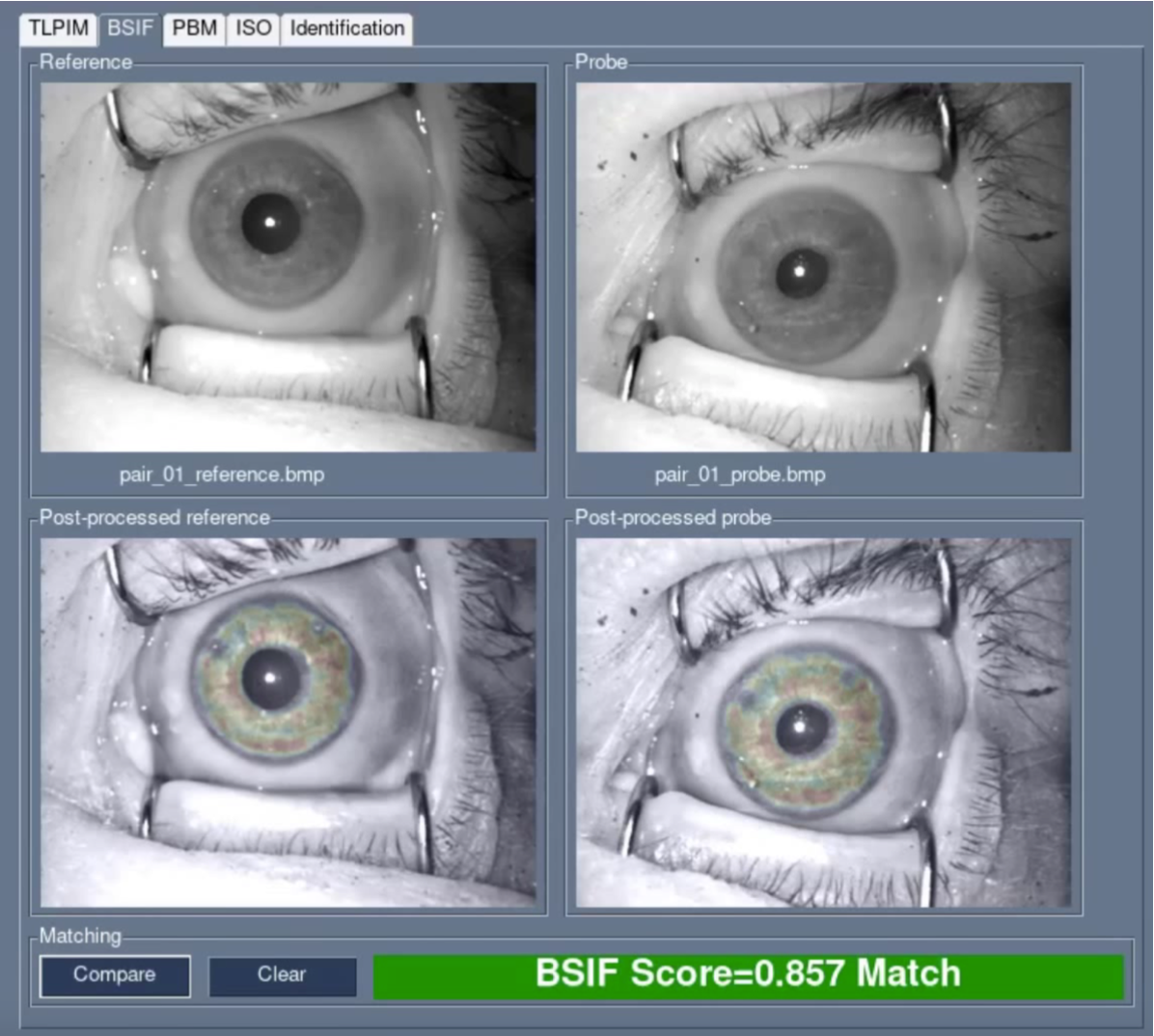}\hskip1mm
    \includegraphics[height=0.21\linewidth]{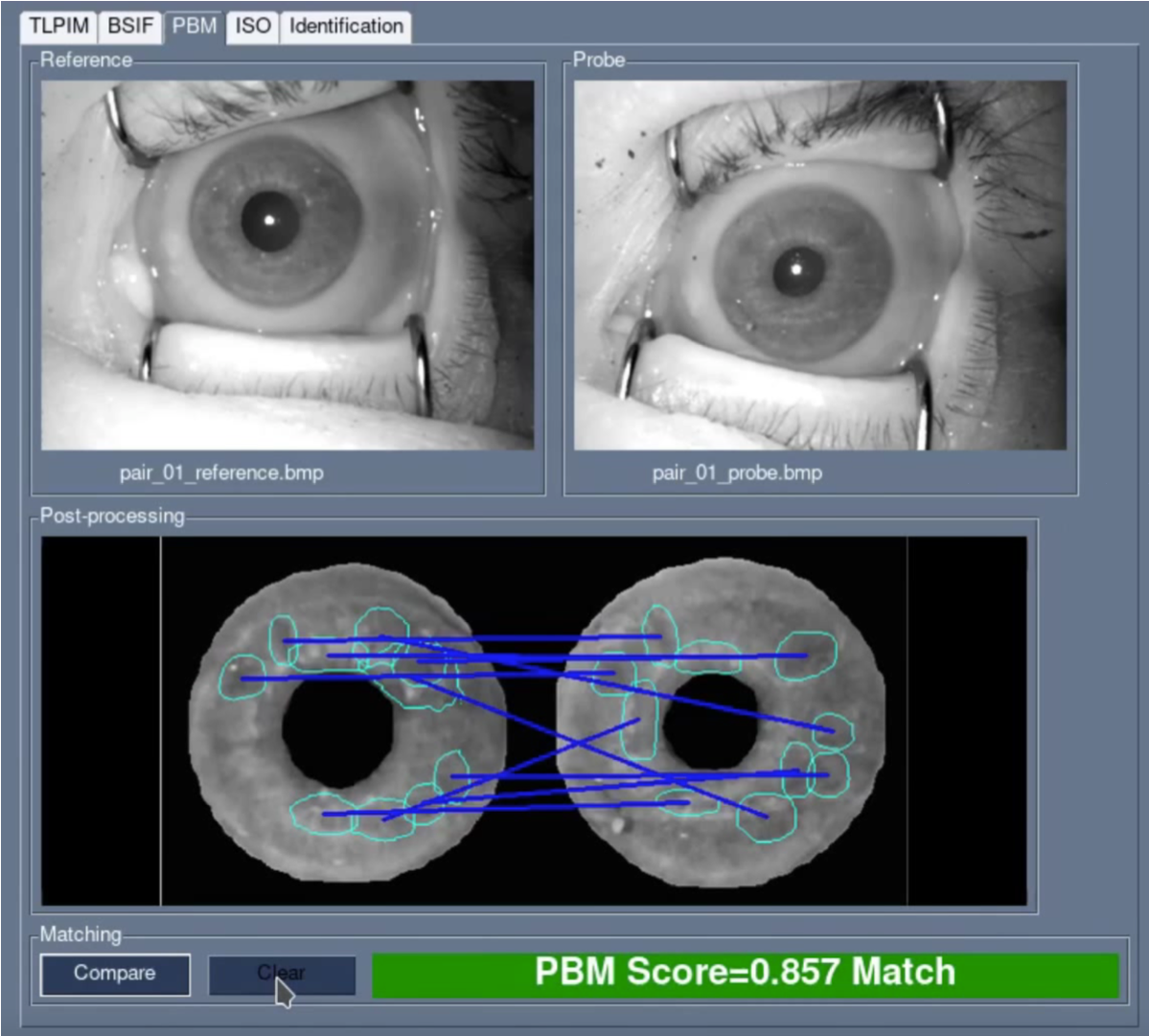}\hskip1mm
    \includegraphics[height=0.21\linewidth]{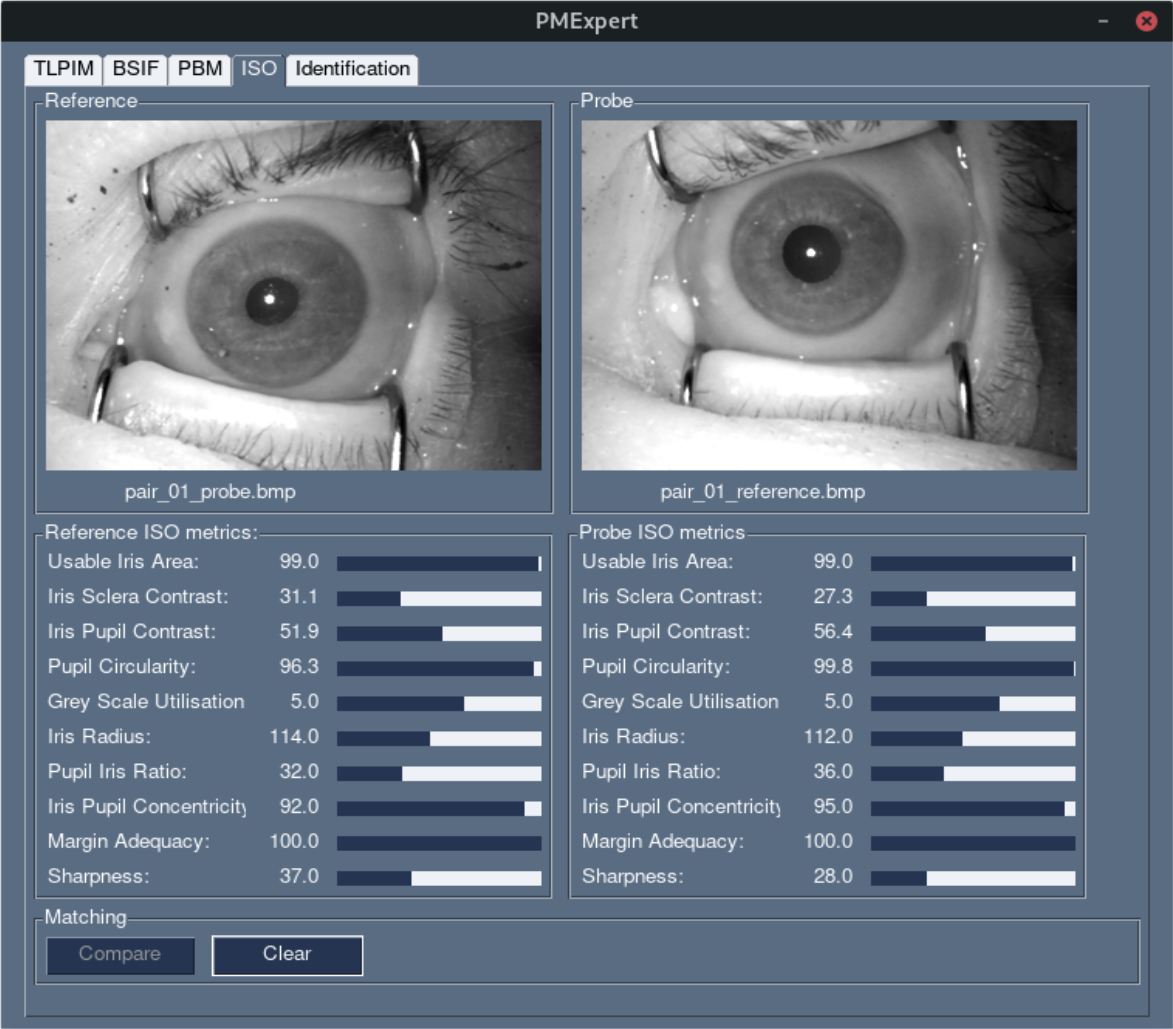}
    \cprotect\caption{Four screens of the \verb|PMExpert| tool illustrating how the forensic experts can evaluate matching results using different post mortem-specific iris recognition methods along with their visual explanations (from left to right: TLPIM, HDBIF and PBM methods), as well as get the ISO metrics for each image (the upmost right screen).}
    \label{fig:pmexpert}
\end{figure*}

The designed and implemented tool, \verb+PMExpert+, is an open-source software, \added{it was delivered to and integrated into the forensic workflow at the DCMEO, and has been used by a real forensic expert conducting routine postmortem subjects’ examinations since May 2021. The DCMEO experts concluded that the use of the iris capture tool and the matching software was easy to learn and deploy. In the morgue, scans were performed both prior to and after completion of postmortem examinations (autopsies) by autopsy technician or physician with comparable results.} \verb+PMExpert+ implements the most successful methods for postmortem iris identification at the time of its creation: 

\begin{enumerate}
    \item three post-mortem iris-specific recognition methods: 
    \begin{enumerate}
        \item Triplet Loss Postmortem Iris Matching (TLPIM) \cite{kuehlkamp2022interpretable}, with class activation maps to visualize the salient iris regions used by the method in matching,
        \item Patch-Based Matching (PBM) \cite{Boyd_WACVW_2023}, visualizing the localized and matched irregular iris regions, and
        \item Human-Driven Binarized Image Features (HDBIF) \cite{czajka2019domain}, complemented with a visualization capability added in this study that highlights the most similar regions of the iris texture;
    \end{enumerate}
    \item identification feature, maintaining a local repository of image templates, to facilitate searches for the ``best matches'' against an unknown query image;
    \item ISO/IEC 29794-6 iris image quality metrics, offering examiners critical data points for evaluating image quality and determining usability of the examined samples.
\end{enumerate}

\verb+PMExpert+ is designed with two primary components: a command line interface (CLI) and a graphical user interface (GUI, illustrated in Fig. \ref{fig:pmexpert}). The CLI caters to users who prefer a scriptable environment, while the GUI streamlines the process for technicians, allowing rapid comparison of image pairs using multiple techniques to evaluate and compare iris comparison scores, candidate lists and image quality metrics. 

From a software engineering standpoint, \verb+PMExpert+ is structured into a front-end module comprising the CLI and GUI that operate natively across multiple platforms, including Windows, Linux, and Mac OS, and a back-end module housed within a Docker container. The front-end is responsible for user interface and file handling, whereas the back-end manages the complex image processing tasks.

\section{Summary}

\paragraph{Conclusions} Post-mortem biometric recognition, including iris recognition, has been, to date, restricted to the domain of professional forensic lab examination due to exceptionally difficult data collection. This has resulted in limited research on automatic iris recognition as applied to deceased subjects. This study makes important contributions to accelerate these research efforts: (a) the dataset acquired from 259 deceased subjects with several unique properties, such as samples collected before and after death from the same subject, and the largest post-mortem interval of almost 70 days, (b) experiments with few-shot deep learning demonstrating a possibility to detect cadaver eye images as presentation attacks, (c) an open-source forensic iris examination tool implementing three different post-mortem iris recognition methods, and (d) post-mortem iris recognition results, broken down by demographic and biological factors to provide the most recent assessment of forensic iris recognition capabilities.

\paragraph{Dataset and software availability} This study follows good practices in making the research results fully replicable. Thus, except for the commercial VeriEye matcher, all other methods used in this work are open-sourced, and both the newly-collected \datasetname~dataset as well as the designed forensic examination tool \verb+PMExpert+ are available at no cost via the National Archive of Criminal Justice Data (NACJD)~\cite{Czajka_NACJD_2023}.

\section*{Acknowledgment}

\added{This work was supported by the National Institute of Justice, Office of Justice Programs, U.S. Department of Justice, under Award 2018-DU-BX-0215. The opinions, findings, and conclusions or recommendations expressed in this publication are those of the authors and do not necessarily reflect those of the Department of Justice.}

\bibliography{ref.bib}{}
\bibliographystyle{IEEEtran}

\section{Biography Section}

\vspace{-33pt}
\begin{IEEEbiography}[{\includegraphics[width=1in,height=1.25in,clip,keepaspectratio]{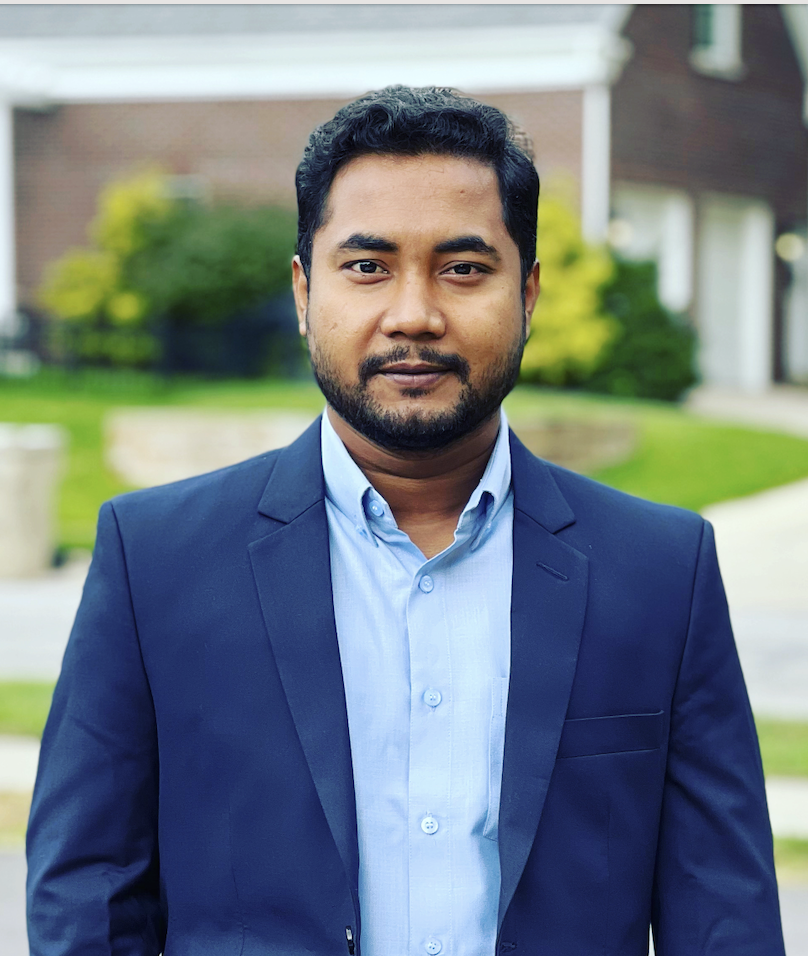}}]{Rasel Ahmed Bhuiyan} (Student Member, IEEE) is a Ph.D. candidate in the Department of Computer Science and Engineering at the University of Notre Dame, where he conducts research under the guidance of Prof. Adam Czajka in the Computer Vision Research Laboratory (CVRL). His research focuses on advancing machine learning and computer vision techniques to address complex challenges in human biometrics, particularly in iris recognition under extreme conditions such as post-mortem and infant scenarios, with applications in identity verification and security. He holds an M.S. in Computer Science and Engineering from the University of Notre Dame and a B.Sc. in Computer Science and Engineering from the University of Asia Pacific.
\end{IEEEbiography}

\vspace{-33pt}
\begin{IEEEbiography}[{\includegraphics[width=1in,height=1.25in,clip,keepaspectratio]{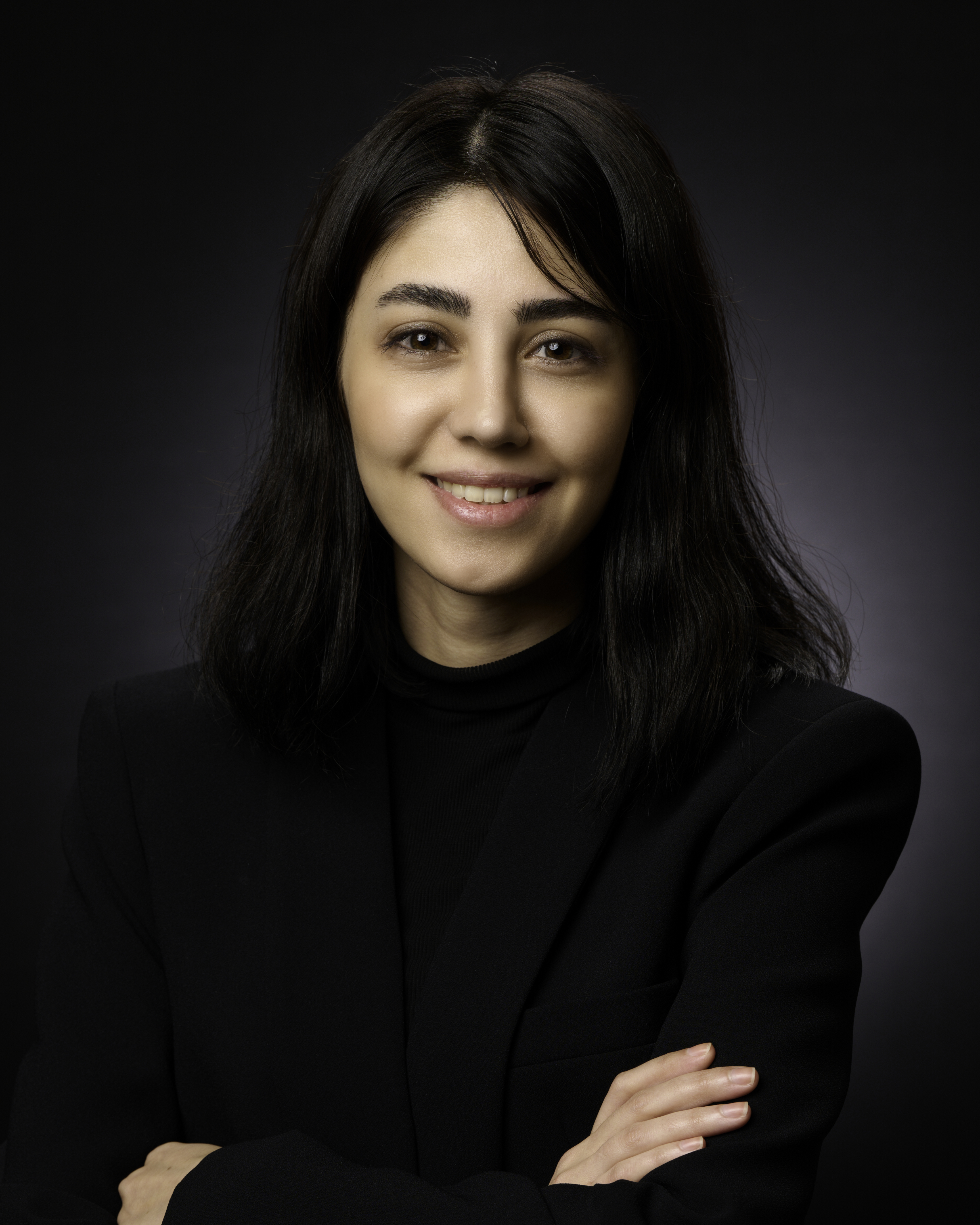}}]{Parisa Farmanifard} is a Ph.D. student in the Department of Computer Science and Engineering at Michigan State University, working in the iPRoBe Lab under the supervision of Dr. Arun Ross. She also completed her M.S. in Computer Science at MSU. Her research focuses on biometrics, with an emphasis on iris recognition, presentation attack detection, and the use of foundation models (FMs) and large language models (LLMs) in biometrics.
\end{IEEEbiography}

\vspace{-33pt}
\begin{IEEEbiographynophoto}{Renu Sharma} completed her PhD in the Computer Science and Engineering Department at Michigan State University (MSU) in 2022. Currently, she is with Amazon as Applied Scientist. Renu also has 8 years of experience in Research \& Development at CDAC Mumbai, India.
\end{IEEEbiographynophoto}

\vspace{-33pt}
\begin{IEEEbiography}[{\includegraphics[width=1in,height=1.25in,clip,keepaspectratio]{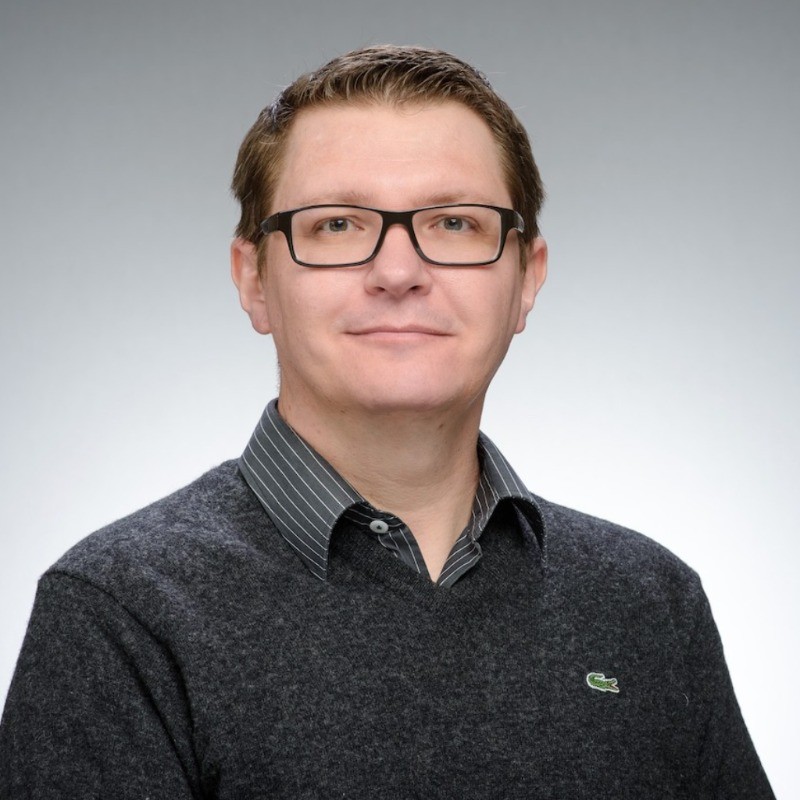}}]{Andrey Kuehlkamp} is currently a Postdoctoral Research Associate in the Center for Research Computing at the University of Notre Dame, where he also obtained his PhD in Computer Science and Engineering. He has experience as researcher, developer and also as a teacher in Computer Science. His interests are Biometrics, Distributed Systems, Computer Vision and Machine Learning.
\end{IEEEbiography}

\vspace{-33pt}
\begin{IEEEbiography}[{\includegraphics[width=1in,height=1.25in,clip,keepaspectratio]{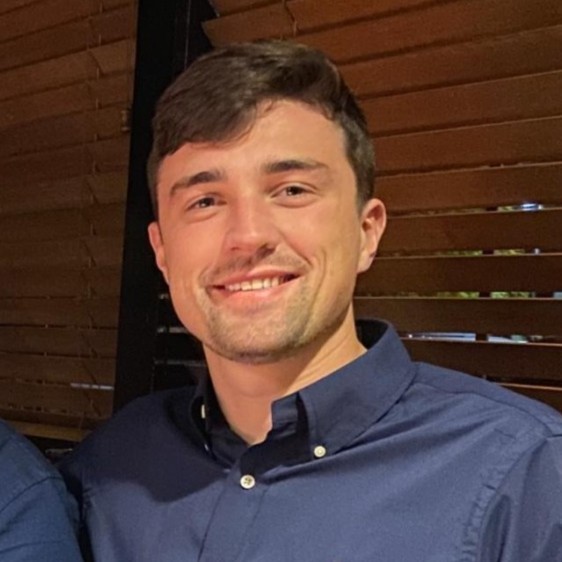}}]{Aidan Boyd} received his Ph.D from the University of Notre Dame in 2023 and is currently with Meta, New York, NY. He is passionate about the advancement of computer vision technology using artificial intelligence and is especially interested in leveraging human decision making to improve robustness and trustworthiness of AI models.
\end{IEEEbiography}

\vspace{-33pt}
\begin{IEEEbiography}[{\includegraphics[width=1in,height=1.25in,clip,keepaspectratio]{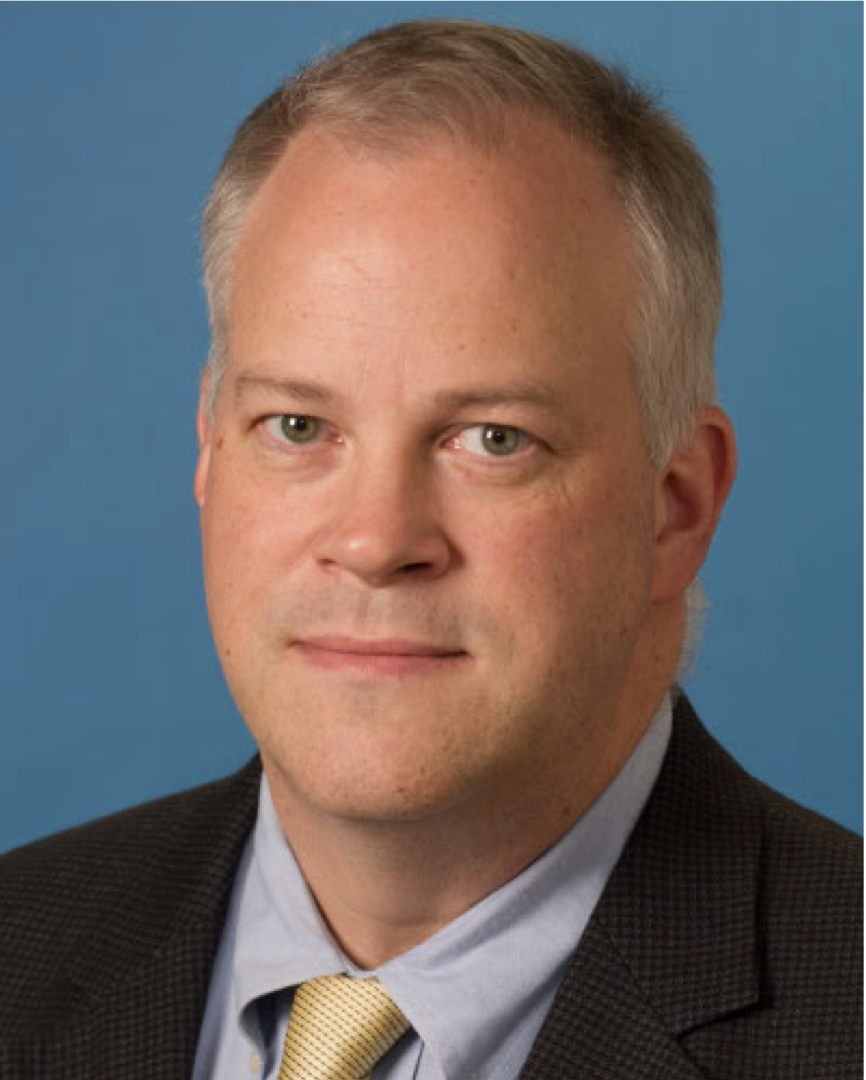}}]{Patrick J. Flynn} (Fellow, IEEE) received the Ph.D. degree in computer science from Michigan State University, East Lansing, MI, USA, in 1990. He is the Fritz Duda Family Professor of Engineering at the University of Notre Dame, Notre Dame, IN, USA. He has also held faculty positions at Washington State University, Pullman, WA, USA, and Ohio State University, Columbus, OH, USA. His research interests include computer vision, biometrics, and image processing. Dr. Flynn is a fellow of IAPR. He is a past Editor-in-Chief of IEEE Biometrics Compendium, a past Associate Editor-in-Chief of IEEE Transactions on PAMI, and a past Associate Editor of IEEE Transactions on Image Processing and IEEE Transactions on Information Forensics and Security.
\end{IEEEbiography}

\vspace{-33pt}
\begin{IEEEbiography}[{\includegraphics[width=1in,height=1.25in,clip,keepaspectratio]{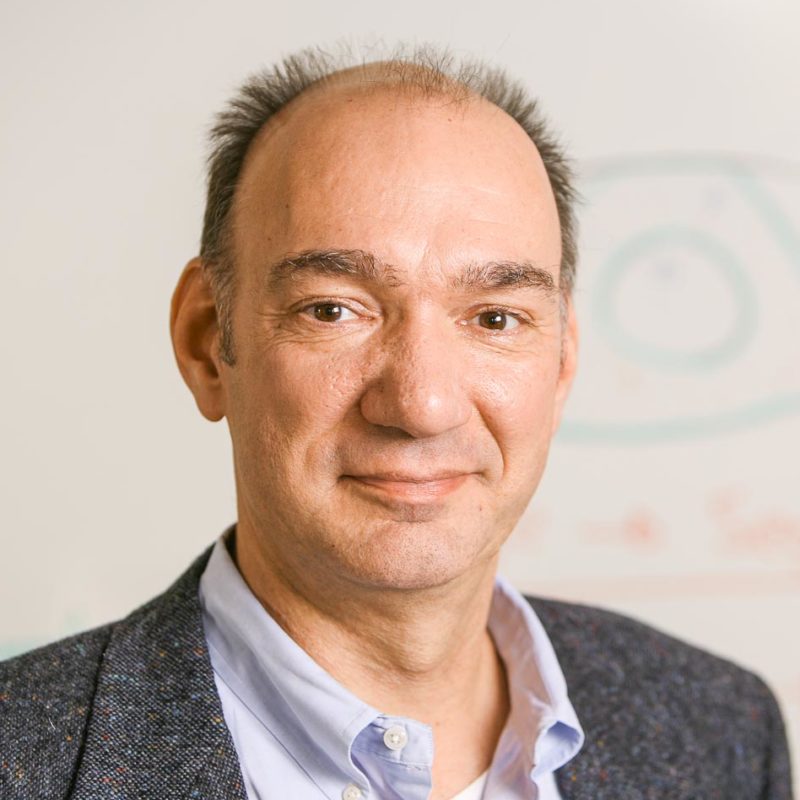}}]{Kevin Bowyer} (Fellow, IEEE) is the Schubmehl-Prein Family Professor of computer science and engineering with the University of Notre Dame. In 2019, he was elected as a Fellow of the American Association for the Advancement of Science. He is also a Fellow of the IAPR. He received the Technical Achievement Award from the IEEE Computer Society, with the citation for pioneering contributions to the science and engineering of biometrics. He has served as the Editor-in-Chief for the IEEE Transactions on Pattern Analysis and Machine Intelligence. He currently serves as the Editor-in-Chief for the IEEE Transactions on Biometrics, Behavior, and Identity Science.
\end{IEEEbiography}

\vspace{-33pt}
\begin{IEEEbiography}[{\includegraphics[width=1in,height=1.25in,clip,keepaspectratio]{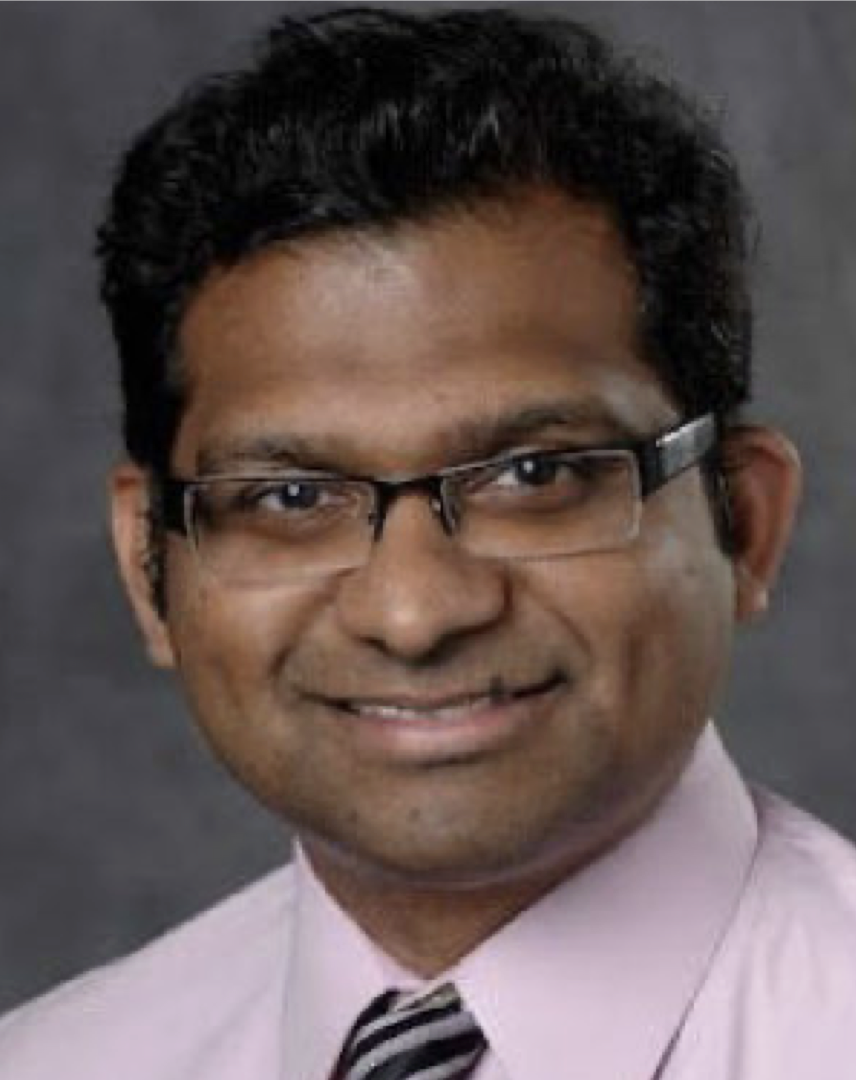}}]{Arun Ross} (Senior Member, IEEE) (Senior Member, IEEE) received the B.E. degree (Hons.) in computer science from BITS Pilani, India, and the M.S. and Ph.D. degrees in computer science and engineering from Michigan State University, USA. He is currently the Martin J. Vanderploeg Endowed Professor with Michigan State University and the Site Director of the NSF Center for Identification Technology Research. He has co-authored the textbook, Introduction to Biometrics and the monograph, Handbook of Multibiometrics. He was a recipient of the NSF CAREER Award and was designated as a Kavli Fellow by the U.S. National Academy of Sciences, in 2006. He received the J. K. Aggarwal Prize, in 2014, and the Young Biometrics Investigator Award, in 2013, from the International Association of Pattern Recognition. He has advocated for the responsible use of biometrics in multiple forums including the NATO Advanced Research Workshop on Identity and Security in Switzerland, in 2018. He testified as an Expert Panelist in an event organized by the United Nations Counter-Terrorism Committee at the UN Headquarters, in 2013. In June 2022, he testified at the U.S. House Science, Space, and Technology Committee on the topic of Biometrics and Personal Privacy.
\end{IEEEbiography}

\vspace{-33pt}
\begin{IEEEbiography}[{\includegraphics[width=1in,height=1.25in,clip,keepaspectratio]{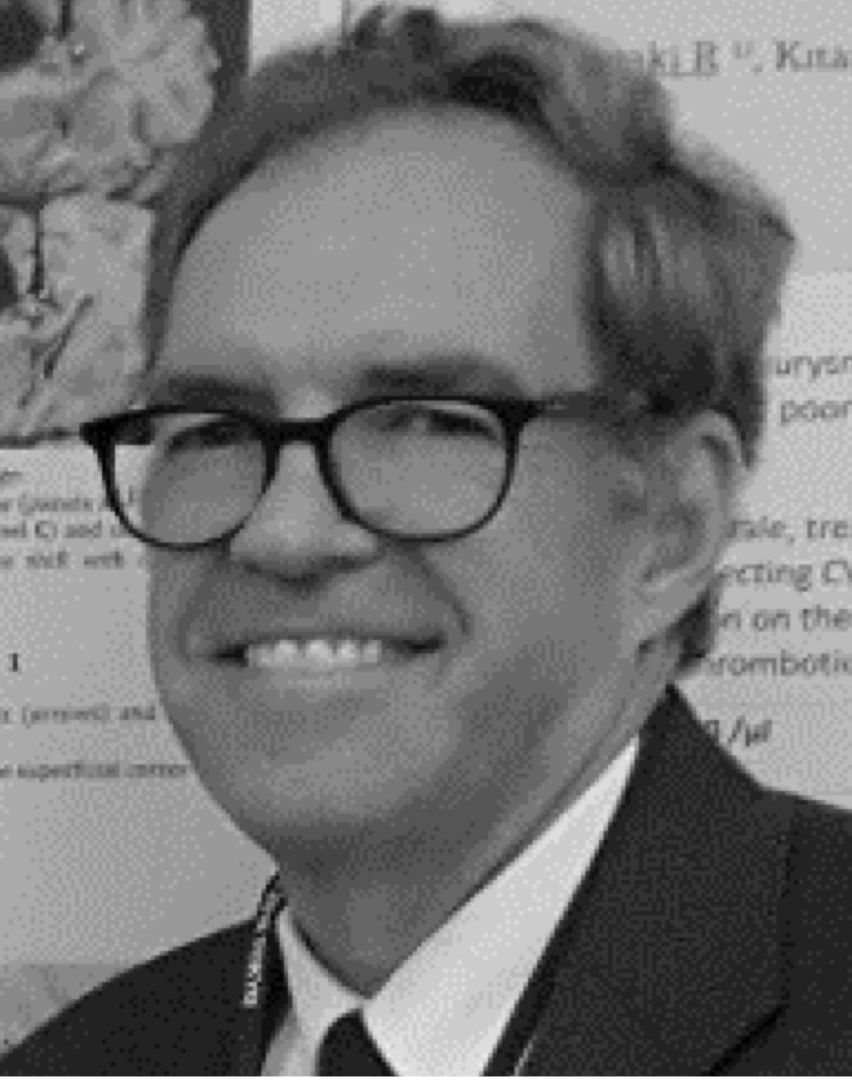}}]{Dennis Chute} received the Bachelor of Science degree from the University of Notre Dame, in 1980, and the Medical degree from the Hahnemann Medical School, Drexel University Medical School, in 1985. He completed a residency in Anatomic and Clinical Pathology with the Danbury Hospital, in 1989, and a fellowship in Forensic Pathology at the Office of the Chief Medical Examiner for the State of Maryland, Baltimore, in 1990. He was employed as an Assistant Medical Examiner for the State of Maryland, for ten years, before doing a fellowship in Neuropathology at UCLA after which he worked for two years as an Assistant Professor with the Division of Neuropathology, The David Geffen School of Medicine, UCLA. In 2005, he became the Deputy Medical Examiner for Dutchess County and, then, became the Chief Medical Examiner, in 2014. His research interests include forensic and forensic neuropathology.
\end{IEEEbiography}

\vspace{-33pt}
\begin{IEEEbiography}[{\includegraphics[width=1in,height=1.25in,clip,keepaspectratio]{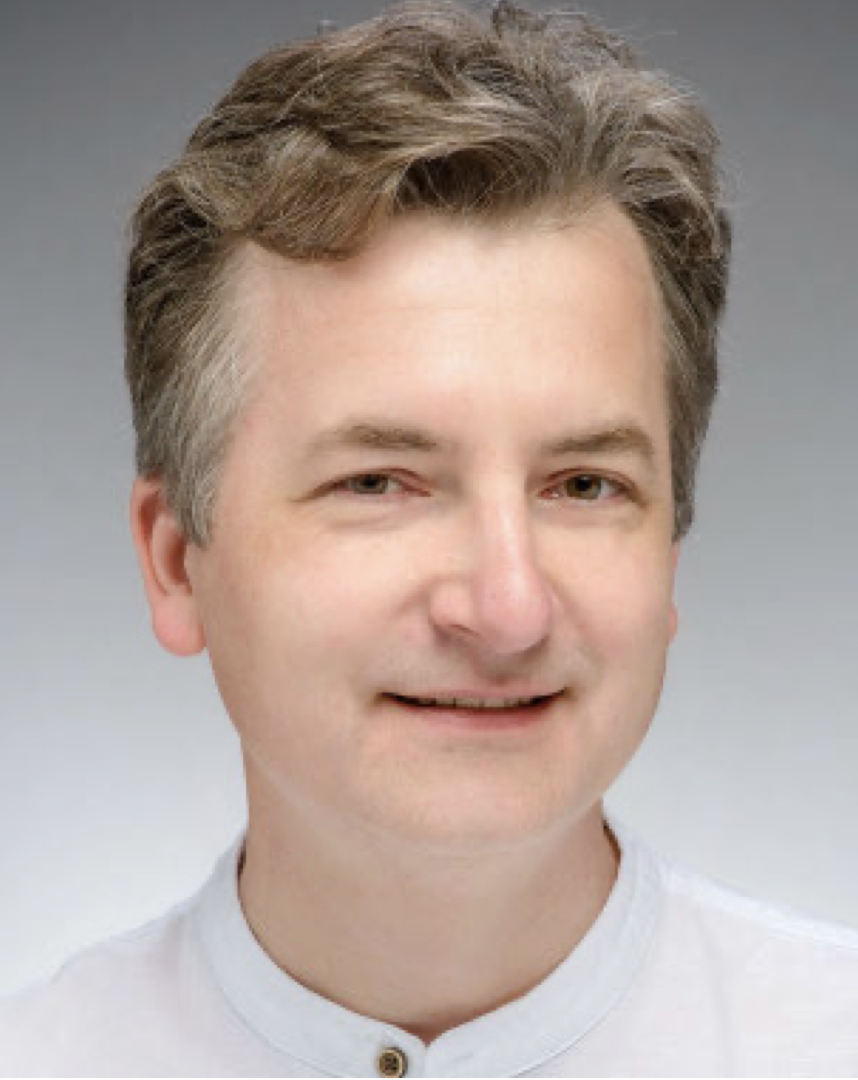}}]{Adam Czajka} (Senior Member, IEEE) is an Associate Professor at the University of Notre Dame, Notre Dame, IN, USA, where he co-directs the Computer Vision Research Lab, and directs the AI Trust and Reliability (AITAR) Lab. His research focuses on
human biometrics, especially iris recognition and methods of detecting biometric presentation attacks. In general, Dr. Czajka is interested in a wide spectrum of research in computer vision, pattern recognition, and machine learning, and the non-obvious intersections with psychology, medical sciences, and art. Dr. Czajka is a recipient of the NSF CAREER Award. He serves as the Associate Editor of the IEEE Transactions on Pattern Analysis and Machine Intelligence, is past Senior Associate Editor of the IEEE Transactions on Information Forensics and Security, and past Associate Editor of the IEEE Access and the IEEE Biometrics Compendium.
\end{IEEEbiography}

\onecolumn

\begin{center}
    \Huge{Beyond Mortality: Advancements in Post-Mortem Iris Recognition through Data Collection and Computer-Aided Forensic Examination}\vskip3mm 
    \large{Supplementary Materials}
\end{center}

\vskip3mm\noindent
These Supplementary Materials offer heatmaps illustrating selected ISO/IEC 29794-6 quality metrics (\verb+USABLE_IRIS_AREA+, \verb+IRIS_SCLERA_CONTRAST+, \verb+GRAY_SCALE_UTILIZATION+, \verb+IRIS_RADIUS+, \verb+PUPIL_IRIS_RATIO+, \verb+IRIS_PUPIL_CONCENTRICITY+, and \verb+SHARPNESS+) calculated for iris image pairs in various matching experiments. Since such matching incorporates two images, and hence two quality scores $f_1$ and $f_2$ are available, the presented heatmaps combine the average quality score $\frac{f_1+f_2}{2}$ and the difference $f_1-f_2$ between quality scores.

\twocolumn

\begin{figure}[hbt!]
    \begin{minipage}[c]{3.45in}
        \centering
        \subfloat[USABLE\_IRIS\_AREA\label{1a:sex-heatmap}]{%
           \includegraphics[width=0.50\linewidth]{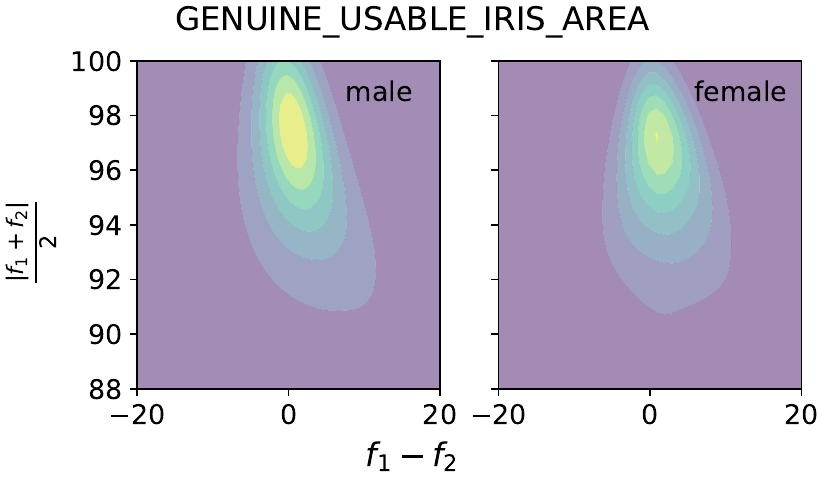}
           \includegraphics[width=0.50\linewidth]{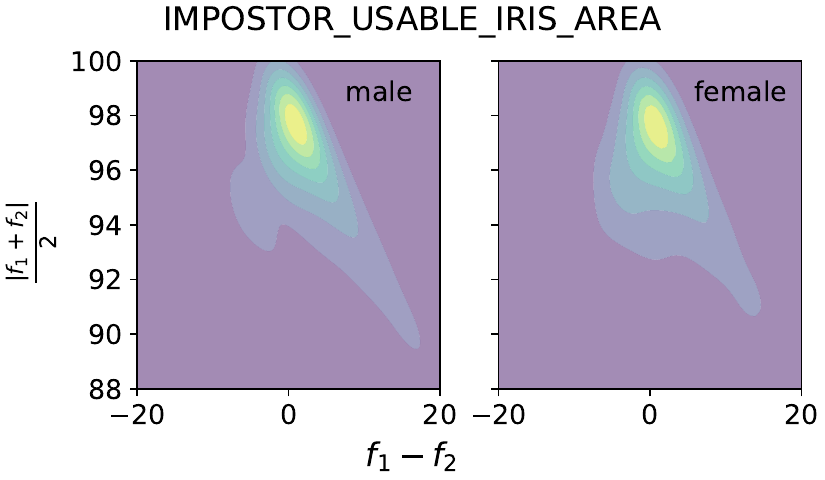}}
        \\
        \subfloat[IRIS\_SCLERA\_CONTRAST\label{1b:sex-heatmap}]{%
            \includegraphics[width=0.50\linewidth]{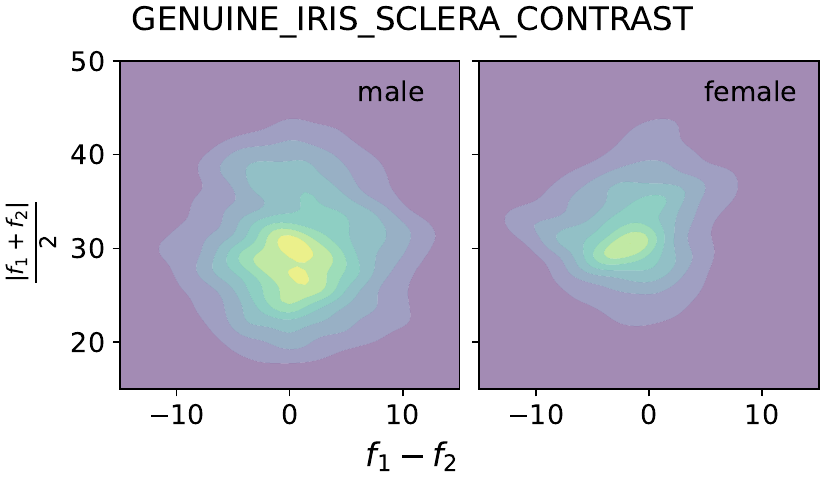}
           \includegraphics[width=0.50\linewidth]{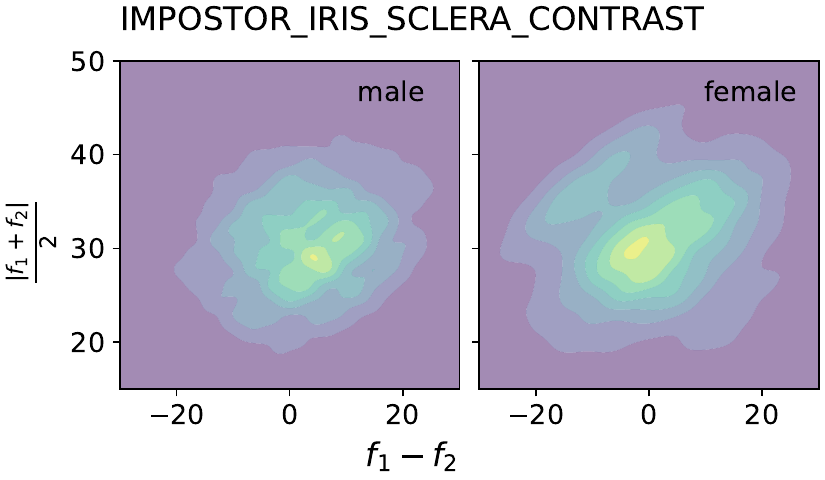}}
        \\
        \subfloat[GREY\_SCALE\_UTILISATION\label{1d:sex-heatmap}]{%
            \includegraphics[width=0.50\linewidth]{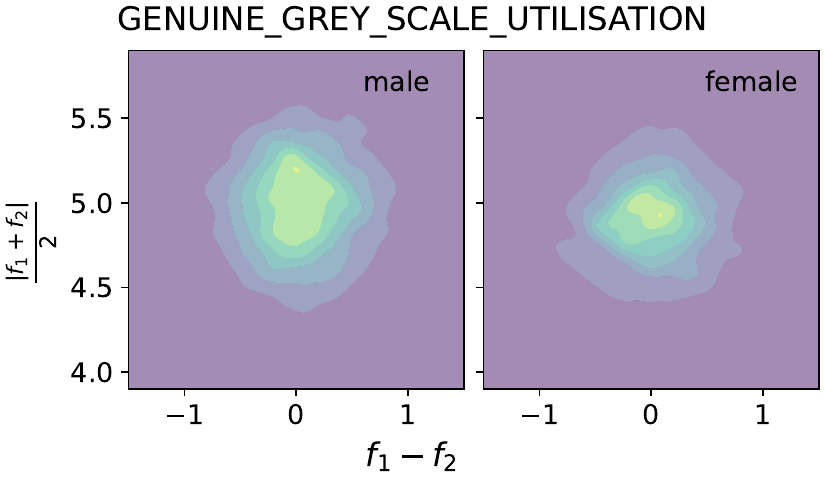}
           \includegraphics[width=0.50\linewidth]{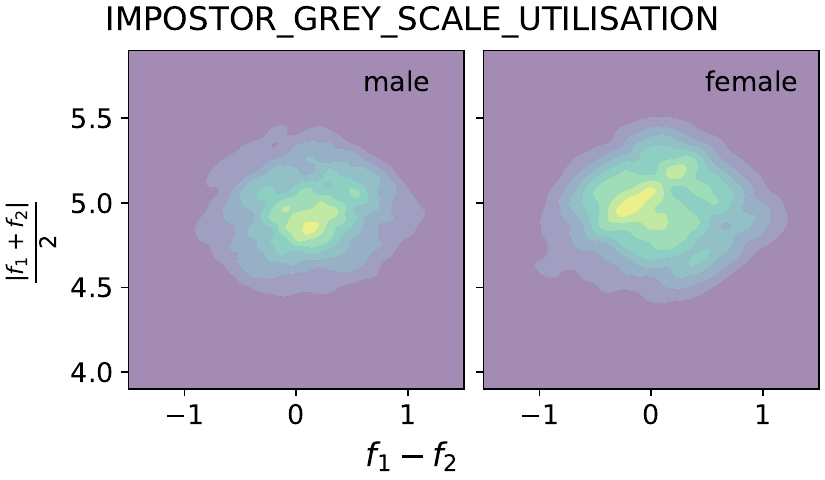}}
        \\
       \subfloat[IRIS\_RADIUS\label{1e:sex-heatmap}]{%
           \includegraphics[width=0.50\linewidth]{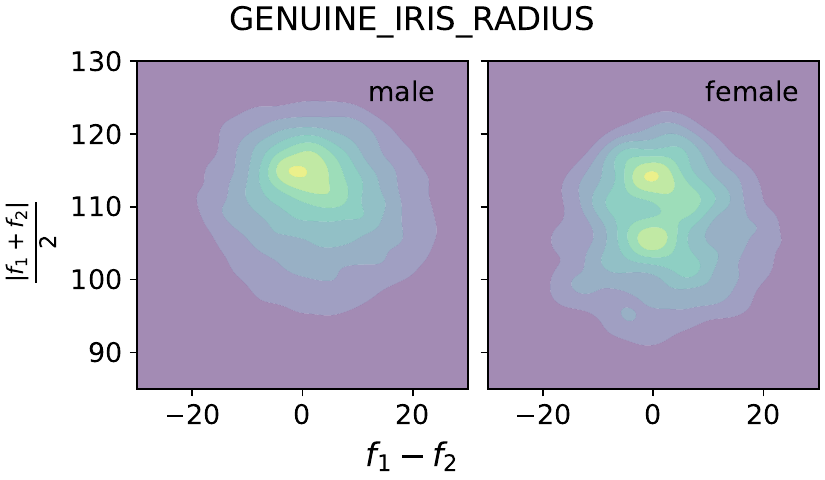}
           \includegraphics[width=0.50\linewidth]{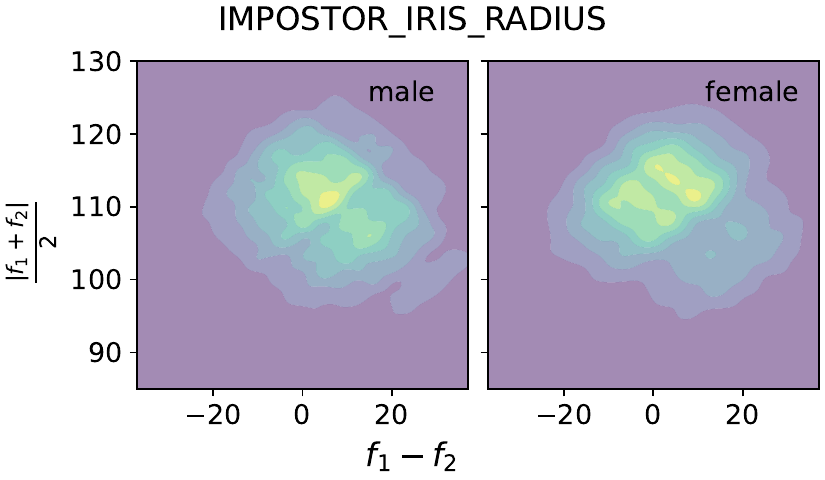}}
        \\
        \subfloat[PUPIL\_IRIS\_RATIO\label{1f:sex-heatmap}]{%
           \includegraphics[width=0.50\linewidth]{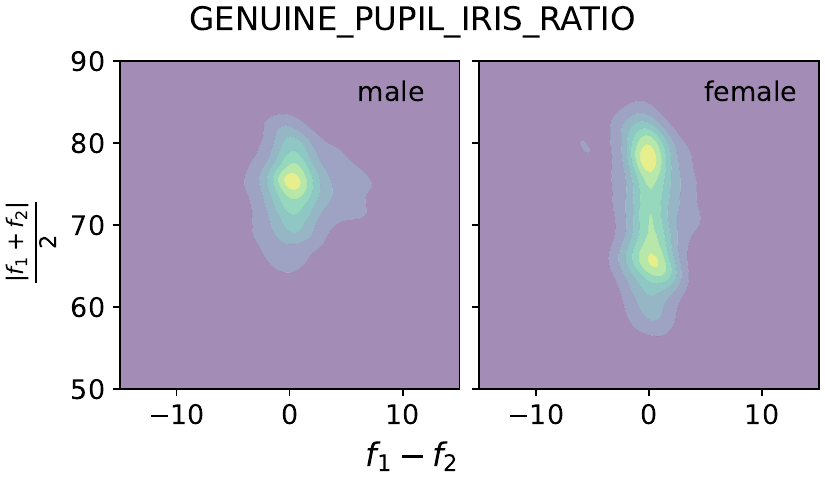}
           \includegraphics[width=0.50\linewidth]{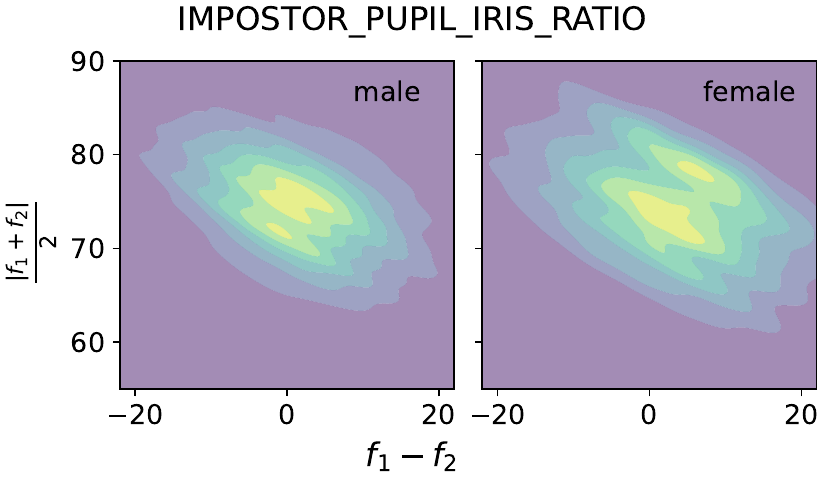}}
        \\
        \subfloat[IRIS\_PUPIL\_CONCENTRICITY\label{1g:sex-heatmap}]{%
           \includegraphics[width=0.50\linewidth]{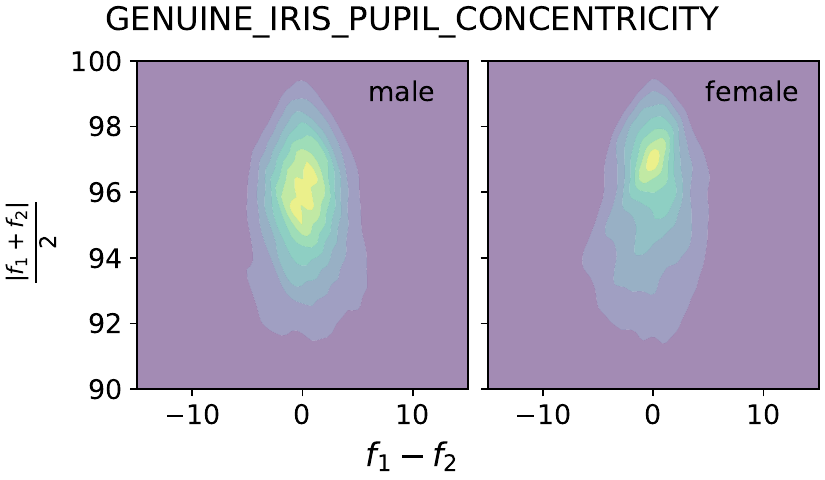}
           \includegraphics[width=0.50\linewidth]{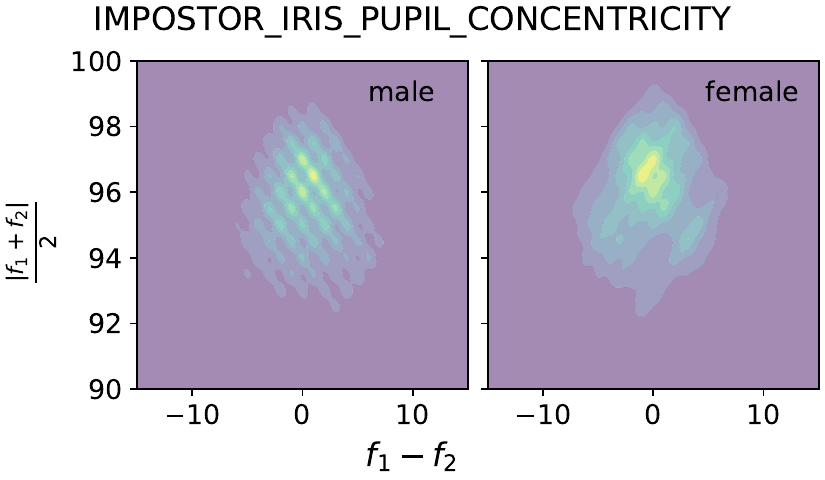}}
        \\
        \subfloat[SHARPNESS\label{1h:sex-heatmap}]{%
            \includegraphics[width=0.50\linewidth]{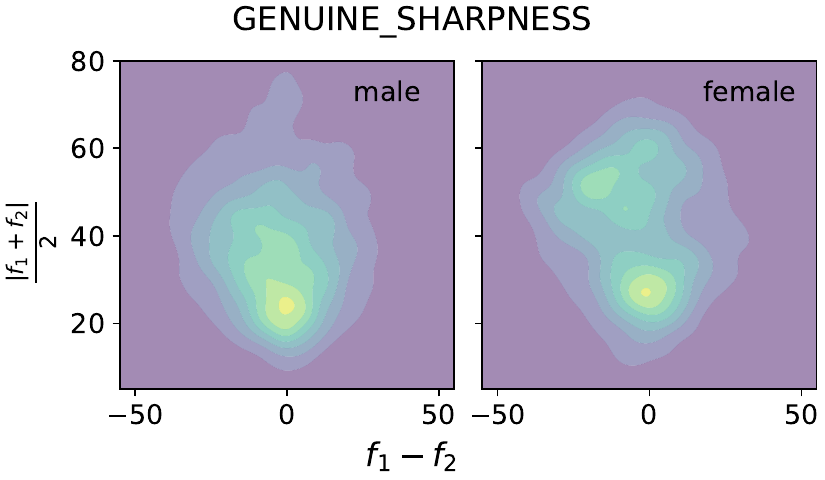}
            \includegraphics[width=0.50\linewidth]{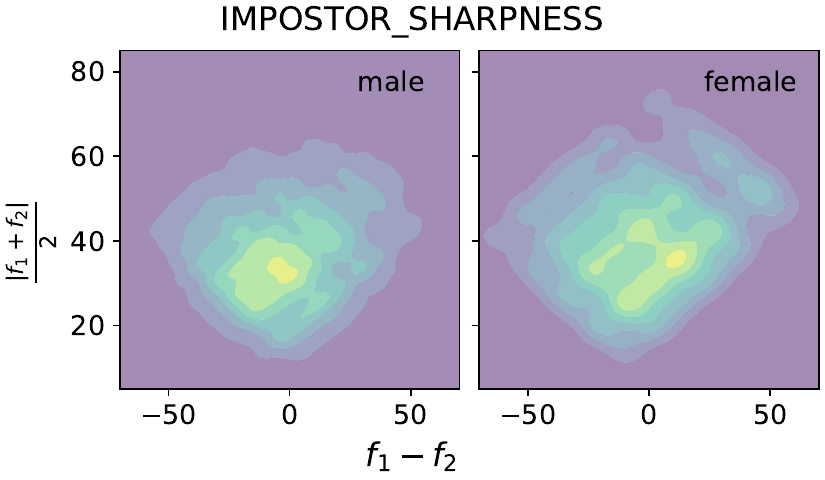}}
      \caption{ISO/IEC 29794-6 quality metrics density heatmaps split by gender and type of comparison (genuine and impostor) for the combined \datasetname~and Warsaw datasets.}
      \label{fig:sex-heatmap} 
    \end{minipage}
\end{figure}

\begin{figure} [hbt!]
\begin{minipage}[c]{3.45in}
    \centering
    \subfloat[USABLE\_IRIS\_AREA\label{1a:age-heatmap}]{%
       \includegraphics[width=0.50\textwidth]{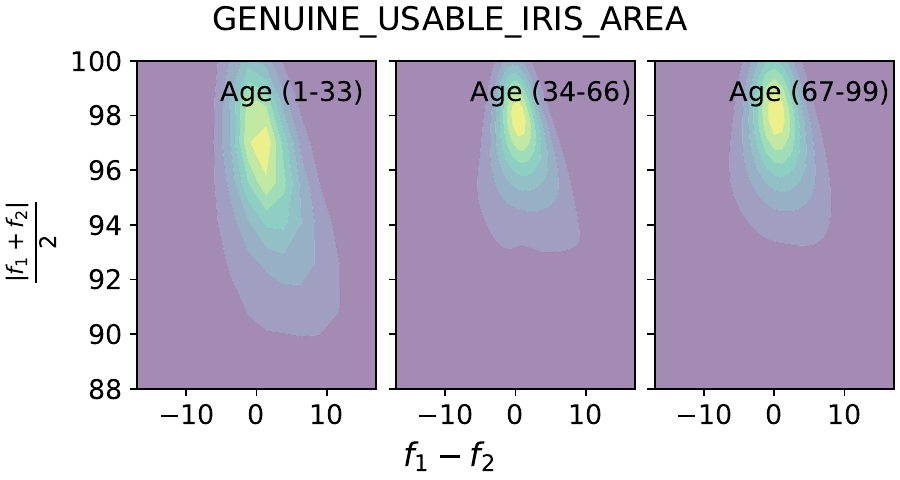}
       \includegraphics[width=0.50\textwidth]{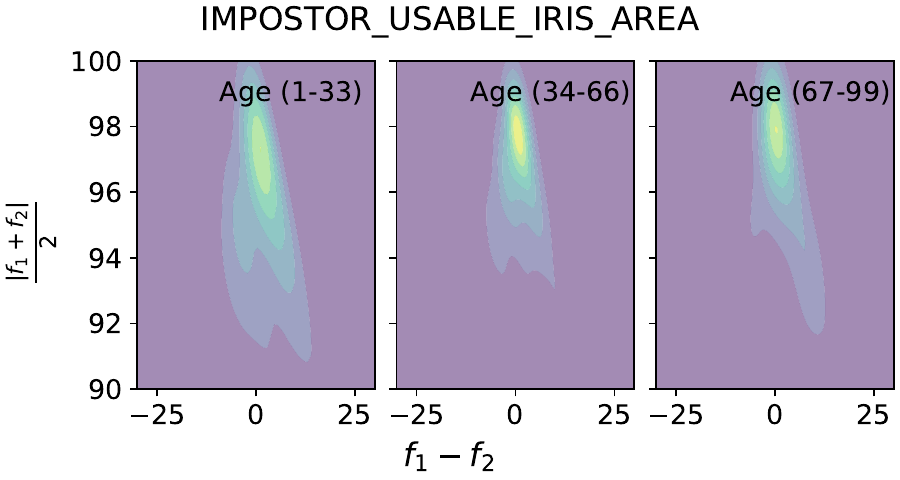}}
    \\
    \subfloat[IRIS\_SCLERA\_CONTRAST\label{1b:age-heatmap}]{%
        \includegraphics[width=0.50\textwidth]{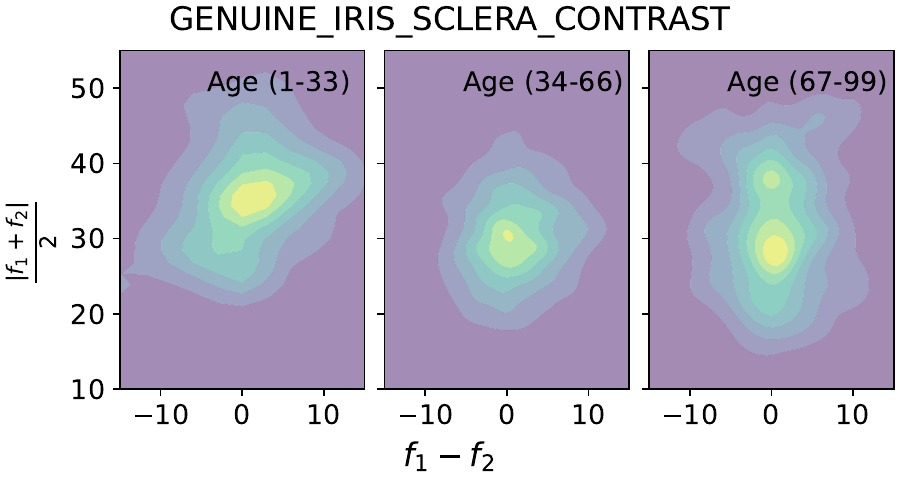}
       \includegraphics[width=0.50\textwidth]{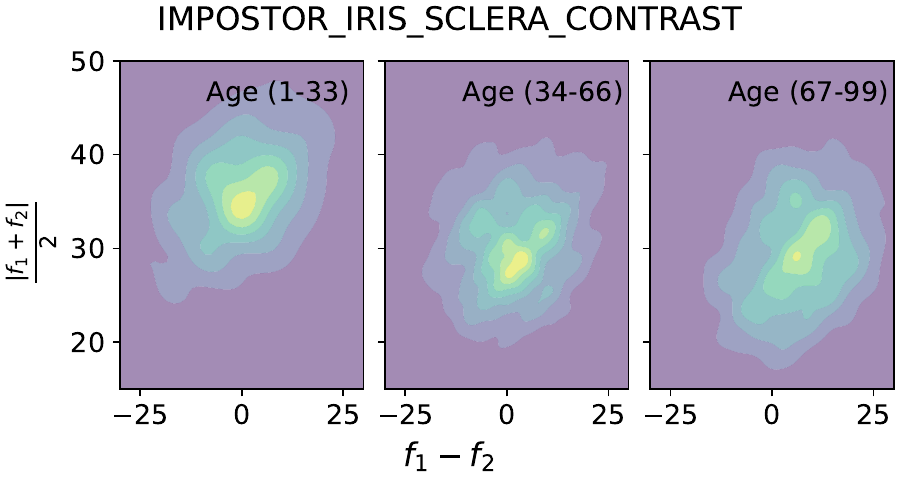}}
    \\
    \subfloat[GREY\_SCALE\_UTILISATION\label{1d:age-heatmap}]{%
        \includegraphics[width=0.50\textwidth]{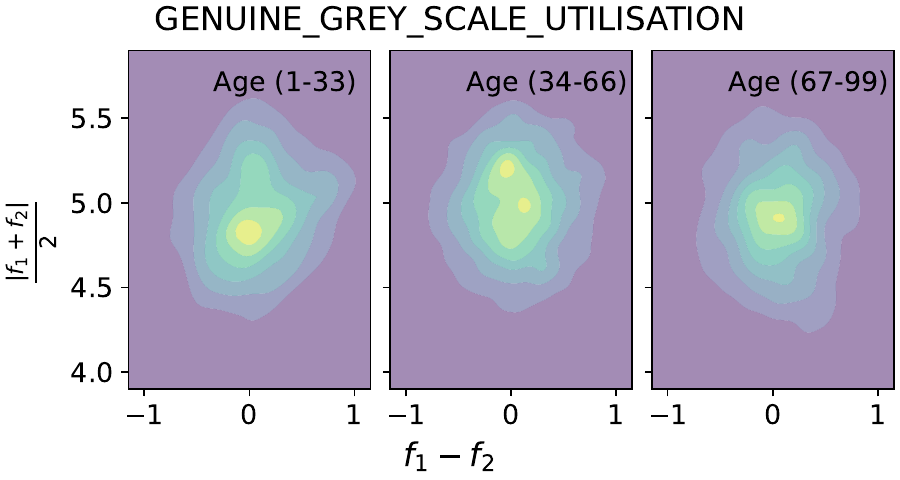}
       \includegraphics[width=0.50\textwidth]{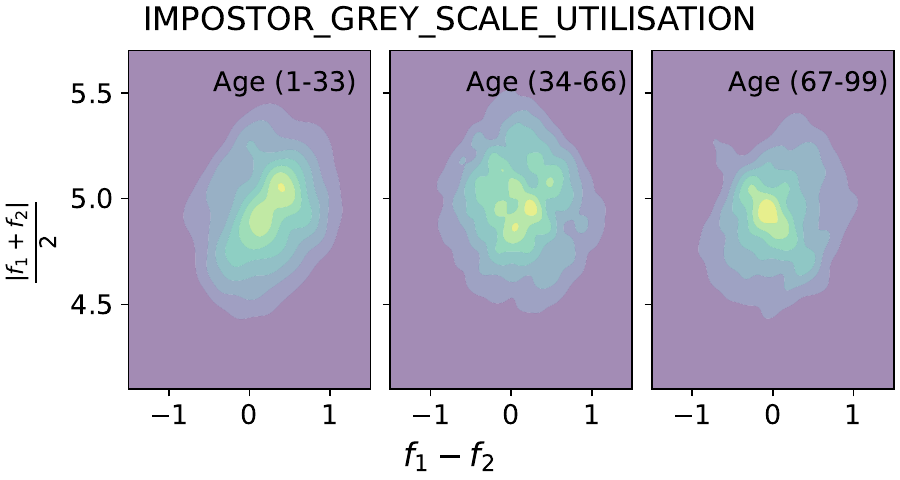}}
    \\
    \subfloat[IRIS\_RADIUS\label{1e:age-heatmap}]{%
       \includegraphics[width=0.50\textwidth]{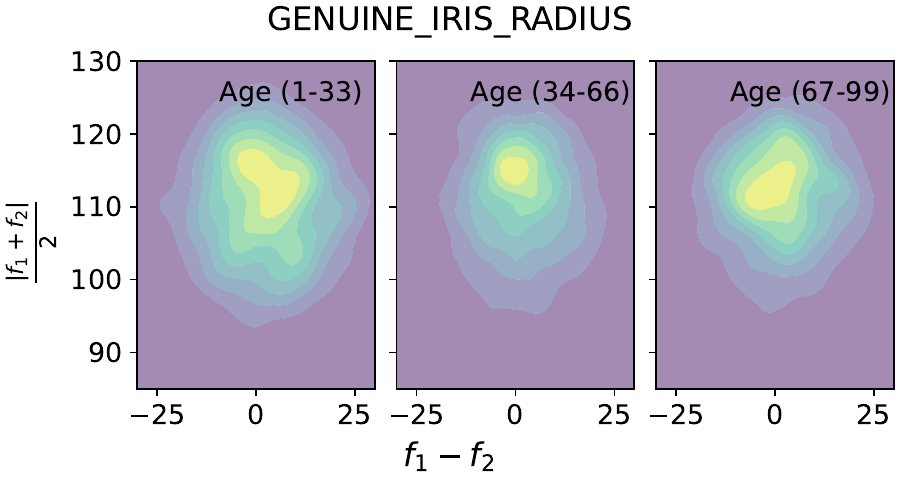}
       \includegraphics[width=0.50\textwidth]{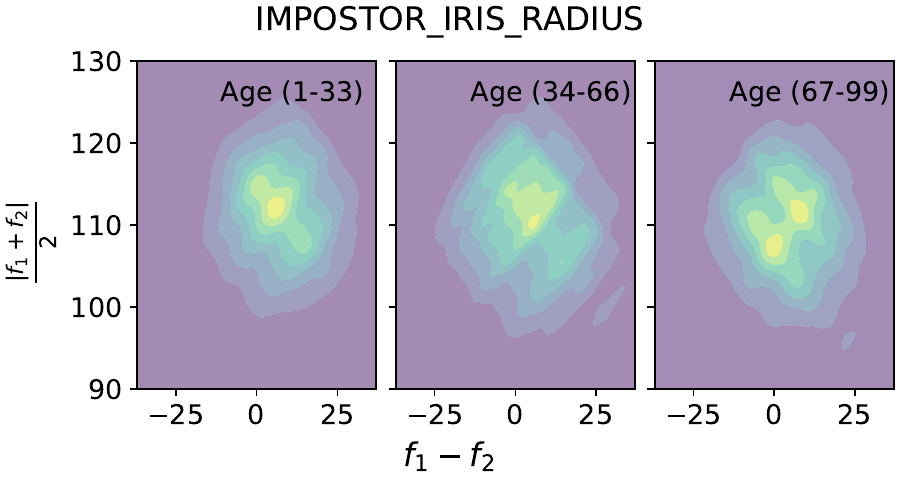}}
    \\
    \subfloat[PUPIL\_IRIS\_RATIO\label{1f:age-heatmap}]{%
       \includegraphics[width=0.50\textwidth]{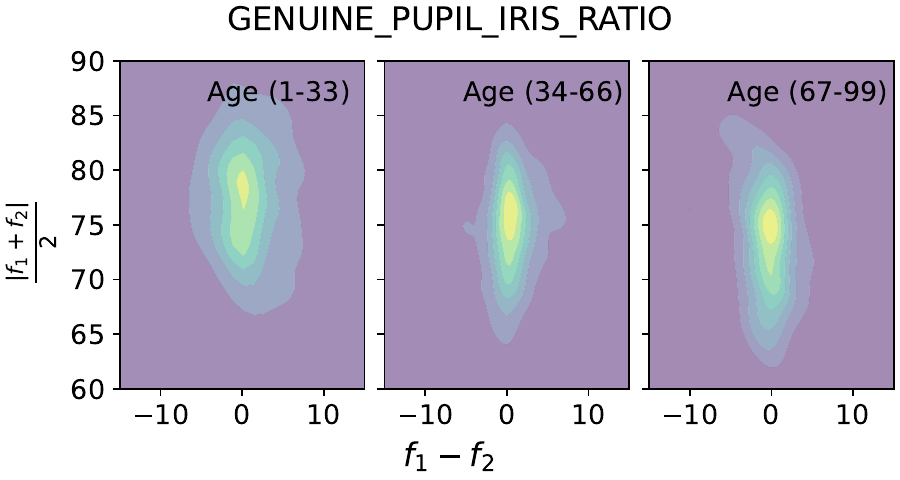}
       \includegraphics[width=0.50\textwidth]{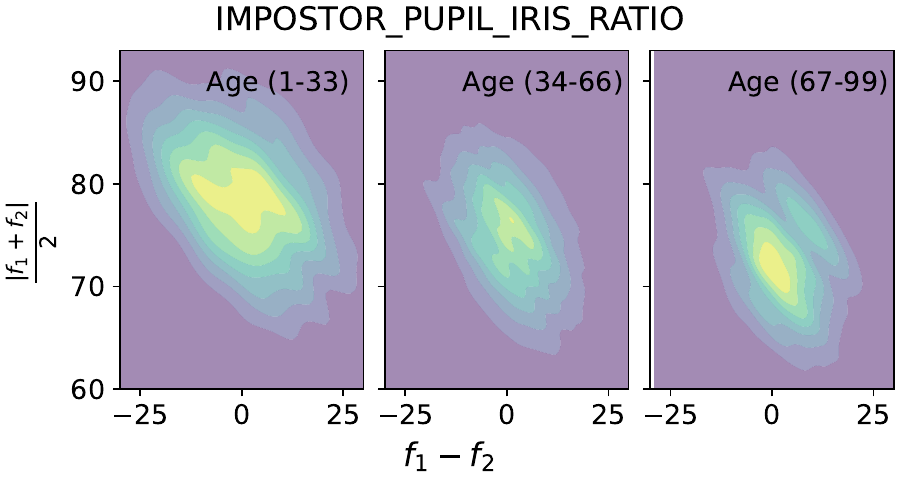}}
    \\
    \subfloat[IRIS\_PUPIL\_CONCENTRICITY\label{1g:age-heatmap}]{%
       \includegraphics[width=0.50\textwidth]{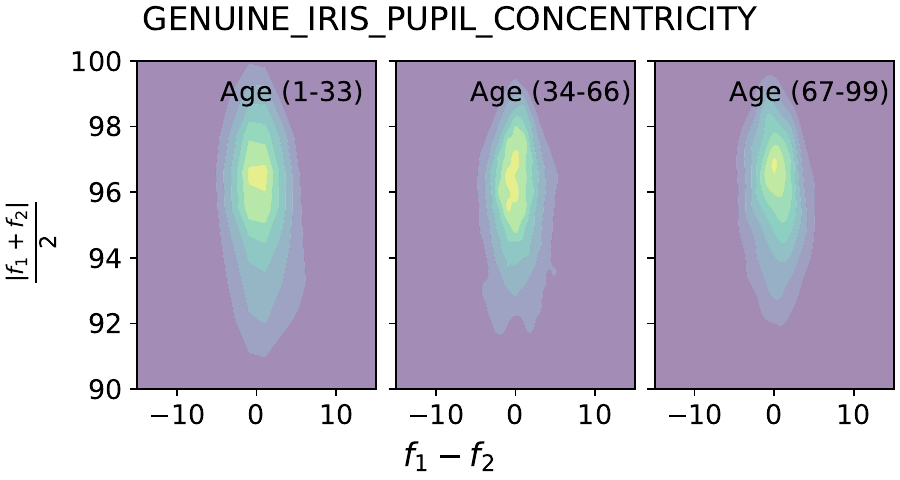}
       \includegraphics[width=0.50\textwidth]{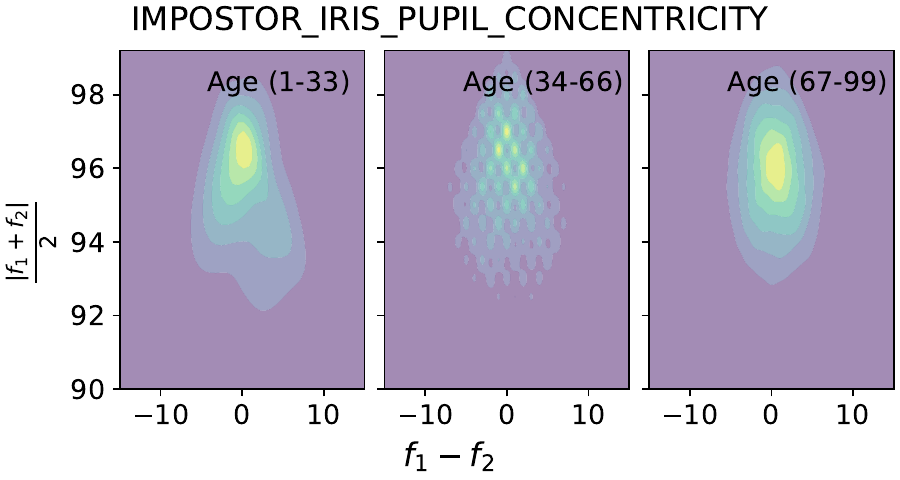}}
    \\
    \subfloat[SHARPNESS\label{1h:age-heatmap}]{%
        \includegraphics[width=0.50\textwidth]{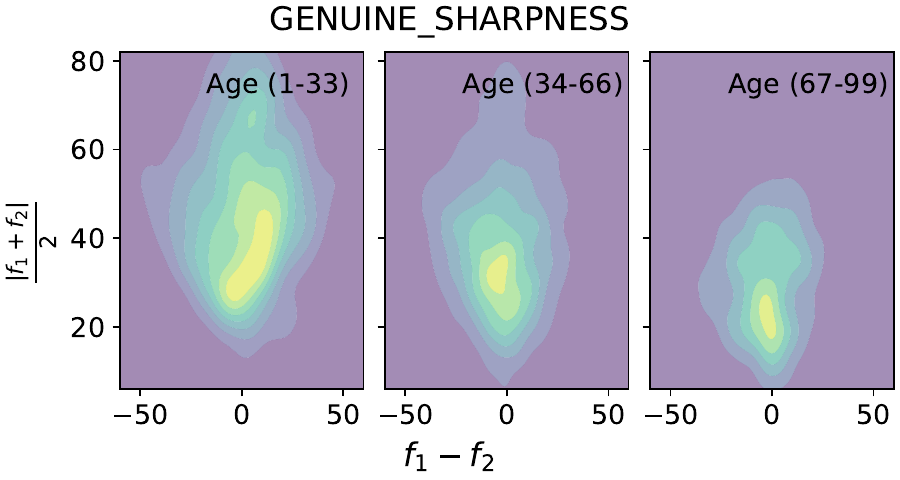}
        \includegraphics[width=0.50\textwidth]{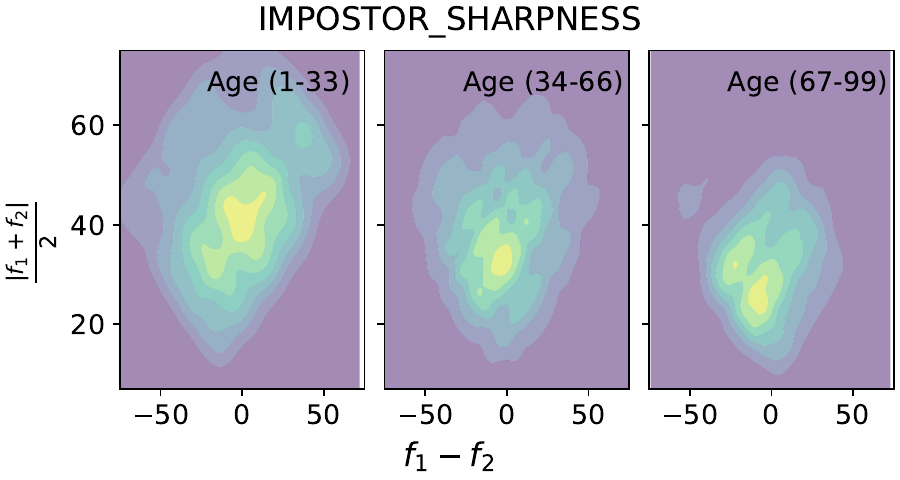}}
    \caption{ISO/IEC 29794-6 quality metrics density heatmaps for three age groups, and split by the type of comparison (genuine and impostor) for the combined \datasetname~and Warsaw datasets.}
    \label{fig:age-heatmap} 
  \end{minipage}  
\end{figure}

\end{document}